\documentclass[10pt]{article}
\usepackage{multirow}
\usepackage{tabularx}
\usepackage{comment}
\usepackage{url}
\usepackage{xcolor}
\usepackage{amsmath}
\usepackage{subcaption}
\usepackage{graphicx}
\usepackage{hyperref}
\usepackage{tikz}
\usetikzlibrary{fit,arrows,calc,positioning,shadows.blur}
\usepackage[edges]{forest}
\usepackage{color}
\usepackage{amssymb}
\usepackage{amsfonts}
\usepackage{mathtools}
\usepackage[capitalise]{cleveref}
\usepackage{makecell}
\newcommand{\verteq}{\rotatebox{90}{$\,=$}}
\newcommand{\equalto}[2]{\underset{\scriptstyle\overset{\mkern4mu\verteq}{#2}}{#1}}

\newtheorem{definition}{Definition}[section]

\newcommand*{\eqtext}{=}
\newcommand*{\comma}{,}
\newcommand{\name}{\textsc{GraphEDM}}
\newcommand{\framework}{\textsc{GCF}}
\usepackage{times}

\pdfoutput=1

\usepackage{authblk}
\usepackage[numbers,sort]{natbib}

\usepackage{hyperref}       
\hypersetup{colorlinks,linkcolor=blue,citecolor=blue,urlcolor=blue}
\newlength{\defbaselineskip}
\usepackage{etoolbox}

\newtoggle{arxiv}
\toggletrue{arxiv}
\usepackage[numbers,sort]{natbib}


\begin{document}
\title{Machine Learning on Graphs:\\ A Model and Comprehensive Taxonomy}
\author[$\dagger$]{Ines Chami\thanks{Work partially done during an internship at Google Research.}}
\author[$\ddagger$]{Sami Abu-El-Haija}
\author[$\dagger\dagger$]{Bryan Perozzi}
\author[$\ddagger\ddagger$]{Christopher R\'e}
\author[$\dagger\dagger$]{Kevin Murphy}

\affil[$\dagger$]{\small{Stanford University, Institute for Computational and Mathematical Engineering}}
\affil[$\ddagger$]{\small{University of Southern California, Information Sciences Institute}}
\affil[$\ddagger\ddagger$]{\small{Stanford University, Department of Computer Science}}
\affil[$\dagger\dagger$]{\small{Google Research}}
\affil[ ]{\footnotesize{\texttt{\{chami,chrismre\}@cs.stanford.edu},\ \texttt{sami@haija.org},\ \texttt{bperozzi@acm.org},\ \texttt{kpmurphy@google.com}}}
\maketitle
\begin{abstract}
There has been a surge of recent interest in graph representation learning (GRL). 
GRL methods have generally fallen into three main categories, based on the availability of labeled data.
The first, network embedding, focuses on learning unsupervised representations of relational structure.
The second, graph regularized neural networks, leverages graphs to augment neural network losses with a regularization objective for semi-supervised learning. 
The third, graph neural networks, aims to learn differentiable functions over discrete topologies with arbitrary structure.
However, despite the popularity of these areas there has been surprisingly little work on unifying the three paradigms.
Here, we aim to bridge the gap between network embedding, graph regularization and graph neural networks.
We propose a comprehensive taxonomy of GRL methods, aiming to unify several disparate bodies of work.
Specifically, we propose the \name{} framework, which generalizes popular algorithms for semi-supervised learning (e.g.\ GraphSage, GCN, GAT), and unsupervised learning (e.g.\ DeepWalk, node2vec) of graph representations into a single consistent approach.
To illustrate the generality of \name{}, we fit over thirty existing methods into this framework.
We believe that this unifying view both provides a solid foundation for understanding the intuition behind these methods, and enables future research in the area.

\end{abstract}
\newpage
\tableofcontents
\newpage
\defcitealias{10.1093/nar/gkj109}{Stark, 2006}
\defcitealias{abu2016metric}{Abu-ata, 2016}
\defcitealias{abu2017watch}{Abu-el-haija, 2018}
\defcitealias{abu2019mixhop}{Abu-el-haija, 2019}
\defcitealias{adcock2013tree}{Adcock, 2013}
\defcitealias{alanis2016efficient}{Alanis-lobato, 2016}
\defcitealias{ahmed2013distributed}{Ahmed, 2013}
\defcitealias{almeida1987learning}{Almeida, 1987}
\defcitealias{alrfou2019ddgk}{Al-rfou, 2019}
\defcitealias{atwood2016diffusion}{Atwood, 2016}
\defcitealias{balazevic2019multi}{Balazevic, 2019}
\defcitealias{battaglia2016interaction}{Battaglia, 2016}
\defcitealias{battaglia2018relational}{Battaglia, 2018}
\defcitealias{barabasi1999emergence}{Barab{\'a}si, 1999}
\defcitealias{becigneul2018riemannian}{Becigneul, 2018}
\defcitealias{belkin2002laplacian}{Belkin, 2002}
\defcitealias{belkin2004semi}{Belkin, 2004}
\defcitealias{belkin2006manifold}{Belkin, 2006}
\defcitealias{bengio200611}{Bengio, 2006}
\defcitealias{bengio2018machine}{Bengio, 2018}
\defcitealias{berg2017graph}{Berg, 2017}
\defcitealias{bojchevski2018netgan}{Bojchevski, 2018}
\defcitealias{boscaini2016learning}{Boscaini, 2016}
\defcitealias{bonnabel2013stochastic}{Bonnabel, 2013}
\defcitealias{bordes2013translating}{Bordes, 2013}
\defcitealias{bronstein2017geometric}{Bronstein, 2017}
\defcitealias{bruna2013spectral}{Bruna, 2014}
\defcitealias{bruna18fewshot}{Garcia, 2018}
\defcitealias{bui2017neural}{Bui, 2018}
\defcitealias{cai2018comprehensive}{Cai, 2018}
\defcitealias{cao2015grarep}{Cao, 2015}
\defcitealias{cao2016deep}{Cao, 2016}
\defcitealias{chamberlain2017neural}{Chamberlain, 2017}
\defcitealias{chami2019hyperbolic}{Chami, 2019}
\defcitealias{chami2020low}{Chami, 2020}
\defcitealias{chapellesemi}{Chapelle, 2009}
\defcitealias{chen2013hyperbolicity}{Chen, 2013}
\defcitealias{chen2017harp}{Chen, 2018}
\defcitealias{chen2018edgelabels}{Chen, 2018}
\defcitealias{chen2018fastgcn}{Chen, 2018}
\defcitealias{chen2018tutorial}{Chen, 2018}
\defcitealias{chen2019supervised}{Chen, 2019}
\defcitealias{chen2019equivalence}{Chen, 2019}
\defcitealias{chiang2019clustergcn}{Chiang, 2019}
\defcitealias{cho2014properties}{Cho, 2014}
\defcitealias{dai2018adversarial}{Dai, 2018}
\defcitealias{de2018molgan}{De cao, 2018}
\defcitealias{de2018representation}{Sala, 2018}
\defcitealias{dean08mapreduce}{Dean, 2008}
\defcitealias{defferrard2016convolutional}{Defferrard, 2016}
\defcitealias{deng2009imagenet}{Deng, 2009}
\defcitealias{duvenaud2015convolutional}{Duvenaud, 2015}
\defcitealias{dwivedi2020benchmarking}{Dwivedi, 2020}
\defcitealias{elton2019molecular}{Elton, 2019}
\defcitealias{feng2018padme}{Feng, 2018}
\defcitealias{fey2019fast}{Fey, 2019}
\defcitealias{ganea2018hyperbolic}{Ganea, 2018}
\defcitealias{gao2018large}{Gao, 2018}
\defcitealias{garg2020generalization}{Garg, 2020}
\defcitealias{gilmer2017neural}{Gilmer, 2017}
\defcitealias{godec_2018}{Godec, 2018}
\defcitealias{gori2005new}{Gori, 2005}
\defcitealias{lerer2019biggraph}{Lerer, 2019}
\defcitealias{goyal2018gem}{Goyal, 2018}
\defcitealias{goyal2018graph}{Goyal, 2018}
\defcitealias{grover2016node2vec}{Grover, 2016}
\defcitealias{grover2018graphite}{Grover, 2019}
\defcitealias{hamilton2017inductive}{Hamilton, 2017}
\defcitealias{gu2018learning}{Gu, 2018}
\defcitealias{haija2017lowrank}{Abu-el-haija, 2017}
\defcitealias{hamilton2017representation}{Hamilton, 2017}
\defcitealias{hammond2011wavelets}{Hammond, 2011}
\defcitealias{henaff2015deep}{Henaff, 2015}
\defcitealias{hinton2006reducing}{Hinton, 2006}
\defcitealias{hochreiter1997long}{Hochreiter, 1997}
\defcitealias{huang2020personalized}{Huang, 2020}
\defcitealias{jin2018junction}{Jin, 2018}
\defcitealias{jolliffe2011principal}{Jolliffe, 2011}
\defcitealias{jonckheere2008scaled}{Jonckheere, 2008}
\defcitealias{kearnes2016molecular}{Kearnes, 2016}
\defcitealias{khalil2017learning}{Khalil, 2017}
\defcitealias{kipf2016semi}{Kipf, 2016}
\defcitealias{kipf2016variational}{Kipf, 2016}
\defcitealias{kipf18a}{Kipf, 2018}
\defcitealias{kleinberg2007geographic}{Kleinberg, 2007}
\defcitealias{krishna2018referring}{Krishna, 2018}
\defcitealias{krioukov2010hyperbolic}{Krioukov, 2010}
\defcitealias{kruskal1964multidimensional}{Kruskal, 1964}
\defcitealias{kuznetsova2020openimages}{Kuznetsova, 2020}
\defcitealias{lamb2020graph}{Lamb, 2020}
\defcitealias{le2019inferring}{Le, 2019}
\defcitealias{lecun1989backpropagation}{Lecun, 1989}
\defcitealias{leskovec10www}{Leskovec, 2010}
\defcitealias{li2017diffusion}{Li, 2017}
\defcitealias{liben2007link}{Liben-nowell, 2007}
\defcitealias{li2015gated}{Li, 2015}
\defcitealias{li2018learning}{Li, 2018}
\defcitealias{liu2018constrained}{Liu, 2018}
\defcitealias{liu2019hyperbolic}{Liu, 2019}
\defcitealias{litany2017deformable}{Litany, 2017}
\defcitealias{loukas2019graph}{Loukas, 2019}
\defcitealias{maaten2008visualizing}{Maaten, 2008}
\defcitealias{maron2018invariant}{Maron, 2018}
\defcitealias{masci2015geodesic}{Masci, 2015}
\defcitealias{mikolov2013distributed}{Mikolov, 2013}
\defcitealias{mikolov2013efficient}{Mikolov, 2013}
\defcitealias{mcauley12nips}{Mcauley, 2012}
\defcitealias{monti2017geometric}{Monti, 2017}
\defcitealias{morris2019weisfeiler}{Morris, 2019}
\defcitealias{mrowca2018flexible}{Mrowca, 2018}
\defcitealias{muscoloni2017machine}{Muscoloni, 2017}
\defcitealias{nickel2017poincare}{Nickel, 2017}
\defcitealias{nickel2018learning}{Nickel, 2018}
\defcitealias{niepert2016learning}{Niepert, 2016}
\defcitealias{nowak2017revised}{Nowak, 2017}
\defcitealias{ogb_2020}{Hu, 2020}
\defcitealias{oliver2018realistic}{Oliver, 2018}
\defcitealias{ou2016asymmetric}{Ou, 2016}
\defcitealias{markowitz2021graph}{Markowitz, 2021}
\defcitealias{papadopoulos2012popularity}{Papadopoulos, 2012}
\defcitealias{papadopoulos2014network}{Papadopoulos, 2014}
\defcitealias{peng2020graph}{Peng, 2020}
\defcitealias{pennington2014glove}{Pennington, 2014}
\defcitealias{perozzi2014deepwalk}{Perozzi, 2014}
\defcitealias{perozzi2017don}{Perozzi, 2017}
\defcitealias{pham2017column}{Pham, 2017}
\defcitealias{pineda1988generalization}{Pineda, 1988}
\defcitealias{prates2019learning}{Prates, 2019}
\defcitealias{Ragoza2017proteinligand}{Ragoza, 2017}
\defcitealias{qm9dataset}{Ramakrishnan, 2014}
\defcitealias{qi2017pointnet}{Qi, 2017}
\defcitealias{qi2017pointnet++}{Qi, 2017}
\defcitealias{ribeiro2017struc2vec}{Ribeiro, 2017}
\defcitealias{roweis2000nonlinear}{Roweis, 2000}
\defcitealias{rozemberczki2019gemsec}{Rozemberczki, 2019}
\defcitealias{sarkar2011low}{Sarkar, 2011}
\defcitealias{scarselli2009graph}{Scarselli, 2009}
\defcitealias{schlichtkrull2018modeling}{Schlichtkrull, 2018}
\defcitealias{shchur2018pitfalls}{Shchur, 2018}
\defcitealias{selsam2018learning}{Selsam, 2018}
\defcitealias{shaw2009structure}{Shaw, 2009}
\defcitealias{simonovsky2017dynamic}{Simonovsky, 2017}
\defcitealias{simonovsky2018graphvae}{Simonovsky, 2018}
\defcitealias{sinha2020evaluating}{Sinha, 2020}
\defcitealias{subramanian05}{Subramanian, 2005}
\defcitealias{socrecgraph}{Konstas, 2009}
\defcitealias{tang09relational}{Tang, 2009}
\defcitealias{tang2015line}{Tang, 2015}
\defcitealias{tenenbaum2000global}{Tenenbaum, 2000}
\defcitealias{tifrea2018poincare}{Tifrea, 2018}
\defcitealias{tsitsulin2018netlsd}{Tsitsulin, 2018}
\defcitealias{tsitsulin2020slaq}{Tsitsulin, 2020}
\defcitealias{vaswani2017attention}{Vaswani, 2017}
\defcitealias{vincent2010stacked}{Vincent, 2010}
\defcitealias{velickovic2018dgi}{Veli{\v{c}}kovi{\'{c}}, 2019}
\defcitealias{velickovic2017graph}{Veli{\v{c}}kovi{\'{c}}, 2018}
\defcitealias{verma2018feastnet}{Verma, 2018}
\defcitealias{verma2019stability}{Verma, 2019}
\defcitealias{von2007tutorial}{Von luxburg, 2007}
\defcitealias{wang2016structural}{Wang, 2016}
\defcitealias{wang2019deep}{Wang, 2019}
\defcitealias{weston2012deep}{Weston, 2008}
\defcitealias{wold1987principal}{Wold, 1987}
\defcitealias{wu2019comprehensive}{Wu, 2019}
\defcitealias{wu2019simplifying}{Wu, 2019}
\defcitealias{xu2018powerful}{Xu, 2018}
\defcitealias{yang2016revisiting}{Yang, 2016}
\defcitealias{ying2018hierarchical}{Ying, 2018}
\defcitealias{ying2018pinsage}{Ying, 2018}
\defcitealias{you2018graphrnn}{You, 2018}
\defcitealias{yu2018spatio}{Yu, 2018}
\defcitealias{yu2019numerically}{Yu, 2019}
\defcitealias{zhang2018deep}{Zhang, 2018}
\defcitealias{zhang2018end}{Zhang, 2018}
\defcitealias{zhang2018network}{Zhang, 2018}
\defcitealias{zhou2004learning}{Zhou, 2004}
\defcitealias{zhou2018graph}{Zhou, 2018}
\defcitealias{zhu2002learning}{Zhu, 2002}
\defcitealias{zhu2003semi}{Zhu, 2003}
\defcitealias{zugner2018adversarial}{Z{\"u}gner, 2018}
\defcitealias{mehrabi2019survey}{Mehrabi, 2019}
\defcitealias{bose2019compositional}{Bose, 2019}
\defcitealias{gonen2019lipstick}{Gonen, 2019}
\defcitealias{palowitch2019monet}{Palowitch, 2019}
\defcitealias{levy2014neural}{Levy, 2014}
\defcitealias{qiu2018network}{Qiu, 2018}
\defcitealias{qiu2019NetSMF}{Qiu, 2019}
\defcitealias{epasto2019splitter}{Epasto, 2019}
\defcitealias{epasto2017persona}{Epasto, 2017}
\defcitealias{bojchevski2019pagerank}{Bojchevski, 2019}
\defcitealias{Srinivasan2020On}{Srinivasan, 2020}
\defcitealias{you2019position}{You, 2019}

\section{Introduction}
Learning representations for complex structured data is a challenging task. 
In the last decade, many successful models have been developed for certain kinds of structured data, including data defined on a discretized Euclidean domain. 
For instance, sequential data, such as text or videos, can be modelled via recurrent neural networks, which can capture sequential information, yielding efficient representations as measured on machine translation and speech recognition tasks.
Another example is convolutional neural networks (CNNs), which parameterize neural networks according to structural priors such as shift-invariance, and have achieved unprecedented performance in pattern recognition tasks such as image classification or speech recognition. 
These major successes have been restricted to particular types of data that have a simple relational structure (e.g. sequential data, or data following regular patterns).  

\begin{figure*}
\centering
\begin{subfigure}[b]{0.35\textwidth}
        \vspace{20pt}
        \includegraphics[width=\textwidth]{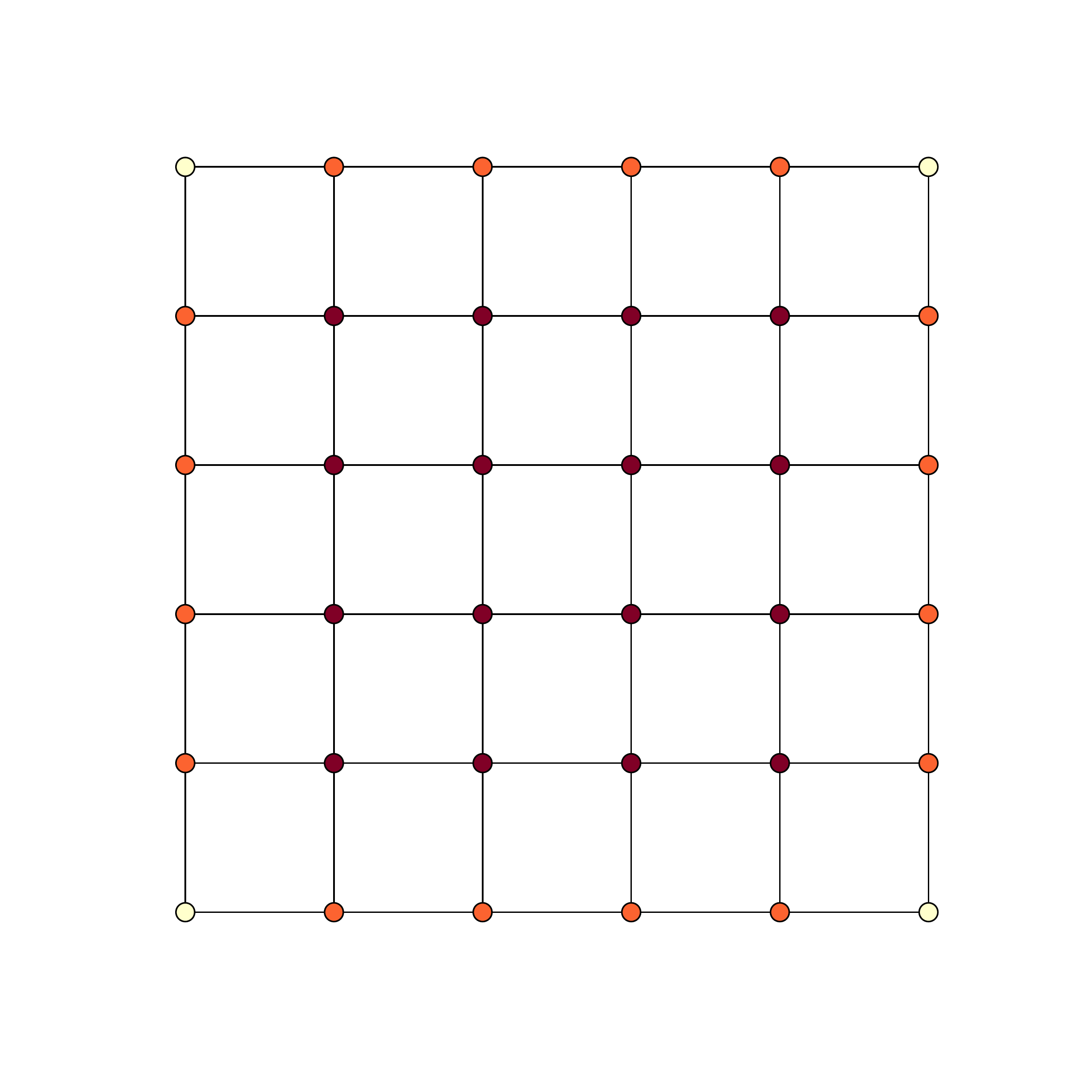}
        \caption{Grid (Euclidean).}
        \label{fig:euclidean}
    \end{subfigure}
    \begin{subfigure}[b]{0.45\textwidth}
        \includegraphics[width=\textwidth]{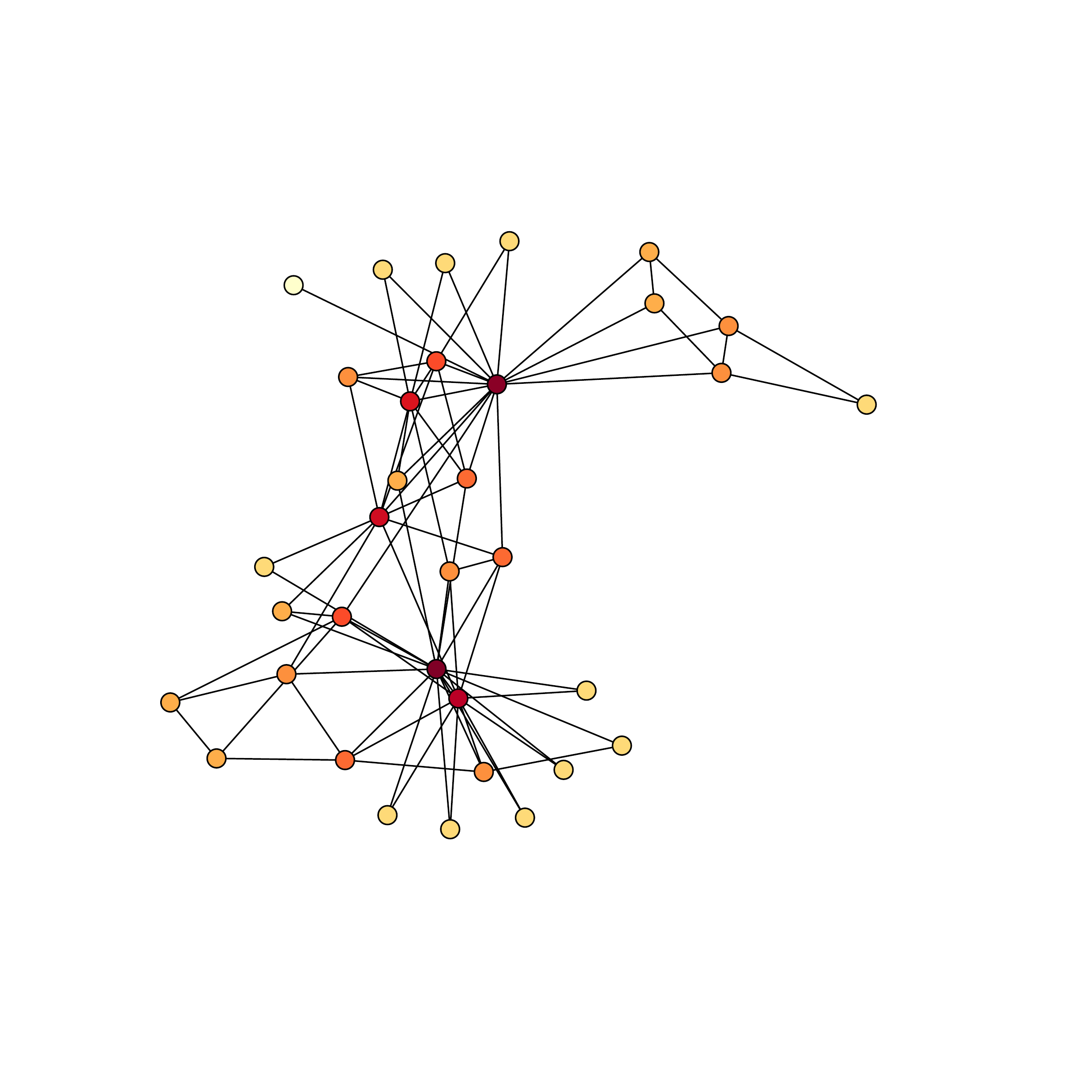}
        \vspace{-45pt}
        \caption{Arbitrary graph (Non-Euclidean).}
        \label{fig:non_euclidean}
\end{subfigure}
\caption{An illustration of Euclidean vs. non-Euclidean graphs.}\label{fig:non_euclidean_vs_euclidean}
\end{figure*}

In many settings, data is not nearly as regular: complex relational structures commonly arise, and extracting information from that structure is key to understanding how objects interact with each other.
Graphs are a universal data structures that can represent complex relational data (composed of nodes and edges), and appear in multiple domains such as social networks, computational chemistry \citep{gilmer2017neural}, biology \citep{10.1093/nar/gkj109}, recommendation systems \citep{socrecgraph}, semi-supervised learning \citep{bruna18fewshot}, and others.
For graph-structured data, it is challenging to define networks with strong structural priors, as structures can be arbitrary, and can vary significantly across different graphs and even different nodes within the same graph. 
In particular, operations like convolutions cannot be directly applied on irregular graph domains. For instance in images, each pixel has the same neighborhood structure, allowing to apply the same filter weights at multiple locations in the image. 
However in graphs, one can’t define an ordering of node since each node might have a different neighborhood structure~(\cref{fig:non_euclidean_vs_euclidean}).
Furthermore, Euclidean convolutions strongly rely on geometric priors (e.g. shift invariance) which don't generalize to non-Euclidean domains (e.g. translations might not even be defined on non-Euclidean domains).

These challenges led to the development of Geometric Deep Learning (GDL) research which aims at applying deep learning techniques to non-Euclidean data.
In particular, given the widespread prevalence of graphs in real-world applications, there has been a surge of interest in applying machine learning methods to graph-structured data. 
Among these, Graph Representation Learning (GRL) methods aim at learning low-dimensional continuous vector representations for graph-structured data, also called embeddings.

Broadly speaking, GRL can be divided into two classes of learning problems, \textbf{unsupervised} and \textbf{supervised} (or semi-supervised) GRL. 
The first family aims at learning low-dimensional Euclidean representations that preserve the structure of an input graph. 
The second family also learns low-dimensional Euclidean representations but for a specific downstream prediction task such as node or graph classification. 
Different from the unsupervised setting where inputs are usually graph structures, inputs in supervised settings are usually composed of different signals defined on graphs, commonly known as \textbf{node features}. 
Additionally, the underlying discrete graph domain can be fixed, which is the \textbf{transductive} learning setting (e.g. predicting user properties in a large social network), but can also vary in the \textbf{inductive} learning setting (e.g. predicting molecules attribute where each molecule is a graph).
Finally, note that while most supervised and unsupervised methods learn representations in Euclidean vector spaces, there recently has been interest for \textbf{non-Euclidean representation learning}, which aims at learning non-Euclidean embedding spaces such as hyperbolic or spherical spaces.
The main motivations for this body of work is to use a {continuous embedding space that resembles the underlying discrete structure of the input data} it tries to embed (e.g. the hyperbolic space is a continuous version of trees \citep{sarkar2011low}). 

Given the impressive pace at which the field of GRL is growing, we believe it is important to summarize and describe all methods in one unified and comprehensible framework. 
The goal of this survey is to provide a unified view of representation learning methods for graph-structured data, to better understand the different ways to leverage graph structure in deep learning models. 

A number of graph representation learning surveys exist. 
First, there exist several surveys that cover shallow network embedding and auto-encoding techniques and we refer to \citep{cai2018comprehensive,chen2018tutorial,goyal2018graph,hamilton2017representation,zhang2018network} for a detailed overview of these methods. 
Second, \citet{bronstein2017geometric} also gives an extensive overview of deep learning models for non-Euclidean data such as graphs or manifolds. 
Third, there have been several recent surveys \citep{battaglia2018relational,wu2019comprehensive,zhang2018deep,zhou2018graph} covering methods applying deep learning to graphs, including graph neural networks. Most of these surveys focus on a specific sub-field of graph representation learning and do not draw connections between each sub-field. 

In this work, we extend the encoder-decoder framework proposed by \citet{hamilton2017representation} and introduce a general framework, the  Graph Encoder Decoder Model (\name{}), which allows us to group existing work into four major categories: (i) shallow embedding methods, (ii) auto-encoding methods, (iii) graph regularization methods, and (iv) graph neural networks (GNNs). 
Additionally, we introduce a Graph Convolution Framework (\framework), specifically designed to describe convolution-based GNNs, which have achieved state-of-the art performance in a broad range of applications. 
This allows us to analyze and compare a variety of GNNs, ranging in construction from methods operating in the Graph Fourier\footnote{As defined by the eigenspace of the graph Laplacian.} domain to methods applying self-attention as a neighborhood aggregation function \citep{velickovic2017graph}. 
We hope that this unified formalization of recent work would help the reader gain insights into the various learning methods on graphs to reason about similarities, differences, and point out potential extensions and limitations.
That said, our contribution with regards to previous surveys are threefold:
\begin{itemize}
    \itemsep -2pt
    \item We introduce a general framework, \name, to describe a broad range of supervised and unsupervised methods that operate on graph-structured data, namely shallow embedding methods, graph regularization methods, graph auto-encoding methods and graph neural networks. 
    \item Our survey is the first attempt to unify and view these different lines of work from the same perspective, and we provide a general taxonomy~(\cref{fig:taxonomy}) to understand differences and similarities between these methods. 
    In particular, this taxonomy encapsulates over thirty existing GRL methods. 
    Describing these methods within a comprehensive taxonomy gives insight to exactly how these methods differ. 
    \item We release an open-source library for GRL which includes state-of-the-art GRL methods and important graph applications, including node classification and link prediction. 
    Our implementation is publicly available at \url{https://github.com/google/gcnn-survey-paper}.
\end{itemize}
\paragraph{Organization of the survey}
We first review basic graph definitions and clearly state the problem setting for GRL~(\cref{sec:setup}).
In particular, we define and discuss the differences between important concepts in GRL, including the role of node features in GRL and how they relate to supervised GRL~(\cref{subsec:node_features}), the distinctions between inductive and transductive learning~(\cref{subsec:inductive_transductive}), positional and structural embeddings~(\cref{subsec:pos_struct}) and the differences between supervised and unsupervised embeddings~(\cref{subsec:supervision}). 
We then introduce \name{}~(\cref{sec:framework}) a general framework to describe both supervised and unsupervised GRL methods, with or without the presence of node features, which can be applied in both inductive and transductive learning settings. 
Based on \name{}, we introduce a general taxonomy of GRL methods~(\cref{fig:taxonomy}) which encapsulates over thirty recent GRL models, and we describe both unsupervised~(\cref{sec:unsup_models}) and supervised~(\cref{sec:sup_models}) methods using this taxonomy.
Finally, we survey graph applications~(\cref{sec:applications}).

\section{Preliminaries}
\label{sec:setup}
Here we introduce the notation used throughout this article (see~\cref{tab:notation} for a summary), and the generalized network embedding problem which graph representation learning methods aim to solve.

\begin{table}
\centering
\resizebox{\textwidth}{!}{\renewcommand{\arraystretch}{1.1}
\begin{tabular}{c | c | l}
\hline
&  \bfseries Notation & \bfseries Meaning \\
\hline
\multirow{4}{*}{{Abbreviations}} & GRL & Graph Representation Learning \\ 
& \name{} & Graph Encoder Decoder Model \\
& GNN & Graph Neural Network \\
& \framework{} & Graph Convolution Framework \\
\hline
\multirow{11}{*}{{Graph notation}} & $G=(V, E)$ & Graph with vertices (nodes) $V$ and edges $E$ \\
& $v_i\in V$ & Graph vertex \\
& $d_G(\cdot,\cdot)$ & Graph distance (length of shortest path) \\
& $\mathrm{deg}(\cdot)$ & Node degree \\
& $D\in\mathbb{R}^{|V|\times |V|}$ & Diagonal degree matrix \\
& $W\in\mathbb{R}^{|V|\times |V|}$ & Graph weighted adjacency matrix \\
& $\widetilde{W}\in\mathbb{R}^{|V|\times |V|}$ & Symmetric normalized adjacency matrix ($\widetilde{W}=D^{-1/2}WD^{-1/2}$)\\
& $A\in\{0,1\}^{|V|\times |V|}$ & Graph unweighted weighted adjacency matrix \\
& $L\in\mathbb{R}^{|V|\times |V|}$ & Graph unnormalized Laplacian matrix ($L=D-W$) \\
& $\widetilde{L}\in\mathbb{R}^{|V|\times |V|}$ & Graph normalized Laplacian matrix ($\widetilde{L}=I-D^{-1/2}WD^{-1/2}$) \\
& $L^{\mathrm{rw}}\in\mathbb{R}^{|V|\times |V|}$ & Random walk normalized Laplacian ($L^{\mathrm{rw}}=I-D^{-1}W$) \\
\hline
\multirow{21}{*}{{\name{} notation}} &  $d_0$ & Input feature dimension \\

& $X\in\mathbb{R}^{|V|\times d_0}$ & Node feature matrix \\
& $d$ & Final embedding dimension \\ 
& $Z\in\mathbb{R}^{|V|\times d}$ & Node embedding matrix \\
& $d_\ell$ & Intermediate hidden embedding dimension at layer $\ell$ \\ 
& $H^\ell\in\mathbb{R}^{|V|\times d_\ell}$ & Hidden representation at layer $\ell$ \\
& $\mathcal{Y}$ & Label space \\
& $y^S\in\mathbb{R}^{|V|\times |\mathcal{Y}|}$ & Graph ($S=G$) or node ($S=N$) ground truth labels \\
& $\hat{y}^S\in\mathbb{R}^{|V|\times |\mathcal{Y}|}$ & Predicted labels \\
& $s(W)\in\mathbb{R}^{|V|\times |V|}$ & Target similarity or dissimilarity matrix in graph regularization \\
& $\widehat{W}\in\mathbb{R}^{|V|\times |V|}$ & Predicted similarity or dissimilarity matrix \\
& $\mathrm{ENC}(\cdot;\Theta^E)$ & Encoder network with parameters $\Theta^E$ \\
& $\mathrm{DEC}(\cdot;\Theta^D)$ & Graph decoder network with parameters $\Theta^D$ \\
& $\mathrm{DEC}(\cdot;\Theta^S)$ & Label decoder network with parameters $\Theta^S$ \\
& $\mathcal{L}^S_\mathrm{SUP}(y^S, \hat{y}^S;\Theta)$ & Supervised loss \\
& $\mathcal{L}_{G,\mathrm{REG}}(W, \widehat{W};\Theta)$ & Graph regularization loss \\
& $\mathcal{L}_{\mathrm{REG}}(\Theta)$ & Parameters' regularization loss \\
& $d_1(\cdot, \cdot)$ & Matrix distance used for to compute the graph regularization loss \\
& $d_2(\cdot, \cdot)$ & Embedding distance for distance-based decoders \\
& $||\cdot||_p$ & $p-$norm \\
& $||\cdot||_F$ & Frobenuis norm \\
\hline
\end{tabular}}
\caption{Summary of the notation used in the paper.}\label{tab:notation}
\end{table}
\subsection{Definitions}\label{subsec:def}
\begin{definition}{(\textit{Graph})}. A graph $G$ given as a pair:  ${G}=(V, E)$, comprises a set of vertices (or nodes) $V=\{v_1,\dots,v_{|V|}\}$ connected by edges $E=\{e_1,\ldots,e_{|E|}\}$, where each edge $e_k$ is a pair $(v_i,v_j)$ with $v_i,v_j \in V$. 
A graph is \textit{weighted} if there exist a weight function: $w:(v_i,v_j)\rightarrow w_{ij}$ that assigns weight $w_{ij}$ to edge connecting nodes $v_i, v_j \in V$. Otherwise, we say that the graph is \textit{unweighted}. 
A graph is \textit{undirected} if $(v_i, v_j)\in E$ implies $(v_j, v_i)\in E$, i.e. the relationships are symmetric, and \textit{directed} if the existence of edge $(v_i, v_j)\in E$ does not necessarily imply $(v_j, v_i)\in E$.
Finally, a graph can be \textit{homogeneous} if nodes refer to one type of entity and edges to one relationship. It can be \textit{heterogeneous} if it contains different types of nodes and edges. 
\end{definition}
\noindent For instance, social networks are homogeneous graphs that can be undirected (e.g. to encode symmetric relations like friendship) or directed (e.g. to encode the relation {following}); weighted (e.g. co-activities) or unweighted. 

\begin{definition}{(\textit{Path})}. A path $P$ is a sequence of edges $(u_{i_1},u_{i_2}),(u_{i_2},u_{i_3}),\ldots,(u_{i_k},u_{i_{k+1}})$ of length $k$. 
	A path is called simple if all $u_{i_j}$ are distinct from each other. Otherwise, if a path visits a node more than once, it is said to contain a cycle.
\end{definition}

\begin{definition}{(\textit{Distance}).}
	Given two nodes $(u, v)$ in a graph $G$, we define the distance from $u$ to $v$, denoted $d_G(u, v)$, to be the length of the shortest path from $u$ to $v$, or $\infty$	if there exist no path from $u$ to $v$. 
\end{definition}
\noindent The graph distance between two nodes is the analog of geodesic lengths on manifolds.

\begin{definition}{(\textit{Vertex degree}).}
The \textit{degree}, $\mathrm{deg}(v_i)$, of a vertex $v_i$ in an unweighted graph is the number of edges incident to it. 
Similarly, the degree of a vertex $v_i$ in a weighted graph is the sum of incident edges weights. 
The degree matrix $D$ of a graph with vertex set $V$ is the $|V|\times|V|$ diagonal matrix such that $D_{ii}=\mathrm{deg}(v_i)$.
\end{definition}

\begin{definition}{(\textit{Adjacency matrix}).} 
A finite graph ${G}=(V, E)$
can be represented as a square $|V|\times|V|$ \textit{adjacency matrix}, where the elements of the matrix indicate whether pairs of nodes are adjacent or not. 
The adjacency matrix is binary for unweighted graph, $A \in \{0, 1\}^{|V|\times|V|}$, and non-binary for weighted graphs $W \in \mathbb{R}^{|V|\times|V|}$. 
Undirected graphs have symmetric adjacency matrices, in which case, $\widetilde{W}$ denotes symmetrically-normalized adjacency matrix: $\widetilde{W}=D^{-1/2}WD^{-1/2}$, where $D$ is the degree matrix.
\end{definition}

\begin{definition}{(\textit{Laplacian}).}
The \textit{unnormalized Laplacian} of an undirected graph is the $|V|\times|V|$ matrix
$L=D-W$. The \textit{symmetric normalized Laplacian} is
$\widetilde{L}=I-D^{-1/2}WD^{-1/2}.$ The \textit{random walk normalized Laplacian} is the matrix $L^{\mathrm{rw}}=I-D^{-1}W$. 
\end{definition}
\noindent The name random walk comes from the fact that $D^{-1}W$ is a stochastic transition matrix that can be interpreted as the transition probability matrix of a random walk on the graph.
The graph Laplacian is a key operator on graphs and can be interpreted as the analogue of the continuous Laplace-Beltrami operator on manifolds. 
Its eigenspace capture important properties about a graph (e.g. cut information often used for spectral graph clustering) but can also serve as a basis for smooth functions defined on the graph for semi-supervised learning~\citep{belkin2004semi}. 
The graph Laplacian is also closely related to the heat equation on graphs as it is the generator of diffusion processes on graphs and can be used to derive algorithms for semi-supervised learning on graphs~\citep{zhou2004learning}.

\begin{definition}{(\textit{First order proximity}).}
	The \textit{first order proximity} between two nodes $v_i$ and $v_j$ is a \textit{local} similarity measure indicated by the edge weight $w_{ij}$. 
	In other words, the first-order proximity captures the strength of an edge between node $v_i$ and node $v_j$ (should it exist). 
\end{definition}
\begin{definition}{(\textit{Second-order proximity}).}
	The \textit{second order proximity} between two nodes $v_i$ and $v_j$ is measures the similarity of their neighborhood structures. 
	Two nodes in a network will have a high second-order proximity if they tend to share many neighbors. 
\end{definition}
\noindent Note that there exist higher-order measures of proximity between nodes such as Katz Index, Adamic Adar or Rooted PageRank \citep{liben2007link}. 
These notions of node proximity are particularly important in network embedding as many algorithms are optimized to preserve some order of node proximity in the graph. 
\subsection{The generalized network embedding problem}
\textit{Network embedding} is the task that aims at learning a mapping function from a discrete graph to a  continuous domain. 
Formally, given a graph $G=(V,E)$ with weighted adjacency matrix $W\in\mathbb{R}^{|V|\times|V|}$, the goal is to learn low-dimensional vector representations $\{Z_i\}_{i\in V}$ (embeddings) for nodes in the graph $\{v_i\}_{i\in V}$,
such that important graph properties (e.g. local or global structure) are preserved in the embedding space.
For instance, if two nodes have similar connections in the original graph, their learned vector representations should be close. 
Let $Z\in\mathbb{R}^{|V|\times d}$ denote the node\footnote{Although we present the model taxonomy via embedding nodes yielding $Z\in\mathbb{R}^{|V|\times d}$, it can also be extended for models that embed an entire graph i.e. with $Z \in \mathbb{R}^d$ as a $d$-dimensional vector for the whole graph (e.g. \citep{duvenaud2015convolutional, alrfou2019ddgk}), or embed graph edges $Z \in \mathbb{R}^{|V| \times |V| \times d}$ as a (potentially sparse) 3D matrix with $Z_{u, v} \in \mathbb{R}^{d}$ representing the embedding of edge $(u, v)$.}  embedding matrix.
In practice, we often want low-dimensional embeddings ($d\ll |V|$) for 
scalability purposes.
That is, network embedding can be viewed as a dimensionality reduction technique for graph structured data, where the input data is defined on a non-Euclidean, high-dimensional, discrete domain.

\subsubsection{Node features in network embedding}\label{subsec:node_features}
\begin{definition}{(\textit{Vertex and edge fields}).} A \textit{vertex field} is a function defined on vertices $f:V\rightarrow \mathbb{R}$ and similarly an \textit{edge field} is a function defined on edges: $F:E\rightarrow \mathbb{R}$. Vertex fields and edge fields can be viewed as analogs of scalar fields and tensor fields on manifolds. 
\end{definition}

\noindent Graphs may have node attributes (e.g. gender or age in social networks; article contents for citation networks) which can be represented as multiple vertex fields, commonly referred to as \textit{node features}. 
In this survey, we denote node features with $X\in \mathbb{R}^{|V|\times d_0}$, where $d_0$ is the input feature dimension. 
Node features might provide useful information about a graph. Some network embedding algorithms leverage this information by learning mappings:
$$W, X\rightarrow Z.$$
In other scenarios, node features might be unavailable or not useful for a given task: network embedding can be \textit{featureless}. 
That is, the goal is to learn graph representations via mappings:
$$W \rightarrow Z.$$

\noindent Note that depending on whether node features are used or not in the embedding algorithm, the learned representation could capture different aspects about the graph. 
If nodes features are being used, embeddings could capture both \textit{structural} and \textit{semantic} graph information.
On the other hand, if node features are not being used, embeddings will only preserve structural information of the graph. 

Finally, note that edge features are less common than node features in practice, but can also be used by embedding algorithms.
For instance, edge features can be used as regularization for node embeddings \citep{chen2018edgelabels}, or to compute messages from neighbors as in message passing networks \citep{gilmer2017neural}.

\subsubsection{Transductive and inductive network embedding}\label{subsec:inductive_transductive}
Historically, a popular way of categorizing a network embedding method has been by whether the model can generalize to unseen data instances -- methods are referred to as operating in either a \textit{transductive} or \textit{inductive} setting \citep{yang2016revisiting}.  While we do not use this concept for constructing our taxonomy, we include a brief discussion here for completeness.

In transductive settings, it assumed that all nodes in the graph are observed in training (typically the nodes all come from one fixed graph).
These methods are used to infer information about or between observed nodes in the graph (e.g. predicting labels for all nodes, given a partial labeling).
For instance, if a transductive method is used to embed the nodes of a social network, it can be used to suggest new edges (e.g.\ friendships) between the nodes of the graph.
One major limitation of models learned in transductive settings is that they fail to generalize to new nodes (e.g. evolving graphs) or new graph instances. 

On the other hand, in inductive settings, models are expected to generalize to
new nodes, edges, or graphs that were not observed during training. 
Formally, given training graphs $(G_1,\ldots,G_{k})$, the goal is to learn a mapping to continuous representations that can generalize to unseen test graphs $(G_{k+1},\ldots,G_{k+l})$.
For instance, inductive learning can be used to embed molecular graphs, each representing a molecule structure \citep{gilmer2017neural}, generalizing to new graphs and showing error margins within {chemical accuracy} on many {quantum properties}.  
Embedding dynamic or temporally evolving graphs is also another inductive graph embedding problem.

%
%
%

There is a strong connection between inductive graph embedding and \textit{node features}~(\cref{subsec:node_features}) as the latter are usually necessary for most inductive graph representation learning algorithms.
More concretely, node features can be leveraged to learn embeddings with parametric mappings and instead of directly optimizing the embeddings, one can optimize the mapping’s parameters.
The learned mapping can then be applied to any node (even those that were not present a training time).
On the other hand, when node features are not available, the first mapping from nodes to embeddings is usually a one-hot encoding which fails to generalize to new graphs where the canonical node ordering is not available. 

Finally, we note that this categorization of graph embedding methods is at best an incomplete lens for viewing the landscape.
While some models are inherently better suited to different tasks in practice, recent theoretical results \citep{Srinivasan2020On} show that models previously assumed to be capable of only one setting (e.g. only transductive) can be used in both.

\subsubsection{Positional vs structural network embedding}\label{subsec:pos_struct}
An emerging categorization of graph embedding algorithms is about whether the learned embeddings are positional or structural. 
Position-aware embeddings capture global relative positions of nodes in a graph and it is common to refer to embeddings as positional if they can be used to approximately reconstruct the edges in the graph, preserving distances such as shortest paths in the original graph~\citep{you2019position}. 
Examples of positional embedding algorithms include random walk or matrix factorization methods. 
On the other hand, structure-aware embeddings capture local structural information about nodes in a graph, i.e. nodes with similar node features or similar structural roles in a network should have similar embeddings, regardless of how far they are in the original graph. 
For instance, GNNs usually learn embeddings by incorporating information for each node's neighborhood, and the learned representations are thus structure-aware. 

In the past, positional embeddings have commonly been used for unsupervised tasks where positional information is valuable (e.g. link prediction or clustering) while structural embeddings have been used for supervised tasks (e.g. node classification or whole graph classification).
More recently, there has been attempts to bridge the gap between positional and structural representations, with positional GNNs~\citep{you2019position} and theoretical frameworks showing the equivalence between the two classes of embeddings~\citep{Srinivasan2020On}.

\subsubsection{Unsupervised and supervised network embedding}\label{subsec:supervision}
Network embedding can be \textit{unsupervised} in the sense that the only information available is the graph structure (and possibly node features) or \textit{supervised}, if additional information such as node or graph labels is provided.
In unsupervised network embedding, the goal is to learn embeddings that preserved the graph structure and this is usually achieved by optimizing some reconstruction loss, which measures how well the learned embeddings can approximate the original graph. 
In supervised network embedding, the goal is to learn embeddings for a specific purpose such as predicting node or graph attributes, and models are optimized for a specific task such as graph classification or node classification. 
We use the level of supervision to build our taxonomy and cover differences between supervised and unsupervised methods in more details in~\cref{sec:framework}. 

\section{A Taxonomy of Graph Embedding Models}\label{sec:framework}
We first describe our proposed framework, \name{}, a general framework for GRL~(\cref{subsec:enc-dec}).
In particular, \name{} is general enough that it can be used to succinctly describe over thirty GRL methods (both unsupervised and supervised). 
We use \name{} to introduce a comprehensive taxonomy in~\cref{subsec:optimization} and~\cref{subsec:taxonomy}, which summarizes exiting works with shared notations and simple block diagrams, making it easier to understand similarities and differences between GRL methods.

\subsection{The \name{} framework}\label{subsec:enc-dec}
\tikzstyle{module} = [rectangle, draw, fill=blue!20, node distance=0.8cm, text width=8em, text centered, rounded corners, minimum height=3em, thick]
\tikzstyle{data} = [rectangle, draw, fill=red!20, node distance=0.65cm, text width=2em, text centered, rounded corners, minimum height=3em, thick]
\tikzstyle{loss} = [rectangle, draw, node distance=0.65cm, text width=3em, text centered, minimum height=3em, thick]
\tikzstyle{c} = [rectangle, draw, inner sep=0.2cm, dashed]
\tikzstyle{c2} = [rectangle, inner sep=3cm]
\tikzstyle{l} = [draw, -latex',thick]
\tikzstyle{l2} = [draw, -latex',thick, dashed]
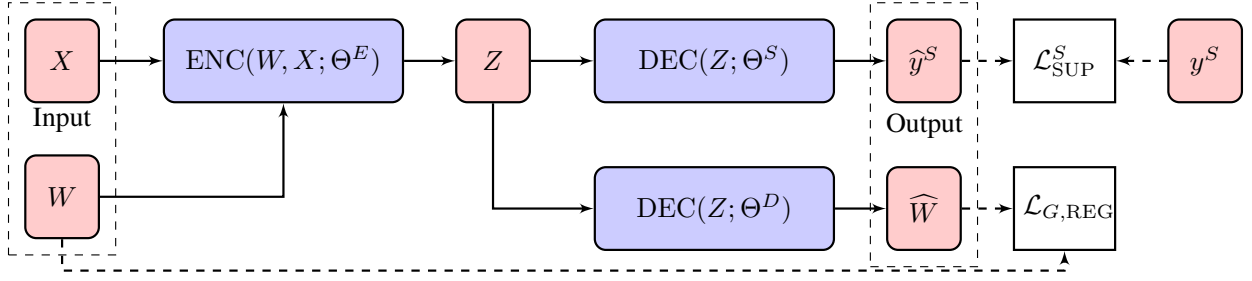
\begin{figure}[t]
\resizebox{\textwidth}{!}{\renewcommand{\arraystretch}{1.1}
    \centering\begin{tikzpicture}[auto]
    \node [data] (X) {$X$};
    \node [data, below=of X] (W) {$W$};
    \node [module, right=of X] (ENC) {$\mathrm{ENC}(W, X;\Theta^E)$};
    \node [data, right=of ENC] (Z) {$Z$};
    \node [module, right=of Z] (DEC1) {$\mathrm{DEC}(Z;\Theta^S)$};
    \node [module, below=of DEC1] (DEC2) {$\mathrm{DEC}(Z;\Theta^D)$};
    \node [data, right=of DEC1] (y_hat) {$\widehat{y}^S$};
    \node [data, right=of DEC2] (W_hat) {$\widehat{W}$};
    \node [c,fit=(X) (W)] (input) {Input};
    \node [c,fit=(W_hat) (y_hat)] (output) {Output};
    \node [loss, right=of y_hat] (L_sup) {$\mathcal{L}^S_{\mathrm{SUP}}$};
    \node [loss, right=of W_hat] (L_reg) {$\mathcal{L}_{G,\mathrm{REG}}$};
    \node [data, right=of L_sup] (y_sup) {$y^S$};

    \path [l] (X) -- (ENC);
    \path [l] (ENC) -- (Z);
    \path [l] (Z) -- (DEC1);

    \path [l] (W) -| (ENC);
    \path [l] (DEC1) -- (y_hat);
    \path [l] (DEC2) -- (W_hat);
    \path [l] (Z) |- (DEC2);
    \path [l2] (y_hat) -- (L_sup);
    \path [l2] (y_sup) -- (L_sup);
    \path [l2] (W_hat) -- (L_reg);
    \path [l2] (W.south) -- ($(W.south) - (0mm, 4mm)$)  -| (L_reg.south);


\end{tikzpicture}}
    \caption{Illustration of the \name{} framework. Based on the supervision available, methods will use some or all of the branches. In particular, unsupervised methods do not leverage label decoding for training and only optimize the similarity or dissimilarity decoder (lower branch). 
    On the other hand, semi-supervised and supervised methods leverage the additional supervision to learn models' parameters (upper branch).}\label{fig:enc-dec}
\end{figure}

The \name{} framework builds on top of the work of \citet{hamilton2017representation}, which describes unsupervised network embedding methods from an encoder-decoder perspective. 
\citet{cruz2019modular} also recently proposed a modular encoder-based framework to describe and compare unsupervised graph embedding methods.
Different from these unsupervised frameworks, we provide a more general framework which additionally encapsulates supervised graph embedding methods, including ones utilizing the graph as a regularizer~(e.g.~\cite{zhu2002learning}), and graph neural networks such as ones based on message passing \citep{gilmer2017neural, scarselli2009graph} or graph convolutions \citep{bruna2013spectral,kipf2016semi}. 

\paragraph{Input} 
The \name{} framework takes as input an undirected weighted graph $G =(V, E)$, with adjacency matrix $W\in\mathbb{R}^{|V|\times |V|}$, and optional node features $X\in\mathbb{R}^{|V|\times d_0}$.
In (semi-)supervised settings, we assume that we are given training target labels for nodes (denoted $N$), edges (denoted $E$), and/or for the entire graph (denoted $G$). 
We denote the supervision signal as $S  \in \{N, E, G\}$, as presented below.

\paragraph{Model} The \name{} framework can be decomposed as follows: 
\begin{itemize}
    \item \textbf{Graph encoder network} $\textrm{ENC}_{\Theta^{E}} : \mathbb{R}^{|V| \times |V|} \times \mathbb{R}^{|V| \times d_0} \rightarrow \mathbb{R}^{|V| \times d}$, parameterized by $\Theta^E$, which combines the graph structure with node features (or not) to produce node embedding matrix $Z \in \mathbb{R}^{|V|\times d}$ as:
    $$Z=\mathrm{ENC}(W, X;\Theta^E).$$
    As we shall see next, this node embedding matrix might capture different graph properties depending on the supervision used for training. 
    \item \textbf{Graph decoder network} $\textrm{DEC}_{\Theta^{D}} : \mathbb{R}^{|V| \times d}  \rightarrow \mathbb{R}^{|V| \times |V|}$, parameterized by $\Theta^D$, which uses the node embeddings $Z$ to compute similarity or dissimilarity scores for all node pairs, producing a matrix $\widehat{W}\in\mathbb{R}^{|V|\times|V|}$ as:
    $$\widehat{W}=\mathrm{DEC}(Z;\Theta^D).$$
    \item \textbf{Classification network} $ \textrm{DEC}_{\Theta^{S}} : \mathbb{R}^{|V| \times d}  \rightarrow \mathbb{R}^{|V| \times |\mathcal{Y}|}$, where $\mathcal{Y}$ is the label space. This network is used in (semi-)supervised settings and parameterized by $\Theta^S$.
    The output is a distribution over the labels $\hat{y}^S$, using node embeddings, as:
    $$\widehat{y}^S=\mathrm{DEC}(Z;\Theta^S).$$
\end{itemize}
Our \name{} framework is general (see~\cref{fig:enc-dec} for an illustration). 
Specific choices of the aforementioned (encoder and decoder) networks allows \name{} to realize specific graph embedding methods. 
Before presenting the taxonomy and showing realizations of various methods using our framework, we briefly discuss an application perspective.

\paragraph{Output}
The \name{} model can return a reconstructed graph similarity or dissimilarity matrix $\widehat{W}$ (often used to train \textit{unsupervised} embedding algorithms), as well as a output labels $\widehat{y}^S$ for \textit{supervised} applications.
The label output space $\mathcal{Y}$ varies depending on the supervised application.
\begin{itemize}
\item \textbf{Node-level supervision,} with $\widehat{y}^N \in \mathcal{Y}^{|V|}$, where $\mathcal{Y}$ represents the node label space. If $\mathcal{Y}$ is categorical, then this is also known as (semi-)supervised node classification ~(\cref{sec:nc}), in which case the label decoder network produces labels for each node in the graph. 
If the embedding dimensions $d$ is such that $d = |\mathcal{Y}|$, then the label decoder network can be just a simple softmax activation across the rows of $Z$, producing a distribution over labels for each node.
Additionally, the graph decoder network might also be used in supervised node-classification tasks, as it can be used to regularize embeddings (e.g. neighbor nodes should have nearby embeddings, regardless of node labels).

\item \textbf{Edge-level supervision,} with $\widehat{y}^E \in \mathcal{Y}^{|V| \times |V|}$, where $\mathcal{Y}$ represents the edge label space. 
For example, $\mathcal{Y}$ can be multinomial in knowledge graphs (for describing the types of relationships between two entities), setting $\mathcal{Y} = \{0, 1\}^\textrm{\#(relation types)}$. 
It is common to have $\textrm{\#(relation types)} = 1$, and this is is known as \textit{link prediction}, where edge relations are binary. 
In this review, when $\widehat{y}^E = \{0, 1\}^{|V| \times |V|}$ (i.e. $\mathcal{Y} = \{0, 1\}$), then rather than naming the output of the decoder as $\widehat{y}^E$, we instead follow the nomenclature and position link prediction as an \textit{unsupervised} task~(\cref{sec:unsup_models}).  Then in lieu of $\widehat{y}^E$ we utilize $\widehat{W}$, the output of the graph decoder network (which is learned to reconstruct a target similarity or dissimilarity matrix) to rank potential edges.

\item \textbf{Graph-level supervision,}
with $\widehat{y}^G \in \mathcal{Y}$, where $\mathcal{Y}$ is the graph label space.
In the graph classification task~(\cref{sec:gc}), the label decoder network converts node embeddings into a single graph labels, using \textit{graph pooling} via the graph edges captured by $W$.
More concretely, the graph pooling operation is similar to pooling in standard CNNs, where the goal is to downsample local feature representations to capture higher-level information. 
However, unlike images, graphs don't have a regular grid structure and it is hard to define a pooling pattern which could be applied to every node in the graph. 
A possible way of doing so is via graph coarsening, which groups similar nodes into clusters to produce smaller graphs \citep{defferrard2016convolutional}.
There exist other pooling methods on graphs such as DiffPool \citep{ying2018hierarchical} or SortPooling \citep{zhang2018end} which creates an ordering of nodes based on their structural roles in the graph. 
Details about graph pooling operators is outside the scope of this work and we refer the reader to recent surveys~\citep{wu2019comprehensive} for a more in-depth treatment. 
\end{itemize} 

\subsection{Taxonomy of objective functions}
\label{subsec:optimization}
We now focus our attention on the optimization of models that can be described in the \name{} framework by describing the loss functions used for training. 
Let $\Theta=\{\Theta^E,\Theta^D,\Theta^S\}$ denote all model parameters.
\name{} models can be optimized using a combination of the following loss terms: 
\begin{itemize}\itemsep -2pt
    \item \textbf{Supervised loss} term, $\mathcal{L}_\mathrm{SUP}^S$, which compares the predicted labels $\hat{y}^S$ to the ground truth labels $y^S$. 
    This term depends on the task the model is being trained for.
    For instance, in semi-supervised node classification tasks ($S=N$), the graph vertices are split into labelled and unlabelled nodes ($V=V_L\cup V_U$), and the supervised loss is computed for each labelled node in the graph:
    $$\mathcal{L}_\mathrm{SUP}^{N}(y^N, \hat{y}^N;\Theta) = \sum_{i|v_i\in V_L} \ell(y^N_i, \hat{y}^N_i;\Theta),$$
    where $\ell(\cdot)$ is the loss function used for classification (e.g. cross-entropy).
    Similarly for graph classification tasks ($S=G$), the supervised loss is computed at the graph-level and can be summed across multiple training graphs:
    $$\mathcal{L}_\mathrm{SUP}^{G}(y^G, \hat{y}^G;\Theta) = \ell(y^G, \hat{y}^G;\Theta).$$
    \item \textbf{Graph regularization loss} term, $\mathcal{L}_{G,\mathrm{REG}}$, which leverages the graph structure to impose regularization constraints on the model parameters. 
    This loss term acts as a smoothing term and measures the distance between the decoded similarity or dissimilarity matrix $\widehat{W}$, and a target similarity or dissimilarity matrix $s(W)$, which might capture higher-order proximities than the adjacency matrix itself: 
    \begin{align}
        \mathcal{L}_{G,\mathrm{REG}}(W, \widehat{W};\Theta)=d_1( s(W), \widehat{W}),\label{eq:reg_loss}
    \end{align}
    where $d_1(\cdot,\cdot)$ is a distance or dissimilarity function. 
    Examples for such regularization are constraining neighboring nodes to share similar embeddings, in terms of their distance in L2 norm. 
    We will cover more examples of regularization functions in~\cref{sec:unsup_models} and~\cref{sec:sup_models}.
    \item \textbf{Weight regularization loss} term, $\mathcal{L}_{\mathrm{REG}}$, e.g. for representing prior, on trainable model parameters for reducing overfitting.
    The most common regularization is L2 regularization (assumes a standard Gaussian prior):
    $$\mathcal{L}_{\mathrm{REG}}(\Theta)=\sum\limits_{\theta\in\Theta}||\theta||_2^2.$$
 \end{itemize}
Finally, models realizable by \name{} framework are trained by minimizing the total loss $\mathcal{L}$ defined as:
\begin{align}
\mathcal{L}=\alpha\mathcal{L}_\mathrm{SUP}^S(y^S, \hat{y}^S;\Theta) + \beta\mathcal{L}_{G,\mathrm{REG}}(W, \widehat{W};\Theta) + \gamma\mathcal{L}_{\mathrm{REG}}(\Theta),\label{eq:loss}
\end{align}
where $\alpha$, $\beta$ and $\gamma$ are hyper-parameters, that can be tuned or set to zero.
Note that graph embedding methods can be trained in a \textit{supervised} (${\alpha\neq 0}$) or \textit{unsupervised} (${\alpha=0}$) fashion. 
Supervised graph embedding approaches leverage an additional source of information to learn embeddings such as node or graph labels. 
On the other hand, unsupervised network embedding approaches rely on the graph structure only to learn node embeddings.

A common approach to solve supervised embedding problems is to first learn embeddings with an unsupervised method~(\cref{sec:unsup_models}) and then train a supervised model on the learned embeddings. 
However, as pointed by \citet{weston2012deep} and others, using a two-step learning algorithm might lead to sub-optimal performances for the supervised task, and in general, supervised methods~(\cref{sec:sup_models}) outperform two-step approaches.

\subsection{Taxonomy of encoders}\label{subsec:taxonomy}
Having introduced all the building blocks of the \name{} framework, we now introduce our graph embedding taxonomy. 
While most methods we describe next fall under the \name{} framework, they will significantly differ based on the encoder used to produce the node embeddings, and the loss function used to learn model parameters. 
We divide graph embedding models into four main categories:
\begin{itemize}
    \item \textbf{Shallow embedding methods}, where the encoder function is a simple embedding lookup. That is, the parameters of the model $\Theta^E$ are directly used as node embeddings:
    \begin{align*}
        Z&=\text{ENC}(\Theta^E)\\
        &=\Theta^E\in\mathbb{R}^{|V|\times d}.
    \end{align*}
    Note that shallow embedding methods rely on an embedding lookup and are therefore \textit{transductive}, i.e. they generally cannot be directly applied in \textit{inductive} settings where the graph structure is not fixed. 
    \item \textbf{Graph regularization methods}, where the encoder network ignores the graph structure and only uses node features as input:
    $$Z=\text{ENC}(X;\Theta^E).$$ 
    As its name suggests, graph regularization methods leverage the graph structure through the graph regularization loss term in~\cref{eq:loss} ($\beta\neq 0$) to regularize node embeddings. 
    \item \textbf{Graph auto-encoding methods}, where the encoder is a function of the graph structure only:
    $$Z=\text{ENC}(W;\Theta^E).$$
    \item \textbf{Neighborhood aggregation methods}, including graph convolutional methods, where both the node features and the graph structure are used in the encoder network.
    Neighborhood aggregation methods use the graph structure to propagate information across nodes and learn embeddings that encode structural properties about the graph:
    $$Z=\text{ENC}(W, X;\Theta^E).$$
\end{itemize}
\subsection{Historical Context}
In most machine learning applications, models follow a rather simple two-step paradigm. 
First they automatically extract meaningful patterns from data, without the need for manual feature engineering.
This is the so-called \emph{representation learning} step~\citep{bengio2013representation}. 
Second, they use these representations in downstream applications which can be \emph{supervised} (e.g.\ classification) or \emph{unsupervised} (e.g.\ clustering, visualization, nearest-neighbor search). 
This is the so-called \emph{downstream task}.\footnote{For supervised tasks, these two steps are often combined into one learning algorithm, which learns both representations and decision rules on top of these representations.}

A good data representation should be expressive enough that it preserves meaningful features found in the original data, but simple enough that it makes the downstream task easier.
For instance, having low-dimensional representations of high-dimensional datasets can help overcome issues caused by the curse of dimensionality such as overfitting.
In the context of GRL, a graph encoder is used to learn representation, and a graph or label decoder is used for downstream tasks (e.g.\ node classification, link prediction).  
Historically, the graph encoder-decoder networks were used for manifold learning.
When input data lies on a high-dimensional Euclidean space, it is common to assume it sits in an intrinsically low-dimensional manifold. This is known as the standard \emph{manifold hypothesis}. 
Methods for \emph{manifold learning} seek to recover this intrinsically low-dimensional manifold. 
This is usually done by first building a discrete approximation of the manifold using a graph in which nearby points in the ambient Euclidean space are connected by an edge. 
Because manifolds are locally Euclidean, graph distances provide a good proxy for both local and global manifold distances.
The second step is to ``flatten'' this graph representation by learning a non-linear mapping from nodes in the graph to points in a low-dimensional Euclidean space, while preserving graph distances as best as possible. 
These representations are usually easier to work with than the original high-dimensional representations, and can then be used in downstream applications. 

In the early 2000s, non-linear\footnote{The non-linearity term here comes to contrast with Euclidean dimensionality reduction methods (such as Principal Component Analysis) which rely on linear projections.} dimensionality reduction methods were extremely popular to solve the manifold learning problem. 
For instance, Laplacian Eigenmaps (LE)~\citep{belkin2002laplacian} use spectral techniques to compute embeddings, and IsoMap~\citep{tenenbaum2000global} use a combination of the Floyd–Warshall algorithm and the classical Multi-dimensional scaling algorithm to preserve global graph geodesics. 
These methods rely on shallow encoders, and we describe some of these in~\cref{subsec:laplacian_unsup_shallow}.

While manifold dimensionality reduction methods have had critical impact in machine learning applications, they do not scale to large datasets. 
For instance, IsoMAP requires computing all pairs of shortest paths which takes more than quadratic time. 
A perhaps more important limitation is their inability to compute embeddings for new datapoints because the mappings from node to embeddings are non-parametric. 

In more recent years, many non-shallow network architectures have been proposed for the problem of graph embedding. 
These include graph regularization networks and graph neural networks which can be described using our \name{} framework.
Because they leverage the expressiveness of deep neural networks, GRL models often yield more expressive, scalable and generalizable embeddings than classical methods. 

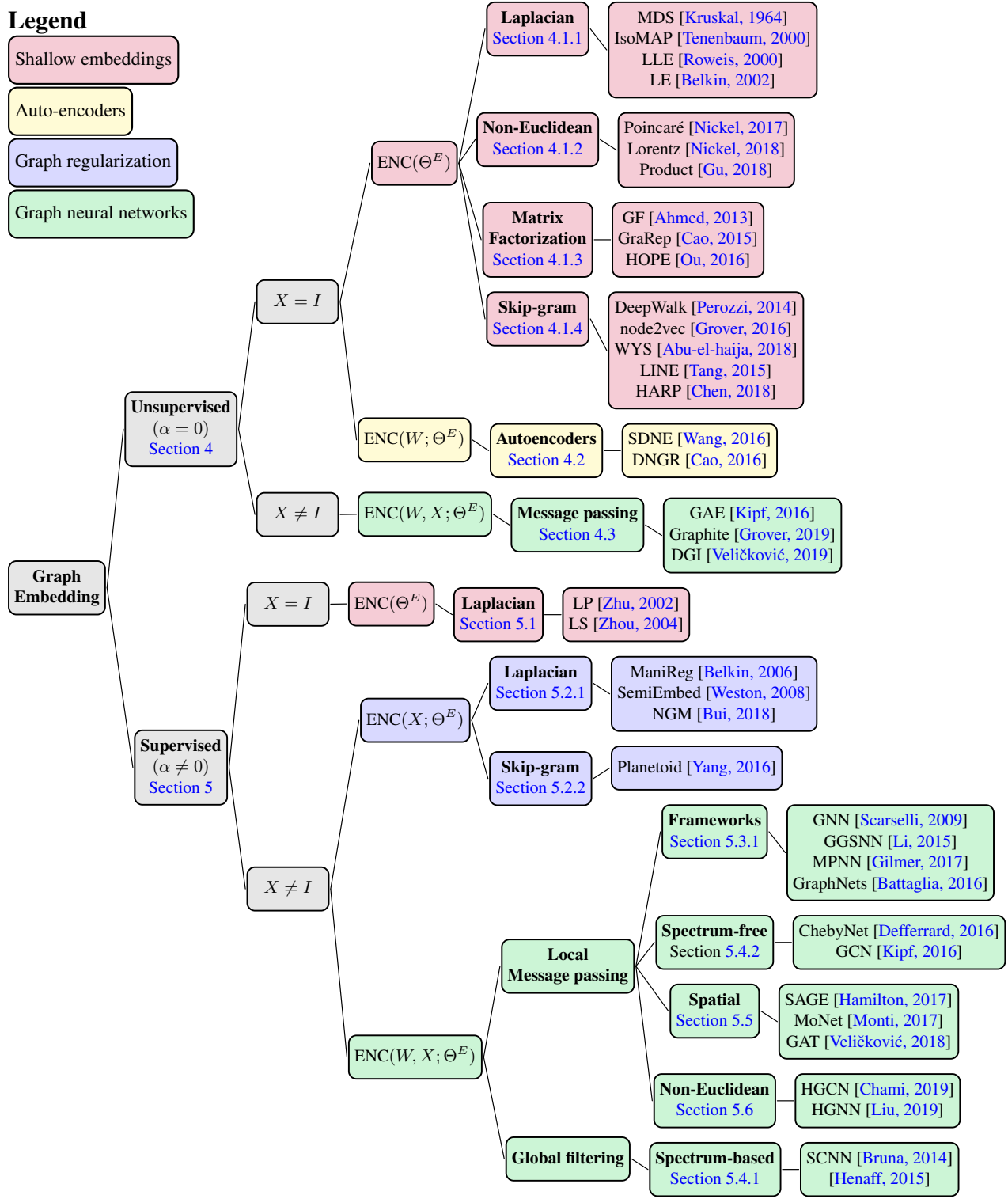
\begin{figure}
    \small
    \forestset{
  L1/.style={rectangle, draw, fill=blue!20!red!20, rounded corners, thick,edge={line width=0.5pt}},
  L2/.style={rectangle, draw, fill=blue!20!green!20, rounded corners, thick,edge={line width=0.5pt}},
  L3/.style={rectangle, draw, fill=gray!20, rounded corners, thick,edge={line width=0.5pt}},
  L4/.style={rectangle, draw, fill=blue!15, rounded corners, thick,edge={line width=0.5pt}},
  L5/.style={rectangle, draw, fill=yellow!20, rounded corners, thick,edge={line width=0.5pt}},
}
\hspace{2in}

%

\begin{minipage}[t]{\textwidth}
    \vspace{0.2cm}
    {\large \textbf{Legend}}
    \\
    \begin{forest}
        for tree={
            align=center, 
            grow=0,reversed, 
            parent anchor=east,child anchor=west, 
            edge={line cap=round},outer sep=+1pt, 
            rounded corners,minimum width=15mm,minimum height=8mm, 
            l sep=2mm 
        }
        [Shallow embeddings,L1]
    \end{forest}
    \\
    \begin{forest}
        for tree={
            align=center, 
            grow=0,reversed, 
            parent anchor=east,child anchor=west, 
            edge={line cap=round},outer sep=+1pt, 
            rounded corners,minimum width=15mm,minimum height=8mm, 
            l sep=5mm 
        }
        [Auto-encoders,L5]
    \end{forest}
    \\
    \begin{forest}
        for tree={
            align=center, 
            grow=0,reversed, 
            parent anchor=east,child anchor=west, 
            edge={line cap=round},outer sep=+1pt, 
            rounded corners,minimum width=15mm,minimum height=8mm, 
            l sep=5mm 
        }
        [Graph regularization,L4]
    \end{forest}
    \\  
    \begin{forest}
        for tree={
            align=center, 
            grow=0,reversed, 
            parent anchor=east,child anchor=west, 
            edge={line cap=round},outer sep=+1pt, 
            rounded corners,minimum width=15mm,minimum height=8mm, 
            l sep=5mm 
        }
        [Graph neural networks,L2]
    \end{forest}
    \\
    \vspace{-3in}
\end{minipage} 
\\
\resizebox{1.02\textwidth}{!}{%
\begin{forest}
    for tree={
        align=center, 
        grow=0,reversed, 
        parent anchor=east,child anchor=west, 
        edge={line cap=round},outer sep=+1pt, 
        rounded corners,minimum width=15mm,minimum height=8mm, 
        l sep=3mm 
    }
    [\textbf{Graph} \\
    \textbf{Embedding},L3
        [\textbf{Unsupervised}\\ $(\alpha \eqtext 0)$\\{\hyperref[sec:unsup_models]{\cref{sec:unsup_models}}},L3
            [$X \eqtext I$,L3
                [ENC$(\Theta^E)$,L1
                    [\textbf{Laplacian} \\  {\hyperref[subsec:laplacian_unsup_shallow]{\cref{subsec:laplacian_unsup_shallow}}},L1
                        [
                            MDS \citepalias{kruskal1964multidimensional} \\
                            IsoMAP \citepalias{tenenbaum2000global} \\
                            LLE \citepalias{roweis2000nonlinear} \\
                            LE \citepalias{belkin2002laplacian},L1
                        ]
                    ]
                    [\textbf{Non-Euclidean}\\ {\hyperref[subsec:non_euclidean_unsup]{\cref{subsec:non_euclidean_unsup}}},L1
                        [
                            Poincar\'e  \citepalias{nickel2017poincare} \\
                            Lorentz \citepalias{nickel2018learning} \\
                            Product \citepalias{gu2018learning},L1
                        ]
                    ]
                    [\textbf{Matrix}\\ \textbf{Factorization} \\ {\hyperref[subsec:matrix]{\cref{subsec:matrix}}},L1
                        [
                            GF \citepalias{ahmed2013distributed} \\
                            GraRep \citepalias{cao2015grarep}\\
                            HOPE \citepalias{ou2016asymmetric},L1
                        ]
                    ]
                    [\textbf{Skip-gram} \\ {\hyperref[subsec:skip_gram_unsup]{\cref{subsec:skip_gram_unsup}}},L1
                        [
                            DeepWalk \citepalias{perozzi2014deepwalk} \\
                            node2vec \citepalias{grover2016node2vec}\\
                            WYS \citepalias{abu2017watch} \\
                            LINE \citepalias{tang2015line} \\
                            HARP \citepalias{chen2017harp},L1
                        ]   
                    ]
                ]
                [ENC$(W; \Theta^E)$,L5
                    [\textbf{Autoencoders} \\ {\hyperref[subsec:auto_enc]{\cref{subsec:auto_enc}}},L5
                        [
                            SDNE \citepalias{wang2016structural} \\
                            DNGR \citepalias{cao2016deep},L5
                        ]
                    ]
                ]
            ]
            [$X \neq I$,L3
                [ENC$(W\comma X ; \Theta^E)$,L2
                    [\textbf{Message passing}  \\ {\hyperref[subsec:neighbour_unsup]{\cref{subsec:neighbour_unsup}}},L2
                        [GAE \citepalias{kipf2016variational}\\ 
                        Graphite \citepalias{grover2018graphite}\\ DGI~\citepalias{velickovic2018dgi},L2]
                    ]
                ]
            ]        
        ]
        [\textbf{Supervised} \\ $(\alpha \neq 0)$\\ {\hyperref[sec:sup_models]{\cref{sec:sup_models}}},L3
            [$X \eqtext I$,L3
                [ENC$(\Theta^E)$,L1
                    [\textbf{Laplacian}  \\ {\hyperref[subsec:shallow_sup]{\cref{subsec:shallow_sup}}},L1
                        [
                            LP \citepalias{zhu2002learning} \\
                            LS \citepalias{zhou2004learning},L1
                        ]
                    ]
                ]
            ]
            [$X \neq I$,L3
                [ENC$(X ; \Theta^E)$,L4
                    [\textbf{Laplacian}  \\ {\hyperref[subsubsec:laplacian_reg_sup]{\cref{subsubsec:laplacian_reg_sup}}},L4
                        [
                            ManiReg \citepalias{belkin2006manifold} \\
                            SemiEmbed \citepalias{weston2012deep}
                            \\ 
                            NGM \citepalias{bui2017neural},L4
                        ]
                    ]
                     [\textbf{Skip-gram}  \\ {\hyperref[subsubsec:skip_gram_sup]{\cref{subsubsec:skip_gram_sup}}},L4
                        [Planetoid \citepalias{yang2016revisiting},L4]
                    ]
                ]
                [ENC$(W\comma X ; \Theta^E)$,L2
                    [\textbf{Local}\\ \textbf{Message passing},L2
                        [\textbf{Frameworks}\\ {\hyperref[subsec:gnn_sup]{\cref{subsec:gnn_sup}}},L2
                            [
                                GNN \citepalias{scarselli2009graph}\\ GGSNN \citepalias{li2015gated} \\
                                MPNN \citepalias{gilmer2017neural}\\ GraphNets \citepalias{battaglia2016interaction},L2
                            ]
                        ]
                        [\textbf{Spectrum-free}\\
                        \cref{subsec:spectrum_free},L2
                            [
                                ChebyNet \citepalias{defferrard2016convolutional}\\ GCN \citepalias{kipf2016semi},L2
                            ]
                        ]
                        [\textbf{Spatial}  \\ {\hyperref[subsec:spatial]{\cref{subsec:spatial}}},L2
                            [
                                SAGE \citepalias{hamilton2017inductive} \\
                                MoNet \citepalias{monti2017geometric}\\ GAT \citepalias{velickovic2017graph},L2
                            ]
                        ]
                        [\textbf{Non-Euclidean} \\ {\hyperref[subsec:non_euclidean_sup]{\cref{subsec:non_euclidean_sup}}},L2
                            [
                            HGCN \citepalias{chami2019hyperbolic} \\
                            HGNN \citepalias{liu2019hyperbolic},L2
                            ]
                        ]
                    ]
                    [\textbf{Global filtering},L2
                        [\textbf{Spectrum-based}\\
                        {\hyperref[subsec:spectral]{\cref{subsec:spectrum_based}}},L2
                            [
                                SCNN \citepalias{bruna2013spectral}\\
                                \citepalias{henaff2015deep},L2
                            ]
                        ]
                    ]
                ]
            ]
        ]
    ]   
\end{forest}
}

    \normalsize
    \caption{Taxonomy of graph representation learning methods.
    Based on what information is used in the encoder network, we categorize graph embedding approaches into four categories: shallow embeddings, graph auto-encoders, graph-based regularization and graph neural networks.
    Note that message passing methods can also be viewed as spatial convolution, since messages are computed over local neighborhood in the graph domain. 
    Reciprocally, spatial convolutions can also be described using message passing frameworks. 
    }
\label{fig:taxonomy}
\end{figure}

In the next sections, we review recent methods for supervised and unsupervised graph embedding techniques using \name{} and summarize the proposed taxonomy in~\cref{fig:taxonomy}.

\section{Unsupervised Graph Embedding}\label{sec:unsup_models}
We now give an overview of recent unsupervised graph embedding approaches using the taxonomy described in the previous section.
These methods map a graph, its nodes, and/or its edges, onto a continuous vector space, without using task-specific labels for the graph or its nodes.
Some of these methods optimize an objective to learn an embedding that preserves the graph structure e.g. by learning to reconstruct some node-to-node similarity or dissimilarity matrix, such as the adjacency matrix.
Some of these methods apply a contrastive objective, e.g. contrasting close-by node-pairs versus distant node-pairs \citep{perozzi2014deepwalk}: nodes co-visited in short random walks should have a similarty score higher than distant ones; or contrasting real graphs versus fake ones \citep{velickovic2018dgi}: the mutual information between a graph and all of its nodes, should be higher in real graphs than in fake graphs.

\subsection{Shallow embedding methods}
\label{sec:unsupervisedshallow}
\begin{small}
\begin{figure}[t]
    \centering\begin{tikzpicture}[auto]
    \node [data, right=of ENC] (Z) {$Z$};
    \node [module, right=of Z] (DEC1) {$\mathrm{DEC}(Z;\Theta^D)$};
    \node [data, right=of DEC1] (W_hat) {$\widehat{W}$};
    
    \node [loss, right=of W_hat] (loss) {$\mathcal{L}_{G,\mathrm{REG}}$};
    \node [data, right=of loss] (W) {$W$};

    \path [l] (Z) -- (DEC1);
    \path [l] (DEC1) -- (W_hat);
    \path [l2] (W_hat) -- (loss);
    \path [l2] (W) -- (loss);

\end{tikzpicture}
    \caption{Shallow embedding methods. The encoder is a simple embedding look-up and the graph structure is only used in the loss function.}\label{fig:shallow}
\end{figure}
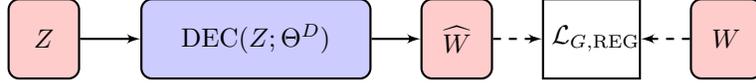
\end{small}

Shallow embedding methods are transductive graph embedding methods where the encoder function is a simple embedding lookup.
More concretely, each node $v_i\in V$ has a corresponding low-dimensional learnable embedding vector $Z_i\in\mathbb{R}^d$ and the shallow encoder function is simply:
\begin{align*}
    Z&=\mathrm{ENC}(\Theta^E)\\
  &=\Theta^E\in\mathbb{R}^{|V|\times d}.
\end{align*}
Embeddings of nodes can be learned such that the structure of the data in the embedding space corresponds to the underlying graph structure. 
At a high level, this is similar to dimensionality reduction methods such as PCA, except that the input data might not have a linear structure. 
In particular, methods used for non-linear dimensionality reduction often start by building a discrete graph from the data (to approximate the manifold) and can be applied to graph embedding problems.  
Here, we analyze two major types of shallow graph embedding methods, namely \textit{distance-based} and \textit{outer product-based} methods.

\paragraph{Distance-based methods} 
These methods optimize embeddings such that points that are close in the graph (as measured by their graph distances for instance) stay as close as possible in the embedding space using a predefined distance function. 
Formally, the decoder network computes pairwise distance for some distance function $d_2(\cdot,\cdot)$, which can lead to Euclidean~(\cref{subsec:laplacian_unsup_shallow}) or non-Euclidean~(\cref{subsec:non_euclidean_unsup}) embeddings:
\begin{align*}
    \widehat{W}&=\mathrm{DEC}(Z;\Theta^D)\\
            \textrm{with } \widehat{W}_{ij} &= d_2(Z_i, Z_j)
\end{align*}

\paragraph{Outer product-based methods} These methods on the other hand rely on pairwise dot-products to compute node similarities and the decoder network can be written as:
\begin{align*}
    \widehat{W}&=\mathrm{DEC}(Z;\Theta^D)\\
            &=ZZ^\top.
\end{align*}
Embeddings are then learned by minimizing the graph regularization loss: $\mathcal{L}_{G,\mathrm{REG}}(W, \widehat{W};\Theta)=d_1( s(W), \widehat{W})$.
Note that for distance-based methods, the function $s(\cdot)$ measures dissimilarity or distances between nodes (higher values mean less similar pairs of nodes), while in outer-product methods, it measures some notion of similarity in the graph (higher values mean more similar pairs). 

\subsubsection{Distance-based: Euclidean methods}\label{subsec:laplacian_unsup_shallow}
Most distance-based methods optimize Euclidean embeddings by minimizing Euclidean distances between similar nodes. 
Among these, we find linear embedding methods such as PCA or MDS, which learn low-dimensional linear projection subspaces, or nonlinear methods such as Laplacian eigenmaps, IsoMAP and Local linear embedding. 
Note that all these methods have originally been introduced for dimensionality reduction or visualization purposes, but can easily be extended to the context of graph embedding.  
\paragraph{Multi-Dimensional Scaling}
(MDS) \citep{kruskal1964multidimensional} refers to a set of embedding techniques used to map objects to positions while preserving the distances between these objects.
In particular, metric MDS (mMDS)~\cite{cox2008multidimensional} minimizes the 
regularization loss in~\cref{eq:reg_loss} with $s(W)$ set to some distance matrix measuring the dissimilarity between objects (e.g. Euclidean distance between points in a high-dimensional space): 
\begin{align*}
    d_1(s(W), \widehat{W})&=\bigg(\frac{\sum_{ij}(s(W)_{ij}-\widehat{W}_{ij})^2}{\sum_{ij}s(W)_{ij}^2}\bigg)^{1/2}\\
   \widehat{W}_{ij}&= d_2(Z_i, Z_j)=||Z_i - Z_j||_2.
\end{align*}
That is, mMDS finds an embedding configuration where distances in the low-dimensional embedding space are preserved by minimizing a residual sum of squares called the \textit{stress} cost function. 
Note that if the dissimilarities are computed from Euclidean distances of a higher-dimensional representation, then mMDS is equivalent to the PCA dimensionality reduction method.
Finally, there exist variants of this algorithm such as non-metric MDS, when the dissimilarity matrix $s(W)$ is not a distance matrix, or classical MDS (cMDS) which can be solved in closed form using a low-rank decomposition of the gram matrix. 
\paragraph{Isometric Mapping} (IsoMap) \citep{tenenbaum2000global} is an algorithm for non-linear dimensionality reduction which estimates the intrinsic geometry of a data lying on a manifold.
This method is similar to MDS, except for a different choice of the distance matrix.
IsoMap approximates manifold distances (in contrast with straight-line Euclidean geodesics) by first constructing a discrete neighborhood graph $G$, and then using the graph distances (length of shortest paths computed using Dijkstra's algorithm for example) to approximate the manifold geodesic distances: 
\begin{align*}
    s(W)_{ij}&=d_G(v_i, v_j).
\end{align*}
IsoMAP then uses the cMDS algorithm to compute representations that preserve these graph geodesic distances.
Different from cMDS, IsoMAP works for distances that do not necessarily come from a Euclidean metric space (e.g. data defined on a Riemannian manifold).
It is however computationally expensive due to the computation of all pairs of shortest path lengths in the neighborhood graph.
\paragraph{Locally Linear Embedding} (LLE) \citep{roweis2000nonlinear} is another non-linear dimensionality reduction technique which was introduced around the same time as IsoMap and improves over its computational complexity via sparse matrix operations. 
Different from IsoMAP which preserves the global geometry of manifolds via geodesics, LLE is based on the local geometry of manifolds and relies on the assumptions that when locally viewed, manifolds are approximately linear. 
The main idea behind LLE is to approximate each point using a linear combination of embeddings in its local neighborhood (linear patches). 
These local neighborhoods are then compared globally to find the best non-linear embedding. 
\paragraph{Laplacian Eigenmaps} (LE)~\citep{belkin2002laplacian} is a non-linear dimensionality reduction methods that seeks to preserve \textit{local} distances. 
Spectral properties of the graph Laplacian matrix capture important structural information about graphs. 
In particular, eigenvectors of the graph Laplacian provide a basis for smooth functions defined on the graph vertices (the ``smoothest'' function being the constant eigenvector corresponding to eigenvalue zero). 
LE is a non-linear dimensionality reduction technique which builds on this intuition. 
LE first constructs a graph from datapoints (e.g. k-NN graph or $\varepsilon$-neighborhood graph) and then represents nodes in the graphs via the Laplacian's eigenvectors corresponding to smaller eigenvalues.
The high-level intuition for LE is that points that are close on the manifold (or graph) will have similar representations, due to the ``smoothness'' of Laplacian's eigenvectors with small eigenvalues. 
Formally, LE learns embeddings by solving the generalized eigenvector problem:
\begin{align*}
    \underset{Z\in \mathbb{R}^{|V|\times d}}{\mathrm{min}}\ &Z^\top L\ Z\\
    \text{subject to }&Z^\top DZ=I\ \ \text{and}\ \ Z^\top D1=0, 
\end{align*}
where the first constraint removes an arbitrary scaling factor in the embedding and the second one removes trivial solutions corresponding to the constant eigenvector (with eigenvalue zero for connected graphs).
Further, note that $Z^\top L\ Z=\frac{1}{2}\sum_{ij}W_{ij}||Z_i-Z_j||_2^2$ and therefore the minimization objective can be equivalently written as a graph regularization term using our notations:
\begin{align*}
    d_1(W, \widehat{W})&=\sum_{ij}W_{ij}\widehat{W}_{ij}\\
    \widehat{W}_{ij}&=d_2(Z_i, Z_j)=||Z_i-Z_j||_2^2.
\end{align*}
Therefore, LE learns embeddings such that the Euclidean distance in the embedding space is small for points that are close on the manifold.

\subsubsection{Distance-based: Non-Euclidean methods}\label{subsec:non_euclidean_unsup}
The distance-based methods described so far assumed embeddings are learned in a Euclidean space.
Graphs are non-Euclidean discrete data structures, and several works proposed to learn graph embeddings into non-Euclidean spaces instead of conventional Euclidean space.
Examples of such spaces include the hyperbolic space, which has a non-Euclidean geometry with a constant negative curvature and is well-suited to represent hierarchical data. 

To give more intuition, the hyperbolic space can be thought of as continuous versions of trees, where geodesics (generalization of shortest paths on manifolds) resemble shortest paths in discrete trees. 
Further, the volume of balls grows exponentially with radius in hyperbolic space, similar to trees where the number of nodes within some distance to the root grows exponentially. 
In contrast, this volume growth is only polynomial in Euclidean space and therefore, the hyperbolic space has more ``room'' to fit complex hierarchies and compress representations.
In particular, hyperbolic embeddings can embed trees with arbitrary low distortion in just two-dimensions \citep{sarkar2011low} whereas this is not possible in Euclidean space. 
This makes hyperbolic space a natural candidate to embed tree-like data and more generally, hyperbolic geometry offers an exciting alternative to Euclidean geometry for graphs that exhibit hierarchical structures, as it enables embeddings with much smaller distortion.

Before its use in machine learning applications, hyperbolic geometry has been extensively studied and used in network science research.
\citet{kleinberg2007geographic} proposed a greedy algorithm for geometric rooting, which maps nodes in sensor networks to coordinates on a hyperbolic plane via spanning trees, and then performs greedy geographic routing. 
Hyperbolic geometry has also been used to study the structural properties of complex networks (networks with non-trivial topological features used to model real-world systems).
\citet{krioukov2010hyperbolic} develop a geometric framework to construct scale-free networks (a family of complex networks with power-law degree distributions), and conversely show that any scale-free graph with some metric structure has an underlying hyperbolic geometry.
\citet{papadopoulos2012popularity} introduce the Popularity-Similarity (PS) framework to model the evolution and growth of complex networks.
In this model, new nodes are likely to be connected to popular nodes (modelled by their radial coordinates in hyperbolic space) as well as similar nodes (modelled by the angular coordinates).  
This framework has further been used to map nodes in graphs to hyperbolic coordinates, by maximising the likelihood that the network is produced by the PS model~\citep{papadopoulos2014network}. 
Further works extend non-linear dimensionality reduction techniques such as LLE \citep{belkin2002laplacian} to efficiently map graphs to hyperbolic coordinates~\citep{alanis2016efficient,muscoloni2017machine}.

More recently, there has been interest in learning hyperbolic representations of hierarchical graphs or trees, via gradient-based optimization. We review some of these machine learning-based algorithms next.
\paragraph{Poincar\'e embeddings} 
\citet{nickel2017poincare} learn embeddings of hierarchical graphs such as lexical databases (e.g. WordNet) in the Poincar\'e model hyperbolic space. 
Using our notations, this approach learns hyperbolic embeddings via the Poincar\'e distance function: 
\begin{align*}
  d_2(Z_i, Z_j)&=d_{\mathrm{Poincar\acute{e}}}(Z_i, Z_j)\\
&=\mathrm{arcosh}\bigg(1+2\frac{||Z_i-Z_j||_2^2}{(1-||Z_i||_2^2)(1-||Z_j||_2^2)}\bigg).  
\end{align*}
Embeddings are then learned by minimizing distances between connected nodes while maximizing distances between disconnected nodes:
\begin{align*}
   d_1(W, \widehat{W})=-\sum_{ij}W_{ij}\mathrm{log}\ \frac{e^{-\widehat{W}_{ij}}}{\sum_{k|W_{ik}=0}e^{-\widehat{W}_{ik}}}=-\sum_{ij}W_{ij}\mathrm{log}\ \mathrm{Softmax}_{k|W_{ik}=0} ( - \widehat{W}_{ij}),
\end{align*}
where the denominator is approximated using negative sampling. 
Note that since the hyperbolic space has a manifold structure, embeddings need to be optimized using Riemannian optimization techniques \citep{bonnabel2013stochastic} to ensure that they remain on the manifold. 

Other variants of these methods have been proposed. 
In particular, \citet{nickel2018learning} explore a different model of hyperbolic space, namely the Lorentz model (also known as the hyperboloid model), and show that it provides better numerical stability than the Poincar\'e model. 
Another line of work extends non-Euclidean embeddings to mixed-curvature product spaces \citep{gu2018learning}, which provide more flexibility for other types of graphs (e.g. ring of trees).  
Finally, \citet{chamberlain2017neural} extend Poincar\'e embeddings to incorporate skip-gram losses using hyperbolic inner products.  

\subsubsection{Outer product-based: Matrix factorization methods}\label{subsec:matrix}
Matrix factorization approaches learn embeddings that lead to a low rank representation of some similarity matrix $s(W)$, where $s : \mathbb{R}^{|V| \times |V|} \rightarrow \mathbb{R}^{|V| \times |V|}$ is a transformation of the weighted adjacency matrix, and many methods set it to the identity, i.e. $s(W)=W$.
Other transformations include the Laplacian matrix or more complex similarities derived from proximity measures such as the Katz Index, Common Neighbours or Adamic Adar. 
The decoder function in matrix factorization methods is a simple outer product:
\begin{equation}
    \widehat{W}=\mathrm{DEC}(Z;\Theta^D)=ZZ^\top.\label{eq:skip_decoder}
\end{equation}
Matrix factorization methods learn embeddings by minimizing the regularization loss in~\cref{eq:reg_loss} with:
\begin{equation}
    \mathcal{L}_{G,\mathrm{REG}}(W, \widehat{W};\Theta)=||s(W)-\widehat{W}||_F^2.\label{eq:matrix_loss}
\end{equation}
That is, $d_1(\cdot,\cdot)$ in~\cref{eq:reg_loss} is the Frobenius norm between the reconstructed matrix and the target similarity matrix. 
By minimizing the regularization loss, graph factorization methods learn low-rank representations that preserve structural information as defined by the similarity matrix $s(W)$. We now review important matrix factorization methods.

\paragraph{Graph factorization} 
(GF) \citep{ahmed2013distributed} learns a low-rank factorization for the adjacency matrix by minimizing graph regularization loss in~\cref{eq:reg_loss} using:
\begin{align*}
    d_1(W, \widehat{W})=\sum_{(v_i,v_j)\in E}(W_{ij}-\widehat{W}_{ij})^2.
\end{align*}
Recall that $A$ is binary adjacency matrix, with $A_{ij}=1$ \textit{iif} $(v_i,v_j)\in E$. We can express the graph regularization loss in terms of Frobenius norm:
\begin{align*}
\mathcal{L}_{G,\mathrm{REG}}(W, \widehat{W};\Theta)=||A\cdot (W-\widehat{W})||_F^2,
\end{align*}
where $\cdot$ is the element-wise matrix multiplication operator.
Therefore, GF also learns a low-rank factorization of the adjacency matrix $W$ measured in Frobenuis norm.
Note that the sum is only over existing edges in the graph, which reduces the computational complexity of this method from $O(|V|^2)$ to $O(|E|)$.

\paragraph{Graph representation with global structure information} (GraRep) \citep{cao2015grarep} 
The methods described so far are all symmetric, that is, the similarity score between two nodes $(v_i, v_j)$ is the same a the score of $(v_j, v_i)$.
This might be a limiting assumption when working with directed graphs as some nodes can be strongly connected in one direction and disconnected in the other direction. 
GraRep overcomes this limitation by learning two embeddings per node, a source embedding $Z^s$ and a target embedding $Z^t$, which capture asymmetric proximity in directed networks. 
GraRep learns embeddings that preserve $k$-hop neighborhoods via powers of the adjacency and minimizes the graph regularization loss with:
\begin{align*}
    \widehat{W}^{(k)}&={Z^{(k),s}}{Z^{(k),t}}^\top\\
    \mathcal{L}_{G,\mathrm{REG}}(W, \widehat{W}^{(k)};\Theta)&=||D^{-k}W^k-\widehat{W}^{(k)}||_F^2,
\end{align*}
for each $1\le k\le K$.
GraRep concatenates all representations to get source embeddings $Z^s=[Z^{(1),s}|\ldots|Z^{(K),s}]$ and target embeddings $Z^t=[Z^{(1),t}|\ldots|Z^{(K),t}]$.
Finally, note that GraRep is not very scalable as the powers of $D^{-1}W$ might be dense matrices.

\paragraph{HOPE} \citep{ou2016asymmetric} 
Similar to GraRep, HOPE learns asymmetric embeddings but uses a different similarity measure. 
The distance function in HOPE is simply the Frobenius norm and the similarity matrix is a high-order proximity matrix (e.g. Adamic-Adar):
\begin{align*}
    \widehat{W}&={Z^s}{Z^t}^\top\\
    \mathcal{L}_{G,\mathrm{REG}}(W, \widehat{W};\Theta)&=||s(W)-\widehat{W}||_F^2.
\end{align*}
The similarity matrix in HOPE is computed with sparse matrices, making this method more efficient and scalable than GraRep.
\begin{figure*}
\centering
\includegraphics[width=\textwidth]{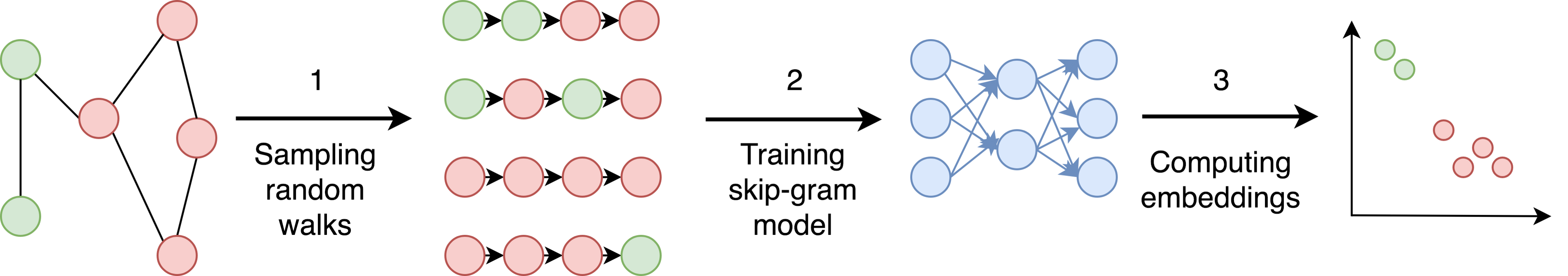}
\caption{An overview of the pipeline for random-walk graph embedding methods. Reprinted with permission from \citep{godec_2018}.}
\label{fig:walk}
\end{figure*}
\subsubsection{Outer product-based: Skip-gram methods}\label{subsec:skip_gram_unsup}
Skip-gram graph embedding models were inspired by efficient NLP methods modeling probability distributions over words for learning word embeddings \citep{mikolov2013distributed,pennington2014glove}.
Skip-gram word embeddings are optimized to predict context words, or surrounding words, for each target word in a sentence.
Given a sequence of words $(w_1,\ldots,w_T)$, skip-gram will minimize the objective:
$$\mathcal{L}=-\sum_{-K\le i\le K,i\neq 0}\mathrm{log}\ \mathbb{P}(w_{k-i}|w_k),$$
for each target words $w_k$.
In practice, the conditional probabilities can be estimated using neural networks, and skip-gram methods can be trained efficiently using negative sampling. 

\citet{perozzi2014deepwalk} empirically show the frequency statistics induced by random walks also follow Zipf's law, thus motivating the development of skip-gram graph embedding methods.
These methods exploit random walks on graphs and produce node sequences that are similar in positional distribution, as to words in sentences.
In skip-gram graph embedding methods, the decoder function is also an outer product~(\cref{eq:skip_decoder}) and the graph regularization term is computed over random walks on the graph.

\paragraph{DeepWalk} \citep{perozzi2014deepwalk} was the first attempt to generalize skip-gram models to graph-structured data.
DeepWalk draws analogies between graphs and language. Specifically, writing a sentence is analogous to performing a random walk, where the sequence of nodes visited during the walk, is treated as the words of the sentence.  
DeepWalk trains neural networks by maximizing the probability of predicting context nodes for each target node in a graph, namely nodes that are close to the target node in terms of hops and graph proximity.
For this purpose, node embeddings are decoded into probability distributions over nodes using row-normalization of the decoded matrix with softmax. 

To train embeddings, DeepWalk generates sequences of nodes using truncated unbiased random walks on the graph---which can be compared to sentences in natural language models---and then maximize their log-likelihood.
Each random walk starts with a node $v_{i_1} \in V$ and repeatedly sample next node at uniform: $v_{i_{j+1}} \in \{ v \in V \mid (v_{i_{j}}, v) \in E \}$. The walk length is a hyperparameter. All generated random-walk can then be passed to an NLP-embedding algorithm e.g. word2vec's Skipgram model. This two-step paradigm introduced by \citet{perozzi2014deepwalk} is followed by many subsequent works, such as node2vec \citep{grover2016node2vec}.

We note that is common for underlying implementations to use two distinct representations for each node (one for when a node is center of a truncated random walk, and one when it is in the context).  The implications of this modeling choice is studied further in \citep{haija2017lowrank}.

\citet{abu2017watch} show that training DeepWalk, in expectation, is equivalent to first sampling integer $q \sim [1, 2, \dots, T_{\max}]$ with mass $ \propto [1,  \frac{T_{\max} - 1}{T_{\max}}, \dots, \frac{1}{T_{\max}} ]$. Specifically, if $s(W) = \mathbb{E}_{q} \left[  \left(D^{-1}W\right)^q \right]$, then training DeepWalk is equivalent to minimizing:
\begin{equation}
    \label{eq:deepwalk}
    \mathcal{L}_{G,\mathrm{REG}}(W, \widehat{W};\Theta)=\mathrm{log}\ C - \sum_{v_i\in V,v_j\in V}s(W)_{ij}  \widehat{W}_{ij},
\end{equation}
where $C=\prod_i\sum_j\mathrm{exp}(\widehat{W}_{ij})$ is a normalizing constant.
Note that computing $C$ requires summing over all nodes in the graph which is computationally expensive. 
DeepWalk overcomes this issue by using a technique called hierarchical softmax, which computes $C$ efficiently using binary trees.
Finally, note that by computing truncated random walks on the graph, DeepWalk embeddings capture high-order node proximity.
\scriptsize
\begin{table}[!t]
\label{tab:unsup_shallow}
\centering
\resizebox{\textwidth}{!}{%
\begin{tabular}{c|c| c| c |c| c| c }
\cline{2-7}
 & \bfseries \multirow{ 2}{*}{Method} & \bfseries \multirow{ 2}{*}{Model} & \multirow{ 2}{*}{$s(W)_{ij}$}  & \multirow{ 2}{*}{$\mathrm{DEC}(Z;\Theta^D)_{ij}$}  & \multirow{ 2}{*}{$d_1(W \leftarrow s(W),\widehat{W} \leftarrow \mathrm{DEC}(Z;\Theta^D))$}  &  \textbf{order of}\\ 
 & & & & & & \textbf{proximity} \\
\hline
\multirow{ 2}{*}{\textbf{Distance}} & 
\multirow{ 2}{*}{\parbox{1.2cm}{Euclidean}} &  mMDS 
& distance matrix & $||Z_i-Z_j||_2$ & $ \bigg(\frac{\sum_{ij}(W_{ij}-\widehat{W}_{ij})^2}{\sum_{ij}W_{ij}^2}\bigg)^{1/2}$ & high \\
\multirow{2}{*}{$\equalto{\mathrm{DEC}(Z;\Theta^D)_{ij}}{d_2(Z_i, Z_j)}$} 
&& LE 
& $W_{ij}$ & $\ ||Z_i-Z_j||_2^2$ & $\sum_{ij}W_{ij}\widehat{W}_{ij}$ & $1^\textrm{st}$  \\
\cline{2-7}
& \multirow{ 2}{*}{Non-Euclidean} & \multirow{ 2}{*}{Poincar\'e} 
& \multirow{ 2}{*}{$W_{ij}$} & \multirow{ 2}{*}{$d_{\mathrm{Poincar\acute{e}}}(Z_i, Z_j)$}  & \multirow{ 2}{*}{$-\sum_{ij}W_{ij}\mathrm{log}\ \mathrm{Softmax}_{k|W_{ik}=0} ( - \widehat{W}_{ij})$} &\multirow{ 2}{*} {$1^\textrm{st}$ }\\ 
& & & & & & \\
\hline
 & \multirow{2}{*}{Matrix} & GF 
 & $W_{ij}$ & $Z_i^\top Z_j$ &$\sum_{ij|W_{ij}>0}(W_{ij}-\widehat{W}_{ij})^2$ & $1^\textrm{st}$ \\
\multirow{2}{*}{\textbf{Outer product}} & \multirow{ 2}{*}{{Factorization}} &GraRep 
&  $(D^{-k}W^k)_{ij}$ & ${Z_i^{(k),s}}^\top {Z_j^{(k),t}}$ & $||W-\widehat{W}||_F^2$ & $k^\textrm{th}$ \\
&& HOPE 
& $s(W)_{ij}$ & ${Z_i^s}^\top Z_j^\top $ & $||W-\widehat{W}||_F^2$ & high  \\
\cline{2-7}
 \multirow{2}{*}{$\equalto{\mathrm{DEC}(Z;\Theta^D)_{ij}}{Z_i^\top Z_j}$} &\multirow{3}{*}{{Skip-gram}} &
 DeepWalk 
 & $\propto\mathbb{E}_{q} \left[  \left(D^{-1}W\right)^q \right]_{ij}$ & $Z_i^\top  Z_j$  &$-\sum_{ij}W_{ij}\log \mathrm{Softmax}_j(\widehat{W}_{ij})$ & high  \\
 && node2vec 
 & $ \textrm{n2vWalk}(W;p, q)_{ij}$ & $Z_i^\top  Z_j$ & $-\sum_{ij}W_{ij}\log \mathrm{Softmax}_j(\widehat{W}_{ij})$ & high  \\
&& WYS 
& $\propto \mathbb{E}_{q} \left[  \left(D^{-1}W\right)^q \right]_{ij}$  & $Z_i^\top  Z_j$   & $\text{BCE}(W, \widehat{W}) $ & high  \\
\hline
\end{tabular}}
\caption{An overview of unsupervised shallow embedding methods, where the encoding function is a simple embedding look-up $Z=\mathrm{ENC}(\Theta^E)$. Softmax represents sampled/hierarchical softmax; $\propto$ for approximating random walks; n2vWalk is a traversal algorithm with (back) teleportation (approximates combination of BFS \& DFS). BCE is the sigmoid cross entropy loss for binary classification. }
\end{table}
\normalsize
\paragraph{node2vec} \citep{grover2016node2vec} is a random-walk based approach for unsupervised network embedding, that extends DeepWalk's sampling strategy. 
The authors introduce a technique to generate biased random walks on the graph, by combining graph exploration through breadth first search (BFS) and through depth first search (DFS). 
Intuitively, node2vec also preserves high order proximities in the graph but the balance between BFS and DFS allows node2vec embeddings to capture local structures in the graph, as well as global community structures, which can lead to more informative embeddings. 
Finally, note that negative sampling \citep{mikolov2013distributed} is used to approximate the normalization factor $C$ in~\cref{eq:deepwalk}. 

\paragraph{Watch Your Step} (WYS) \citep{abu2017watch} Random walk methods are very sensitive to the sampling strategy used to generate random walks.
For instance, some graphs may require shorter walks if local information is more informative that global graph structure, while in other graphs, global structure might be more important.
Both DeepWalk and node2vec sampling strategies use hyper-parameters to control this, such as the length of the walk or ratio between breadth and depth exploration.
Optimizing over these hyper-parameters through grid search can be computationally expensive and can lead to sub-optimal embeddings.
WYS learns such random walk hyper-parameters to minimize the overall objective (in analogy: each graph gets to choose its own preferred ``context size'', such that the probability of predicting random walks is maximized).
WYS shows that, when viewed in expectation,  these hyperparameters only correspond in the objective to coefficients to the powers of the adjacency matrix $(W^k)_{1\le k\le K}$. These coefficients are denoted
$q=(q_k)_{1\le k\le K}$ and are learned through back-propagation.
Should $q$'s learn a left-skewed distribution, then the embedding would prioritize local information and right-skewed distribution will enhance high-order relationships and graph global structure.  This concept has been extended to other forms of attention to the `graph context', such using a personalized context distributions for each node \citep{huang2020personalized}.

\paragraph{Large scale Information Network Embedding} (LINE) \citep{tang2015line} learns embeddings that preserve first and second order proximity.
To learn first order proximity preserving embeddings, LINE minimizes the graph regularization loss:
\begin{align*}
    \widehat{W}^{(1)}_{ij}&={Z_i^{(1)}}^\top Z_j^{(1)}\\
    \mathcal{L}_{G,\mathrm{REG}}(W, \widehat{W}^{(1)};\Theta)&=-\sum_{(i,j)|(v_i,v_j)\in E}W_{ij}\mathrm{log}\ \sigma(\widehat{W}^{(1)}_{ij}).
\end{align*}
LINE also assumes that nodes with multiple edges in common should have similar embeddings and learns second-order proximity preserving embeddings by minimizing:
\begin{align*}
    \widehat{W}^{(2)}_{ij}&={Z_i^{(2)}}^\top Z_j^{(2)}\\
    \mathcal{L}_{G,\mathrm{REG}}(W, \widehat{W}^{(2)};\Theta)&=-\sum_{(i,j)|(v_i,v_j)\in E}W_{ij}\mathrm{log}\ \frac{\mathrm{exp}(\widehat{W}^{(2)}_{ij})}{\sum_{k}\mathrm{exp}(\widehat{W}^{(2)}_{ik})}.
\end{align*}
Intuitively, LINE with second-order proximity decodes embeddings into context conditional distributions for each node $p_2(\cdot|v_i)$.
Note that optimizing the second-order objective is computationally expensive as it requires a sum over the entire set of edges. 
LINE uses negative sampling to sample negative edges according to some noisy distribution over edges.
Finally, as in GraRep, LINE combines first and second order embeddings with concatenation $Z=[Z^{(1)}|Z^{(1)}]$.

\paragraph{Hierarchical representation learning for networks} (HARP)  \citep{chen2017harp}
Both node2vec and DeepWalk learn node embeddings by minimizing non-convex functions, which can lead to local minimas.
HARP introduces a strategy that computes initial embeddings, leading to more stable training and convergence.
More precisely, HARP hierarchically reduces the number of nodes in the graph via graph coarsening.
Nodes are iteratively grouped into super nodes that form a graph with similar properties as the original graph, leading to multiple graphs with decreasing size $(G_1,\ldots,G_T)$.
Node embeddings are then learned for each coarsened graph using existing methods such as LINE or DeepWalk, and at time-step $t$, embeddings learned for $G_t$ are used as initialized embedding for the random walk algorithm on $G_{t-1}$. 
This process is repeated until each node is embedded in the original graph. The authors show that this hierarchical embedding strategy produces stable embeddings that capture macroscopic graph information.
%

\paragraph{Splitter} \citep{epasto2019splitter}
What if a node is not the correct `base unit' of analysis for a graph?  Unlike HARP, which coarsens a graph to preserve high-level topological features, Splitter is a graph embedding approach designed to better model nodes which have membership in multiple communities.
It uses the Persona decomposition \citep{epasto2017persona}, to create a derived graph, $G_P$ which may have multiple \emph{persona} nodes for each original node in $G$ (the edges of each original node are divided among its personas). $G_P$ can then be embedded (with some constraints) using any of the embedding methods discussed so far. The resulting representations allow persona nodes to be separated in the embedding space, and the authors show benefits to this on link prediction tasks.

\paragraph{Matrix view of Skip-gram methods}
As noted by \cite{levy2014neural}, Skip-gram methods can be viewed as matrix factorization, and the methods discussed here are related to those of Matrix Factorization~(\cref{subsec:matrix}).
This relationship is discussed in depth by \cite{qiu2018network}, who propose a general matrix factorization framework, NetMF, which uses the same underlying graph proximity information as DeepWalk, LINE, and node2vec.
Casting the node embedding problem as matrix factorization can offer benefits like easier algorithmic analysis (e.g., convergence guarantees to unique globally-optimal points),
and dense matrix undergoing decomposition can be sampled entry-wise \citep{qiu2019NetSMF}.

\begin{small}
\begin{figure}[t]
    \centering
    \resizebox{\textwidth}{!}{\renewcommand{\arraystretch}{1.1}
    \begin{tikzpicture}[auto]
    \node [data] (W) {$W$};
    \node [module, right=of W] (ENC) {$\mathrm{ENC}(W;\Theta^E)$};
    \node [data, right=of ENC] (Z) {$Z$};
    \node [module, right=of Z] (DEC1) {$\mathrm{DEC}(Z;\Theta^D)$};
    \node [data, right=of DEC1] (W_hat) {$\widehat{W}$};
    
    \node [loss, right=of W_hat] (loss) {$\mathcal{L}_{G,\mathrm{REG}}$};

    \path [l] (X) -- (ENC);
    \path [l] (ENC) -- (Z);
    \path [l] (Z) -- (DEC1);
    \path [l] (DEC1) -- (W_hat);
    \path [l2] (W_hat) -- (loss);
    \path [l2] (W.south) -- ($(W.south) - (0mm, 4mm)$)  -| (loss.south);


\end{tikzpicture}}
    \caption{Auto-encoder methods. The graph structure (stored as the graph adjacency matrix) is encoded and reconstructed using encoder-decoder networks. Models are trained by optimizing the graph regularization loss computed on the reconstructed adjacency matrix.}\label{fig:autoenc}
\end{figure}
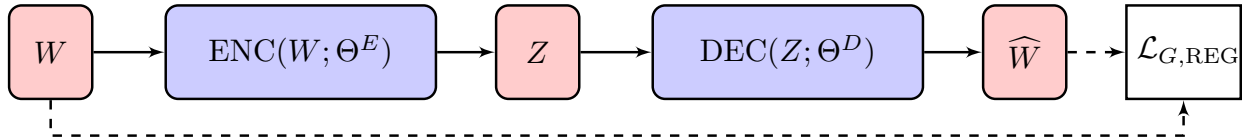
\end{small}

\subsection{Auto-encoders}\label{subsec:auto_enc}
Shallow embedding methods hardly capture non-linear complex structures that might arise in graphs.
Graph auto-encoders were originally introduced to overcome this issue by using deep neural network encoder and decoder functions, due to their ability model non-linearities. 
Instead of exploiting the graph structure through the graph regularization term, auto-encoders directly incorporate the graph adjacency matrix in the encoder function.
Auto-encoders generally have an encoding and decoding network which are multiple layers of non-linear layers. 
For graph auto-encoders, the encoder function has the form:  
$$Z=\mathrm{ENC}(W;\Theta^E).$$
That is, the encoder is a function of the adjacency matrix $W$ only.
These models are trained by minimizing a reconstruction error objective and we review examples of such objectives next.

\begin{figure*}
\centering
\includegraphics[width=0.5\textwidth]{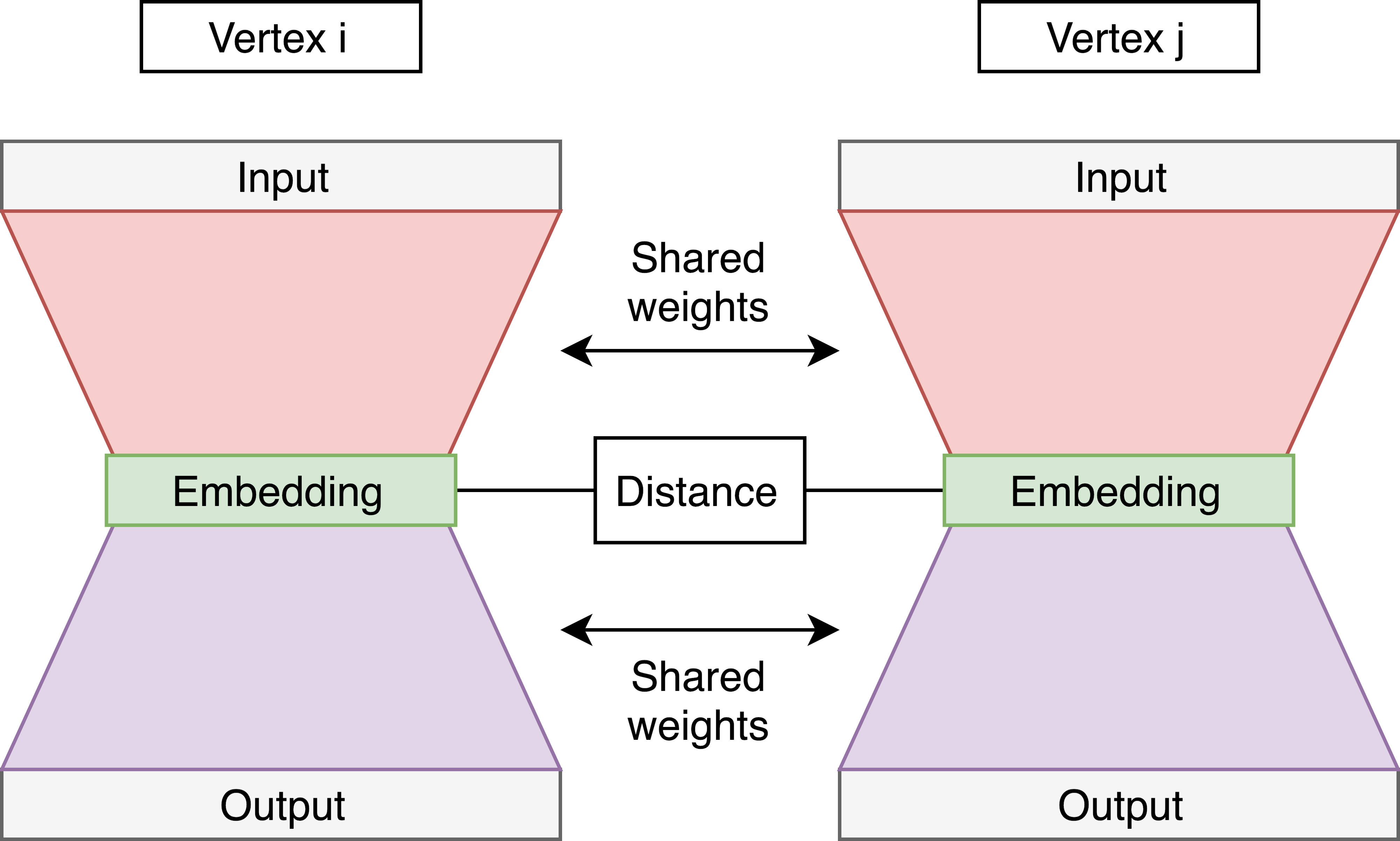}
\caption{Illustration of the SDNE model. The embedding layer (denoted $Z$) is shown in green. Reprinted with permission from \citep{godec_2018}.}
\label{fig:auto-encoder}
\end{figure*}

\paragraph{Structural Deep Network Embedding} (SDNE) \citep{wang2016structural} learns auto-encoders that preserve first and second-order node proximity~(\cref{subsec:def}).
The SDNE encoder takes as input a node vector: a row of the adjacency matrix as they explicitly set $s(W) = W$, and produces node embeddings $Z$.
The SDNE decoder return a reconstruction $\widehat{W}$, which is trained to recover the original graph adjacency matrix~(\cref{fig:auto-encoder}). 
SDNE preserves second order node proximity by minimizing the graph regularization loss:
$$||(s(W)-\widehat{W})\cdot B||_F^2+\alpha_{\mathrm{SDNE}}\sum_{ij}s(W)_{ij}||Z_i-Z_j||_2^2,$$
where $B$ is the indicator matrix for $s(W)$ with $B=1[s(W)>0]$. 
Note that the second term is the regularization loss used by distance-based shallow embedding methods. 
The first term is similar to the matrix factorization regularization objective, except that $\widehat{W}$ is not computed using outer products. 
Instead, SDNE computes a unique embedding for each node in the graph using a decoder network.  

\paragraph{Deep neural Networks for learning Graph Representations} (DNGR)~\citep{cao2016deep} Similar to SDNE, DNGR uses deep auto-encoders to encode and decode a node similarity matrix, $s(W)$. 
The similarity matrix is computed using a probabilistic method called random surfing, that returns a probabilistic similarity matrix through graph exploration with random walks. 
Therefore, DNGR captures higher-order dependencies in the graph. 
The similarity matrix $s(W)$ is then encoded and decoded with stacked denoising auto-encoders \citep{vincent2010stacked}, which allows to reduce the noise in $s(W)$.
DNGR is optimized by minimizing the reconstruction error: $$\mathcal{L}_{G,\mathrm{REG}}(W, \widehat{W};\Theta)=||s(W)-\widehat{W}||_F^2.$$

\subsection{Graph neural networks} \label{subsec:neighbour_unsup}
In graph neural networks, both the graph structure and node features are used in the encoder function to learn structural representations of nodes:
$$Z=\mathrm{ENC(X,W;\Theta^E)}.$$
We first review unsupervised graph neural networks, and will cover supervised graph neural networks in more details in~ \cref{sec:sup_models}. 

\paragraph{Variational Graph Auto-Encoders} (VGAE) \citep{kipf2016variational} use graph convolutions \citep{kipf2016semi} to learn node embeddings $Z=\mathrm{GCN}(W,X;\Theta^E)$ (see~\cref{subsec:gnn_sup} for more details about graph convolutions). 
The decoder is an outer product: $\mathrm{DEC}(Z;\Theta^D)=ZZ^\top.$
The graph regularization term is the sigmoid cross entropy between the true adjacency and the predicted edge similarity scores:
$$\mathcal{L}_{G,\mathrm{REG}}(W, \widehat{W};\Theta)=-\bigg(\sum_{ij}(1-W_{ij})\mathrm{log}(1-\sigma(\widehat{W}_{ij}))+W_{ij}\mathrm{log}\ \sigma(\widehat{W}_{ij})\bigg).$$
Computing the regularization term over all possible nodes pairs is computationally challenging in practice, and the Graph Auto Encoders (GAE) model uses negative sampling to overcome this challenge.

Note that GAE is a deterministic model but the authors also introduce variational graph auto-encoders (VGAE), where they use variational auto-encoders to encode and decode the graph structure.
In VGAE, the embedding $Z$ is modelled as a latent variable with a standard multivariate normal prior $p(Z)=\mathcal{N}(Z|0,I)$ and the amortized inference network $q_{\Phi}(Z|W,X)$ is also a graph convolution network.
VGAE is optimized by minimizing the corresponding negative evidence lower bound:
\begin{align*}
    \mathrm{NELBO}(W,X;\Theta)&=-\mathbb{E}_{q_{\Phi}(Z|W,X)}[\mathrm{log}\ p(W|Z)]+\mathrm{KL}(q_{\Phi}(Z|W,X)||p(Z))\\
    &=\mathcal{L}_{G,\mathrm{REG}}(W, \widehat{W};\Theta) + \mathrm{KL}(q_{\Phi}(Z|W,X)||p(Z)).
\end{align*}
\paragraph{Iterative generative modelling of graphs}
(Graphite) \citep{grover2018graphite} extends GAE and VGAE by introducing a more complex decoder, which iterates between pairwise decoding functions and graph convolutions.
Formally, the graphite decoder repeats the following iteration:
\begin{align*}
    \widehat{W}^{(k)}&=\frac{Z^{(k)}{Z^{(k)}}^\top}{||Z^{(k)}||_2^2}+\frac{11^\top}{|V|}\\
    Z^{(k+1)}&=\mathrm{GCN}(\widehat{W}^{(k)},Z^{(k)})
\end{align*}
where $Z^{(0)}$ are initialized using the output of the encoder network.
By using this parametric iterative decoding process, Graphite learns more expressive decoders than other methods based on non-parametric pairwise decoding. 
Finally, similar to GAE, Graphite can be deterministic or variational. 

\paragraph{Deep Graph Infomax} (DGI) \citep{velickovic2018dgi} is an unsupervised graph-level embedding method. 
Given one or more \textit{real} (positive) graphs, each with its adjacency matrix $W \in \mathbb{R}^{|V| \times |V|}$ and node features $X \in \mathbb{R}^{|V| \times d_0}$, this method creates \textit{fake} (negative) adjacency matrices $W^{-} \in \mathbb{R}^{|V^-| \times |V^-|}$ and their features $X^{-} \in \mathbb{R}^{|V^-| \times d_0}$.
It trains (i) an encoder that processes real and fake samples, respectively giving $Z = \textrm{ENC}(X, W; \Theta^{E}) \in \mathbb{R}^{|V| \times d}$ and $Z^- = \textrm{ENC}(X^-, W^-; \Theta^{E}) \in \mathbb{R}^{|V| \times d}$, (ii) a (readout) graph pooling function $\mathcal{R} : \mathbb{R}^{|V| \times d} \rightarrow \mathbb{R}^{d}$, and (iii) a descriminator function $\mathcal{D} : \mathbb{R}^d \times \mathbb{R}^d \rightarrow [0, 1]$ which is trained to output $\mathcal{D}(Z_i, \mathcal{R}(Z)) \approx 1$ and $\mathcal{D}(Z^-_j, \mathcal{R}(Z^-)) \approx 0$, respectively, for nodes corresponding to given graph $i \in V$ and fake graph $j \in V^-$. Specifically, DGI optimizes:
\begin{equation}
\label{eq:dgi}
    \min_{\Theta} - \mathop{\mathbb{E}}_{X, W} \sum_{i=1}^{|V|} \log \mathcal{D}(Z_i, \mathcal{R}(Z))  - \mathop{\mathbb{E}}_{X^{-}, W^{-}} \sum_{j=1}^{|V^-|} \log \left( 1 - \mathcal{D}(Z^-_j, \mathcal{R}(Z^-)) \right),
\end{equation}
where $\Theta$ contains $\Theta^E$ and the parameters of $\mathcal{R}, \mathcal{D}$. In the first expectation, DGI samples from the real (positive) graphs. If only one graph is given, it could sample some subgraphs from it (e.g. connected components). The second expectation samples fake (negative) graphs. In DGI, fake samples exhibit the real adjacency $W^{-} := W$ but fake features $X^{-}$ are a row-wise random permutation of real $X$, though other negative sampling strategies are plausible.
The $\textrm{ENC}$ used in DGI is a graph convolutional network, though any GNN can be used. The readout $\mathcal{R}$ summarizes an entire (variable-size) graph to a single (fixed-dimension) vector. \citet{velickovic2018dgi} use $\mathcal{R}$ as a row-wise mean, though other graph pooling might be used e.g. ones aware of the adjacency, $\mathcal{R} : \mathbb{R}^{|V| \times d} \times \mathbb{R}^{|V| \times |V|} \rightarrow \mathbb{R}^{d}$.

The optimization~(\cref{eq:dgi}) is shown by \cite{velickovic2018dgi} to maximize a lower-bound on the Mutual Information (MI) between the outputs of the encoder and the graph pooling function. In other words, it maximizes the MI between individual node representations and the graph representation.

Graphical Mutual Information \citep[GMI,][]{peng2020graph} presents another MI alternative: rather than maximizing MI of node information and an entire graph, GMI maximizes the MI between the representation of a node and its neighbors.


\tikzstyle{l} = [draw, -latex',thick]
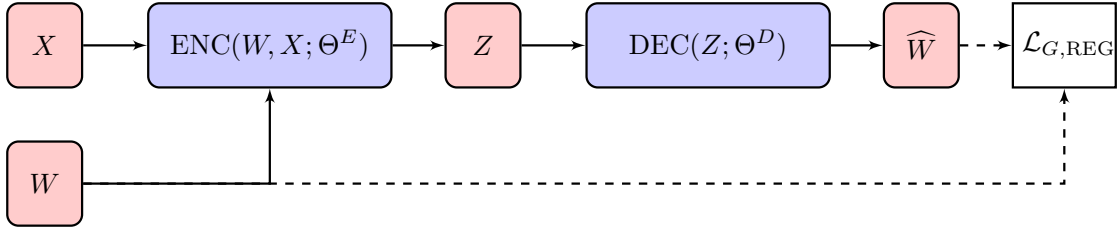
\begin{figure}[t]
    \centering
    \resizebox{0.9\textwidth}{!}{\renewcommand{\arraystretch}{1.1}
    \begin{tikzpicture}[auto]
    \node [data] (X) {$X$};
    \node [data, below=of X] (W) {$W$};
    \node [module, right=of X] (ENC) {$\mathrm{ENC}(W, X;\Theta^E)$};
    \node [data, right=of ENC] (Z) {$Z$};
    \node [module, right=of Z] (DEC1) {$\mathrm{DEC}(Z;\Theta^D)$};
    \node [data, right=of DEC1] (W_hat) {$\widehat{W}$};
    \node [loss, right=of W_hat] (loss) {$\mathcal{L}_{G,\mathrm{REG}}$};

    \path [l] (X) -- (ENC);
    \path [l] (ENC) -- (Z);
    \path [l] (Z) -- (DEC1);
    \path [l] (W) -| (ENC);
    \path [l] (DEC1) -- (W_hat);
    \path [l2] (W_hat) -- (loss);
    \path [l2] (W) -| (loss);


\end{tikzpicture}}
    \caption{Unsupervised graph neural networks. Graph structure and input features are mapped to low-dimensional embeddings using a graph neural network encoder. Embeddings are then decoded to compute a graph regularization loss (unsupervised).}\label{fig:agg_unsup}
\end{figure}

\subsection{Summary of unsupervised embedding methods}
This section presented a number of unsupervised embedding methods. Specifically, the only supervision signal is the graph itself, but no labels for nodes or the graph are processed by these methods.

Some of these methods (Sec.~\ref{sec:unsupervisedshallow}) are shallow, and ignore the node features $X$ even if they exist. These shallow methods program the encoder as a ``look-up table'', parametrizing it by matrix $\in \mathbb{R}^{|V| \times d}$, where each row stores $d$-dimensional embedding vector for a node. These methods are applicable to transductive tasks where is only one graph: it stays fixed between training and inference.

Auto-encoder methods (Sec.~\ref{subsec:auto_enc}) are deeper, though they still ignore node feature matrix $X$.
These are feed-forward neural networks where the network input is the adjacency matrix $W$. These methods are better suited when new nodes are expected at inference test time.

Finally, Graph neural networks (Sec.~\ref{subsec:neighbour_unsup}) are deep methods that process both the adjacency $W$ and node features $X$. These methods are inductive, and are generally empericially outperform the above two classes, for node-classification tasks, especially when nodes have features.

For all these unsupervised methods, the model output on the entire graph is $\in \mathbb{R}^{|V| \times |V|}$ that the objective function encourages to well-predict the adjacency $W$ or its transformation $s(W)$. As such, these models can compute latent representations of nodes that trained to reconstruct the graph structure. This latent representation can subsequently be used for tasks at hand, including, link prediction, node classification, or graph classification.

\section{Supervised Graph Embedding}\label{sec:sup_models}
A common approach for supervised network embedding is to use an unsupervised network embedding method, like the ones described in~\cref{sec:unsup_models} to first map nodes to an embedding vector space, and then use the learned embeddings as input for another neural network.
However, an important limitation with this two-step approach is that the unsupervised node embeddings might not preserve important properties of graphs (e.g. node labels or attributes), that could have been useful for a downstream supervised task.

Recently, methods combining these two steps, namely learning embeddings and predicting node or graph labels, have been proposed.
We describe these methods next.
\subsection{Shallow embedding methods}\label{subsec:shallow_sup}
Similar to unsupervised shallow embedding methods, supervised shallow embedding methods use embedding look-ups to map nodes to embeddings.
However, while the goal in unsupervised shallow embeddings is to learn a good graph representation, supervised shallow embedding methods aim at doing well on some downstream prediction task such as node or graph classification. 
\paragraph{Label propagation} 
(LP)~\citep{zhu2002learning} is a very popular algorithm for graph-based semi-supervised node classification.
It directly learns embeddings in the label space, i.e. the supervised decoder function in LP is simply the identity function:
$$\hat{y}^N=\mathrm{DEC}( Z;\Theta^C)=Z.$$
In particular, LP uses the graph structure to smooth the label distribution over the graph by adding a regularization term to the loss function, where the underlying assumption is that neighbor nodes should have similar labels (i.e. there exist some label consistency between connected nodes). 
The regularization in LP is computed with Laplacian eigenmaps:
\begin{align}
\mathcal{L}_{G,\mathrm{REG}}(W, \widehat{W};\Theta)&=\sum\limits_{ij}W_{ij}\widehat{W}_{ij}\label{eq:lp_loss} \\
    \text{where}\ \ \widehat{W}_{ij}&=||\hat{y}^N_i-\hat{y}^N_j||_2^2.
\end{align}
LP minimizes this energy function over the space of functions that take fixed values on labelled nodes (i.e. $\hat{y}^N_i=y^N_i\ \forall i|v_i\in V_L$) using an iterative algorithm that updates a unlabelled node's label distribution via the weighted average of its neighbors' labels.

There exists variants of this algorithm such as Label Spreading (LS)~\citep{zhou2004learning}, which minimizes the energy function:
\begin{align}
    \mathcal{L}_{G,\mathrm{REG}}(W, \widehat{W};\Theta)&=\sum\limits_{ij}W_{ij}\bigg|\bigg|\frac{\hat{y}^N_i}{\sqrt{D_i}}-\frac{\hat{y}^N_j}{\sqrt{D_j}}\bigg|\bigg|_2^2,\label{eq:label_spreading}
\end{align}
where $D_i=\sum_jW_{ij}$ is the degree of node $v_i$. 
The supervised loss in label spreading is simply the sum of distances between predicted labels and ground truth labels (one-hot vectors):
$$\mathcal{L}_\mathrm{SUP}^N(y^N, \hat{y}^N;\Theta)=\sum\limits_{i|v_i\in V_L}^{}||y^N_i-\hat{y}^N_i||_2^2.$$
Note that the supervised loss is computed over labeled nodes only, while the regularization term is computed over all nodes in the graph.
These methods are expected to work well with \textit{consistent} graphs, that is graphs where node proximity in the graph is positively correlated with label similarity.
%

\tikzstyle{l} = [draw, -latex',thick]
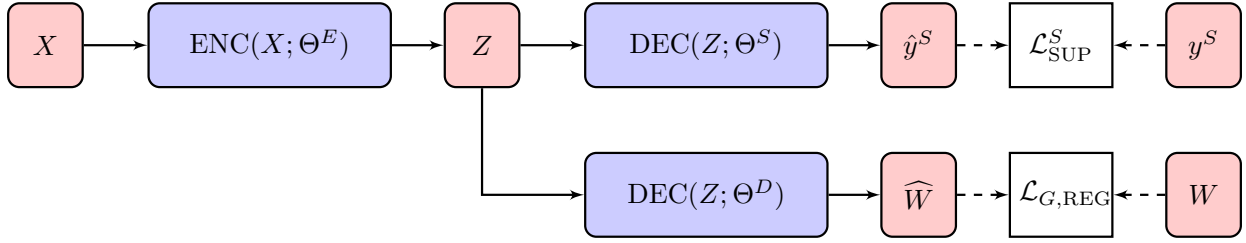
\begin{figure}[t]
    \centering
    \resizebox{\textwidth}{!}{\renewcommand{\arraystretch}{1.1}
    \begin{tikzpicture}[auto]
    \node [data] (X) {$X $};
    \node [module, right=of X] (ENC) {$\mathrm{ENC}(X ;\Theta^E)$};
    \node [data, right=of ENC] (Z) {$Z$};
    \node [module, right=of Z] (DEC1) {$\mathrm{DEC}(Z;\Theta^S)$};
    \node [module, below=of DEC1] (DEC2) {$\mathrm{DEC}(Z;\Theta^D)$};
    \node [data, right=of DEC1] (y_hat) {$\hat{y}^S$};
    \node [data, right=of DEC2] (W_hat) {$\widehat{W}$};
    \node [loss, right=of W_hat] (L_reg) {$\mathcal{L}_{G,\mathrm{REG}}$};
    \node [loss, right=of y_hat] (L_sup) {$\mathcal{L}^S_{\mathrm{SUP}}$};
    \node [data, right=of L_sup] (y) {$y^S$};
    \node [data, right=of L_reg] (W) {$W$};

    \path [l] (X) -- (ENC);
    \path [l] (ENC) -- (Z);
    \path [l] (Z) -- (DEC1);
    \path [l] (DEC1) -- (y_hat);
    \path [l] (Z) |- (DEC2);
    \path [l] (DEC2) -- (W_hat);
    \path [l2] (W_hat) -- (L_reg);
    \path [l2] (y_hat) -- (L_sup);
    \path [l2] (W) -- (L_reg);
    \path [l2] (y) -- (L_sup);

\end{tikzpicture}}
    \caption{Supervised graph regularization methods. The graph structure is not used in the encoder nor the decoder networks. It instead acts as a regularizer in the loss function.}\label{fig:reg}
\end{figure}

\subsection{Graph regularization methods}\label{subsec:reg}
Supervised graph regularization methods also aim at learning to predict graph properties such as node labels. 
Similar to shallow embeddings, these methods compute a graph regularization loss defined over the graph structure, and a supervised loss for the downstream task~(\cref{fig:reg}).
However, the main difference with shallow embeddings lies in the encoder network: rather than using embedding look-ups, graph regularization methods learn embeddings as parametric function defined over node features, which might capture valuable information for downstream applications. 
That is, encoder functions in these methods can be written as: 
$$Z=\mathrm{ENC(X;\Theta^E)}.$$
We review two types of semi-supervised \citep{chapellesemi} graph regularization approaches: Laplacian-based regularization methods and methods that use random walks to regularize embeddings.
\subsubsection{Laplacian}\label{subsubsec:laplacian_reg_sup}
\paragraph{Manifold Regularization} (ManiReg) \citep{belkin2006manifold} builds on the LP model and uses Laplacian Eigenmaps to smoothen the label distribution via the regularization loss in~\cref{eq:lp_loss}.
However, instead of using shallow embeddings to predict labels, ManiReg uses support vector machines to predict labels from node features. 
The supervised loss in ManiReg is computed as:
\begin{equation}\label{eq:hinge}
    \mathcal{L}_\mathrm{SUP}^N(y^N, \hat{y}^N;\Theta)=\sum\limits_{i|v_i\in V_L}^{}\sum_{1\le k\le C}^{} H(y_{ik}^N\hat{y}^N_{ik}),
\end{equation}
where $H(x)=\mathrm{max}(0, 1-x)$ is the hinge loss, $C$ is the number of classes, and $\hat{y}_i^N=f(X_i;\Theta^E)$ are computed using Reproducing Kernel Hilbert Space (RKHS) functions that act on input features. 
\paragraph{Semi-supervised Embeddings} (SemiEmb) \citep{weston2012deep} further extend ManiReg and instead of using simple SVMs, this method uses feed-forward neural networks (FF-NN) to learn embeddings $Z=\mathrm{ENC}(X;\Theta^E)$ and distance-based graph decoders:
\begin{align*}
    \widehat{W}_{ij}&=\mathrm{DEC}(Z;\Theta^D)_{ij}\\
    &=||Z_i - Z_j||^2
    \end{align*}
where $||\cdot||$ can be the L2 or L1 norm.
SemiEmb regularizes intermediate or auxiliary layers in the network using the same regularizer as the LP loss in~\cref{eq:lp_loss}.
SemiEmb uses FF-NN to predict labels from intermediate embeddings, which are then compared to ground truth labels via the Hinge loss in~\cref{eq:hinge}.

Note that SemiEmb leverages multi-layer neural networks and regularizes intermediate hidden representations, while LP does not learn intermediate representations, and ManiReg only regularizes the last layer.

\paragraph{Neural Graph Machines} (NGM)~\citep{bui2017neural} 
More recently, NGM generalize the regularization objective in~\cref{eq:lp_loss} to more complex neural architectures than feed-forward neural networks (FF-NN), such as Long short-term memory (LSTM) networks~\citep{hochreiter1997long} or CNNs~\citep{lecun1989backpropagation}.
in contrast with previous methods, NGM use the cross entropy loss for classification.
\scriptsize
\begin{table}[!t]
\renewcommand{\arraystretch}{1.3}
\label{tab:sup_reg}
\centering
\resizebox{\textwidth}{!}{%
\begin{tabular}{c | c | c | c | c | c }
\hline
 \bfseries Model & \bfseries $Z=\mathrm{ENC}(X ;\Theta^E)$ & $\widehat{y}^N=\mathrm{DEC}(Z ;\Theta^N)$ & $\widehat{W}_{ij}=\mathrm{DEC}(Z ;\Theta^D)_{ij}$ & $\mathcal{L}_{G,\mathrm{REG}}(W, \widehat{W};\Theta)$ & $\mathcal{L}_\mathrm{SUP}^N(y^N, \widehat{y}^N;\Theta)$ \\
\hline
LP 
& Shallow &  $\widehat{y}^N=Z$ & $||Z_i-Z_j||_2^2$ & $\sum_{ij}W_{ij}\widehat{W}_{ij}$ & $\widehat{y}^N_i=y^N_i$ is fixed $\forall v_i\in V_L$\\
LS 
& Shallow &  $\widehat{y}^N=D^{-1/2}Z$ & $||Z_i-Z_j||_2^2$ & $\sum_{ij}W_{ij}\widehat{W}_{ij}$ & $\sum_{i|v_i\in V_L}^{}||y^N_i-\widehat{y}^N_i||_2^2$ \\
ManiReg 
& RKHS &  $\widehat{y}^N=Z$ & $||Z_i-Z_j||_2^2$ & $\sum_{ij}W_{ij}\widehat{W}_{ij}$ & $\sum_{i|v_i\in V_L}^{}\sum_{1\le k\le C}^{} H(y_{ik}^N\widehat{y}^N_{ik})$ \\
SemiEmb 
& FF-NN & FF-NN & $||Z_i-Z_j||^2$ & $\sum_{ij}W_{ij}\widehat{W}_{ij}$ & $\sum_{i|v_i\in V_L}^{}\sum_{1\le k\le C}^{} H(y_{ik}^N\widehat{y}^N_{ik})$ \\
NGM 
& CNN, LSTM, ... & CNN, LSTM, ... & $ ||Z_i-Z_j||^2$ & $\sum_{ij}W_{ij}\widehat{W}_{ij}$ & $-\frac{1}{|V_L|}\sum_{i|v_i\in V_L} \sum_{1\le k \le C} y_{ik}^N\mathrm{log}\ \widehat{y}^N_{ik}$ \\
Planetoid 
& FF-NN & FF-NN & $Z_i^\top Z_j$ &$- \mathbb{E}_{(i, j, \gamma)} \log \sigma\left(\gamma \widehat{W}_{ij}\right)$ & $-\frac{1}{|V_L|}\sum_{i|v_i\in V_L} \sum_{1\le k \le C} y_{ik}^N\mathrm{log}\ \widehat{y}^N_{ik}$ \\ 
\hline
\end{tabular}}
\caption{An overview of supervised shallow and graph regularization methods, where the graph structure is leveraged through the graph regularization term $\mathcal{L}_{G,\mathrm{REG}}(W,\widehat{W};\Theta)$.
}
\end{table}
\normalsize
\subsubsection{Skip-gram}\label{subsubsec:skip_gram_sup}
The Laplacian-based regularization methods covered so far only capture first order proximities in the graphs. 
Skip-gram graph regularization methods further extend these methods to incorporate random walks, which are effective at capturing higher-order proximities. 
\paragraph{Planetoid}~\citep{yang2016revisiting}
Unsupervised skip-gram methods like node2vec and DeepWalk learn embeddings in a multi-step pipeline where random walks are first generated from the graph and then used to learn embeddings. 
These embeddings are not learned for a downstream classification task which might be suboptimal.
Planetoid extends random walk methods to leverage node label information during the embedding algorithm. 

Planetoid first maps nodes to embeddings $Z=[Z^c||Z^F]=\mathrm{ENC}(X;\Theta^E)$ with neural networks, where $Z^c$ are node embeddings that capture structural information while $Z^F$ capture node feature information.
The authors propose two variants, a transductive variant that directly learns embedding $Z^c$ (as an embedding lookup), and an inductive variant where $Z^c$ are computed with parametric mappings that act on input features $X$.
Embeddings are then learned by minimizing a supervised loss and a graph regularization loss, where the regularization loss measures the ability to predict context using nodes embeddings, while the supervised loss measures the ability to predict the correct label. 
More specifically, the regularization loss in Planetoid is given by:
\begin{align*}
    \mathcal{L}_{G,\mathrm{REG}}(W, \widehat{W}; \Theta) = - \mathbb{E}_{(i, j, \gamma)} \log \sigma\left(\gamma \widehat{W}_{ij}\right),
\end{align*}
with $\widehat{W}_{ij} = Z_i^\top Z_j$ and $\gamma\in\{-1, 1\}$ with $\gamma=1$ if $(v_i, v_j) \in E$ is a positive pair and $\gamma=-1$ if $(v_i, v_j)$ is a negative pair. 
The distribution under the expectation is directly defined through a sampling process\footnote{There are two kinds of sampling strategies to sample positive pairs of nodes $i, j$: (i.) samples drawn by conducting random walks, similar to DeepWalk
and (ii.) samples drawn from the same class i.e. $y_i = y_j$. These samples are positive i.e. with $\gamma=1$. The negative samples simply replace one of the nodes with another randomly-sampled (negative) node yielding $\gamma = -1$. The ratio of these kinds of samples are determined by hyperparameters.
}. 
The supervised loss in Planetoid is the negative log-likelihood of predicting the correct labels: 
\begin{align}
    \mathcal{L}_\mathrm{SUP}^{N}(y^N, \widehat{y}^N;\Theta) = -\frac{1}{|V_L|}\sum_{i|v_i\in V_L} \sum_{1\le k \le C} y_{ik}^N\mathrm{log}\ \widehat{y}^N_{ik},\label{eq:ce_loss}
\end{align}
where $i$ is a node's index while $k$ indicates label classes, and $\widehat{y}^N_{i}$ are computed using a neural network followed by a softmax activation, mapping $Z_i$ to predicted labels.
\subsection{Graph convolution framework}
\label{subsec:gc_framework}
We now focus on (semi-)supervised neighborhood aggregation methods, where the encoder uses input features and the graph structure to compute embeddings: $$Z=\mathrm{ENC}(X,W;\Theta^E).$$
We first review the graph neural network model---which was the first attempt to use deep learning techniques on graph-structured data---and other related frameworks such as message passing networks \citep{gilmer2017neural}.
We then introduce a new Graph Convolution Framework (\framework), which is designed specifically for convolution-based graph neural networks. 
While \framework{} and other frameworks overlap on some methods, \framework{} emphasizes the geometric aspects of convolution and propagation, allowing to easily understand similarities and differences between existing convolution-based approaches. 
\subsubsection{The Graph Neural Network model and related frameworks}\label{subsec:gnn_sup}
\paragraph{The Graph Neural Network model} (GNN) \citep{gori2005new, scarselli2009graph}
The first formulation of deep learning methods for graph-structured data dates back to the graph neural network GNN) model of~\citet{gori2005new}. 
This formulation views the supervised graph embedding problem as an information diffusion mechanism, where nodes send information to their neighbors until some stable equilibrium state is reached. 
More concretely, given randomly initialized node embeddings $Z^0$, the following recursion is applied until convergence:
\begin{align}
    Z^{t+1}=\mathrm{ENC}(X, W, Z^{t};\Theta^E),\label{eq:gnn1}
\end{align}
where parameters $\Theta^E$ are reused at every iteration. 
After convergence ($t=T$), the node embeddings $Z^T$ are used to predict the final output such as node or graph labels:
$$\hat{y}^S=\mathrm{DEC}(X, Z^{T};\Theta^S).$$
This process is repeated several times and the GNN parameters $\Theta^E$ and $\Theta^D$ are learned with backpropagation via the Almeda-Pineda algorithm \citep{almeida1987learning,pineda1988generalization}.
Note that by Banach's fixed point theorem, the iteration in~\cref{eq:gnn1} is guaranteed to converge to a unique solution when the iteration ENC is a contraction mapping.
In particular, \citet{scarselli2009graph} explore maps that can be expressed using message passing networks:
\begin{align*}
    Z_i^{t+1}&=\sum_{j|(v_i,v_j)\in E}f(X_i,X_j,Z^{t}_j;\Theta^E),
\end{align*}
where $f(\cdot)$ is a multi-layer perception (MLP) constrained to be a contraction mapping.
On the other hand, the decoder function in GNNs does not need to fulfill any constraint and can be any MLP.
\paragraph{Gated Graph Neural Networks} (GGNN)~\citep{li2015gated} Gated Graph Sequence Neural Networks (GGSNN) or their simpler version GGNN are similar to GNNs but remove the contraction mapping requirement. 
In GGSNNs, the recursive algorithm in~\cref{eq:gnn1} is relaxed and approximated by applying mapping functions for a fixed number of steps, where each mapping function is a gated recurrent unit~\citep{cho2014properties} with parameters shared for every iteration.
The GGSNN model is particularly useful for machine learning tasks with sequential structure (such as temporal graphs) as it outputs predictions at every step.
\paragraph{Message Passing Neural Networks} (MPNN)~\citet{gilmer2017neural}
In the same vein, MPNN provide a framework for graph neural networks, encapsulating many recent graph neural networks. 
In contrast with the GNN model which runs for an indefinite number of iterations, MPNN provide an abstraction for modern graph neural networks, which consist of multi-layer neural networks with a \textit{fixed} number of layers. 
At every layer $\ell$, message functions $f^{\ell}(.)$ compute messages using neighbors' hidden representations, which are then passed to aggregation functions $h^{\ell}(.)$:
\begin{align}
    m^{\ell+1}_i&=\sum_{j|(v_i,v_j)\in E}f^\ell(H^{\ell}_i,H^{\ell}_j)\label{eq:mp1}\\
    H^{\ell+1}_i&=h^{\ell}(H^{\ell}_i,m^{\ell+1}_i).\label{eq:mp2}
\end{align}
After $\ell$ layers of message passing, nodes' hidden representations encode structural information within $\ell$-hop neighborhoods. 
\citet{gilmer2017neural} explore additional variations of message functions within the MPNN framework, and achieve state-of-the-art results for prediction tasks defined on molecular graphs. 
\paragraph{GraphNet}
\cite{battaglia2016interaction} This framework further extends the MPNN framework to learn representations for edges, nodes and the entire graph using message passing functions. 
This framework is more general than the MPNN framework as it incorporates edge and graph representations. 

\tikzstyle{l} = [draw, -latex',thick]
\begin{figure}[t]
    \centering
    \resizebox{\textwidth}{!}{\renewcommand{\arraystretch}{1.1}
    \begin{tikzpicture}[auto]
    \node [data] (X) {$X$};
    \node [data, below=of X] (W) {$W$};
    \node [module, right=of X] (ENC) {$\mathrm{ENC}(W, X ;\Theta^E)$};
    \node [data, right=of ENC] (Z) {$Z$};
    \node [module, right=of Z] (DEC1) {$\mathrm{DEC}(Z;\Theta^S)$};
    \node [data, right=of DEC1] (y_hat) {$\widehat{y}^S$};
    \node [loss, right=of y_hat] (L_sup) {$\mathcal{L}^S_{\mathrm{SUP}}$};
    \node [data, right=of L_sup] (y_sup) {$y^S$};

    \path [l] (X) -- (ENC);
    \path [l] (ENC) -- (Z);
    \path [l] (Z) -- (DEC1);
    \path [l] (W) -| (ENC);
    \path [l] (DEC1) -- (y_hat);
    \path [l2] (y_hat) -- (L_sup);
    \path [l2] (y_sup) -- (L_sup);
\end{tikzpicture}}
    \caption{Supervised graph neural networks (GNNs). Rather than leveraging the graph structure to act as a regularizer, GNNs leverage the graph structure in the encoder to propagate information across neighbouring nodes and learn structural representations.
    Labels are then decoded and compared to ground truth labels (e.g. via the cross-entropy loss). 
    }\label{fig:agg_sup}
\end{figure}
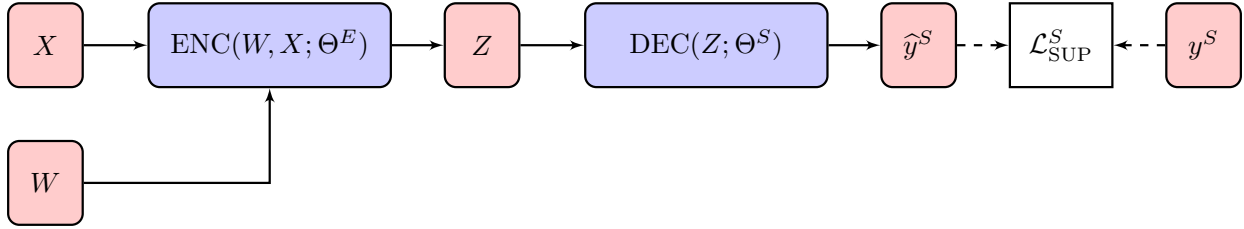

\subsubsection{Graph Convolution Framework}
We now introduce our Graph Convolution Framework (\framework{}); and as we shall see, many recent graph neural networks can be described using this framework.
Different from the MPNN and GraphNet frameworks, our framework focuses on convolution-based methods, and draws direct connections between convolutions on grids and graph convolutions.
While \framework{} does not include sophisticated message passing networks (e.g. messages computed with edge features), it emphasizes geometric properties of convolution operators, and provides a simple way to understand similarities and differences between state-of-the-art graph convolution methods.
\paragraph{\framework{}}
In \framework{}, node embeddings are initialized using input features $H^{0}=X\in\mathbb{R}^{|V|\times d_0}$, and then updated with multiple layers of graph convolutions. 
Graph convolution layers provide a generalization of standard convolutions on grids to graph-structured data and are composed of four main components:
\begin{itemize}\itemsep -2pt
\item \textbf{Patch functions}, which define the shape of convolutional filters (specifies which nodes interact with each other at every step of convolution), that is matrices of size $|V|\times |V|$: 
$$(f_1(W, H^{\ell}), \ldots, f_K(W, H^{\ell})),$$
where $H^{\ell}$ are node features at layer $\ell$ and $K$ is the total number of patches.
Note that the number of patches $K$ might be defined in the spectral domain (e.g. rank of a matrix) or in the spatial domain (e.g. different neighborhood sizes).  
In standard CNNs (which are defined in the spatial pixel domain), these patches usually have rectangular shapes, where nodes (pixels in images) communicate with their top, left, bottom, and right neighbors. 
However, since graphs do not have a grid-like structure, the shape of convolutional filters does not follow regular patterns and is instead defined by the graph structure itself.
While most methods use non-parametric patches at every layer, some methods such as attention-based methods~(\cref{subsec:spatial_attention}) learn patches using parametric functions. 
\item \textbf{Convolution filters' weights} at every layer, which are $d_{\ell}\times d_{\ell+1}$ matrices, representing the filter weights for each patch:
$$(\Theta_1^\ell, \ldots, \Theta_K^\ell).$$ 
Each column can be interpreted as a single convolution filter's weight, and we stack $d_{\ell+1}$ filters filters to compute features in the next layer.
Similarly, $d_\ell$ and $d_{\ell+1}$ are analogous to the number of channels in CNNs for layer $\ell$ and $\ell+1$ respectively. 
At every layer in the \framework{}, hidden representations $H^{\ell}$ are convolved with every patch using the convolution filter weights:
$$m_k^{\ell+1}=f_k(W, H^{\ell})H^{\ell}\Theta_k^\ell\ \  \text{ for }\ \ 1\le k\le K.$$
\item \textbf{Merging functions}, which combine outputs from multiple convolution steps into one representation: 
$$H^{\ell+1} = h(m_1^{\ell+1}, \ldots, m_K^{\ell+1}).$$
For instance, $h(\cdot)$ can be averaging or concatenation along the feature dimension followed by some non-linearity. 
Alternatively, $h(\cdot)$ can also be a more complicated operation parameterized by a neural network. 
\end{itemize}
After $L$ convolution layers, nodes' embeddings $Z=H^{L}$ can be used to decode node or graph labels. 
Next, we review state-of-the-art GNNs, including spectral and spatial graph convolution methods using the proposed \framework{} framework.
\begin{table}[!t]
\renewcommand{\arraystretch}{1.3}
\label{tab:sup_conv}
\centering
\resizebox{\textwidth}{!}{%
\begin{tabular}{c | c | c | c }
\hline
\bfseries Method & \bfseries Model & \bfseries $g_k(.)$ & \bfseries $h(m_1,\ldots,m_k)$ \\
\hline
\multirow{1}{*}{{Spectrum-based:}}
& \multirow{2}{*}{SCNN} 
& \multirow{2}{*}{$g_k(U)=u_ku_k^\top $} & \multirow{2}{*}{$\sigma(\sum_k m_k)$} \\ 
$\widetilde{L}=U\Lambda U^\top,\ f_k(W,H )=g_k(U)$ & & & \\
\hline
\multirow{1}{*}{{Spectrum-free:}}
&  ChebNet 
& $g_k(W, D)=T_k(\frac{2(I-D^{-1/2}WD^{-1/2})}{\lambda_{max}(I-D^{-1/2}WD^{-1/2})}-I)$ & $\sigma(\sum_k m_k)$ \\ 
\cline{2-4}
 $f_k(W,H )=g_k(W, D)$ &  GCN 
&$g_1(W,D)=(D+I)^{-1/2}(W+I)(D+I)^{-1/2}$ & $\sigma(m_1)$\\ \cline{1-4} 
\multirow{1}{*}{{Spatial: }} &  SAGE-mean 
 & $g_1(W,D)=I,g_2(W,D)\sim\mathcal{U}_{\text{norm}}(D^{-1}W, q)$ & $\sigma(m_1+m_2)$\\ \cline{2-4}
$f_k(W,H)=g_k(W, D)$ &  GGNN 
&  $g_1(W,D)=I,g_2(W,D)=W$ & $\mathrm{GRU}(m_1,m2)$\\ \cline{2-4}
\hline
\multirow{1}{*}{{Attention:}}
&  MoNet 
& $g_k(U^{s})=\mathrm{exp}(-\frac{1}{2}(U^{s}-\mu_k)^\top \Sigma_k^{-1}(U^{s}-\mu_k))$ & $\sigma(\sum_km_k)$ \\ \cline{2-4}
$f_k(W,H )=\alpha(W\cdot g_k(H ))$ &  GAT 
& $g_k(H )=\mathrm{LeakyReLU}(H  B^\top b_0 \oplus b_1^\top B{H }^\top)$ & $\sigma([m_1||\ldots||m_k])$\\ \cline{2-4}
\hline
\end{tabular}}
\caption{An overview of graph convolution methods described using \framework{}.}
\end{table}
\subsection{Spectral Graph Convolutions}\label{subsec:spectral}
Spectral methods apply convolutions in the the spectral domain of the graph Laplacian matrix.
These methods broadly fall into two categories: \textit{spectrum-based methods}, which explicitly compute the Laplacian's eigendecomposition, and \textit{spectrum-free} methods, which are motivated by spectral graph theory but do not explicitly compute the Laplacian's eigenvectors.
One disadvantage of spectrum-based methods is that they rely on the spectrum of the graph Laplacian and are therefore domain-dependent (i.e. cannot generalize to new graphs). 
Moreover, computing the Laplacian's spectral decomposition is computationally expensive.
Spectrum-free methods overcome these limitations by providing approximations for spectral filters. 
\subsubsection{Spectrum-based methods}\label{subsec:spectrum_based}
Spectrum-based graph convolutions were the first attempt to generalize convolutions to non-Euclidean graph domains. 
Given a signal $x\in\mathbb{R}^{|V|}$ defined on a Euclidean discrete domain (e.g. grid), applying any linear translation-equivariant operator (i.e. with a Toeplitz structure) $\Theta$ in the discrete domain is equivalent to elementwise multiplication in the Fourier domain:
\begin{equation}\label{eq:cnn_fourier}
  \mathcal{F}({\Theta}x)=\mathcal{F}x\cdot\mathcal{F}\theta.  
\end{equation}
In non-Euclidean domains, the notion of translation (shift) is not defined and it is not trivial to generalize spatial convolutions operators (${\Theta}$) to non-Euclidean domains. 
Note that~\cref{eq:cnn_fourier} can be equivalently written as:
\begin{align*}
       {\Theta}x=\mathcal{F}^{-1}(\mathcal{F}x \cdot\mathcal{F}\theta).  
\end{align*}
While the left hand side is the Euclidean spatial convolution which is not defined for general graphs, the right hand side is a convolution in the Fourier domain which is defined for non-Euclidean domains. 
In particular, if $\widetilde{L}=I-D^{-1/2}WD^{-1/2}$ is the normalized Laplacian of a non-Euclidean graph, it is a real symmetric positive definite matrix and admits an orthonormal eigendecomposition: $\widetilde{L}=U\Lambda U^\top$.
If $x\in\mathbb{R}^{|V|}$ is a signal defined on nodes in the graph, the discrete graph Fourier transform and its inverse can be written as: 
\begin{align*}
    \mathcal{F}x=\hat{x}=U^\top x\ \ \text{and}\ \ \mathcal{F}^{-1} \hat{x}=U \hat{x}.
\end{align*}

Spectral graph convolutions build on this observation to generalize convolutions to graphs, by learning convolution filters in the spectral domain of the normalized Laplacian matrix:
\begin{align*}
    x * \theta&=U(U^\top x \cdot U^\top\theta)\\
    &=U\mathrm{diag}(U^\top\theta)U^\top x
\end{align*}
Using \framework, patch functions in spectrum-based methods can be expressed in terms of eigenvectors of the graph normalized Laplacian:
$$f_k(W, H^{\ell} )=g_k(U)$$
for some function $g_k(.).$
Note that this dependence on the spectrum of the Laplacian makes spectrum-based methods domain-dependent (i.e. they can only be used in transductive settings). 
\paragraph{Spectral Convolutional Neural Networks} (SCNN) \citep{bruna2013spectral} learn convolution filters as multipliers on the eigenvalues of the normalized Laplacian.
SCNN layers compute feature maps at layer $\ell+1$ with:
\begin{equation}\label{eq:scnn_filters}
    H^{\ell+1}_{:,j} = \sigma\bigg(\sum\limits_{i=1}^{d_\ell }U_{K}F_{i,j}^{\ell} U_{K}^\top H^{\ell} _{:,i}\bigg) \text{,  }1\le j\le d_{\ell +1} \text{ and }1\le i\le d_\ell
\end{equation}
where $\sigma(\cdot)$ is a non-linear transformation, $U_{K}$ is a $|V|\times K$ matrix containing the top $K$ eigenvectors of $\widetilde{L}$ and $F_{i,j}^{\ell} $ are $K\times K$ trainable diagonal matrices representing filters' weights in the spectral domain.
We note that this spectral convolution operation can equivalently be written as: 
\begin{align}
    H^{\ell+1}_{:,j} = \sigma\bigg(\sum\limits_{k=1}^{K}u_ku_k^\top \sum\limits_{i=1}^{d_\ell }F_{i,j,k}^{\ell} H^{\ell} _{:,i}\bigg),\label{eq:spectral_conv}
\end{align}
where $(u_k)_{k=1,\ldots,K}$ are the top $K$ eigenvectors of $\widetilde{L}$ and $F_{i,j,k}^{\ell}$ is the $k^{th}$ diagonal element of $F_{i,j}^{\ell}$. 
We can also write~\cref{eq:spectral_conv} using matrix notation as:
$$H^{\ell+1} = \sigma\bigg(\sum\limits_{k=1}^{K}u_ku_k^\top H^{\ell} \Theta^\ell _k\bigg),$$
where $\Theta^\ell _k$ are trainable matrices of shape $d_\ell \times d_{\ell +1}$ containing the filter weights.
Using notation from \framework, SCNNs use patch functions expressed in terms of eigenvectors of the graph Laplacian $g_k(U)=u_ku_k^\top$, and the merging function $h(.)$ is the sum operator followed by a non-linearity $\sigma(.)$.

Euclidean grids have a natural ordering of nodes (top, left, bottom, right) allowing the use of spatially localized convolution filters with fixed size, independent of the input size. 
In contrast, SCNN layers require $\mathcal{O}(d_\ell d_{\ell +1}K)$ parameters, which is not scalable if $K$ is $\mathcal{O}(|V|)$. 
\citet{bruna2013spectral,henaff2015deep} note that spatial localization in the graph domain is equivalent to smoothness in the spectral domain, and propose smooth spectral multipliers in order to reduce the number of parameters in the model and avoid overfitting.
Instead of learning $K$ free parameters for each filter $F_{ij}^\ell $, the idea behind smooth spectral multipliers is to parameterize $F_{ij}^\ell $ with polynomial interpolators such as cubic splines and learn a fixed number of interpolation coefficients.
This modeling assumption leads to a constant number of parameters, independent of the graph size $|V|$.

In practice, SCNNs can be used for node classification or graph classification with graph pooling.
However, SCNNs have two major limitations: (1) computing the eigendecomposition of the graph Laplacian is computationally expensive and (2) this method is domain-dependent, as its filters are eigen-basis dependent and cannot be shared across graphs.  
%
%
%
\subsubsection{Spectrum-free methods}\label{subsec:spectrum_free}
\begin{figure}
\centering
\begin{subfigure}[b]{0.28\textwidth}
    \input{figures/gcn1.tex}
    \caption{GCN aggregation.}
    \label{fig:gcn}
    \end{subfigure}
    \begin{subfigure}[b]{0.28\textwidth}
    \begin{tikzpicture}[scale=0.8, transform shape]
\definecolor{paramcolor}{HTML}{50ae55}
\definecolor{inputcolor}{HTML}{fd9727}
\definecolor{outputcolor}{HTML}{f1453d}
\tikzset{
	inputmat/.style = {rectangle, draw=inputcolor!70, fill=inputcolor!40, thick, minimum width=1.4cm, minimum height = 1cm},
	outputmat/.style = {rectangle, draw=outputcolor!0, fill=outputcolor!50, thick, minimum width=0.6cm, minimum height = 1.0cm},
	thickermat/.style = {rectangle, draw=black!50, fill=black!20, thick, minimum width=1.4cm, minimum height = 1cm},
	parameter/.style = {rectangle, draw=paramcolor!70, fill=paramcolor!40, thick, minimum width=0.6cm, minimum height = 1.4cm},
}
\node [inputmat](x) [draw] { };
\node [above =0cm of x] (xlabel) {$H^{\ell}$ };
\node [draw, below of=x](mult) {$\times$};
\node [left =0.35cm of mult] (a) {$\widetilde{D}^{-1/2}\widetilde{W}\widetilde{D}^{-1/2}$};
\node [draw, below of=mult](wmult) {$\times$};
\node [parameter, right =0.55cm of wmult](w) {};
\node [below=0cm of w](wl) {$\Theta^{\ell}$}; %
\node [outputmat, below of=wmult](y) {};
\node [below =0cm of y] (xlabel) {$H^{\ell + 1}$ };

\draw[-{>}] (x) -- (mult);
\draw[-{>}] (a) -- (mult);
\draw[-{>}] (mult) -- (wmult);
\draw[-{>}] (w) -- (wmult);
\draw[-{>}] (wmult) -- (y);
\end{tikzpicture}
    \caption{GCN layer.}
    \end{subfigure}
    \begin{subfigure}[b]{0.42\textwidth}
    \includegraphics[width=\textwidth]{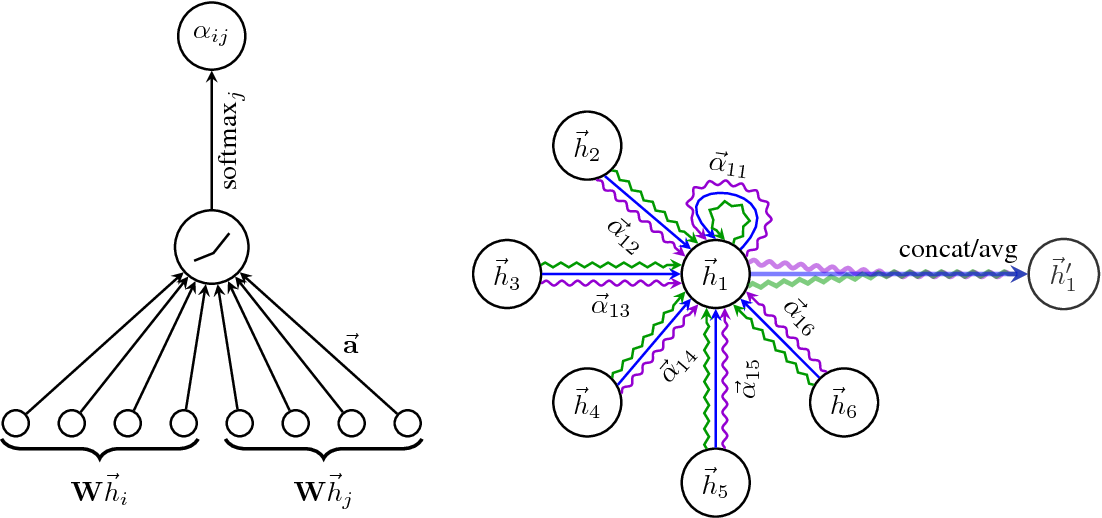}
        \caption{GAT aggregation.}
        \label{fig:gat}
\end{subfigure}
\caption{An illustration of neighborhood aggregation methods. Reprinted with permission from \citep{abu2019mixhop, velickovic2017graph}.}\label{fig:agg}
\end{figure}
We now cover spectrum-free methods, which approximate convolutions in the spectral domain overcoming computational limitations of SCNNs by avoiding explicit computation of the Laplacian's eigendecomposition.
SCNNs filters are neither localized nor parametric, in the sense that the parameters in $F_{ij}^\ell $ in~\cref{eq:spectral_conv} are all free.
To overcome this issue, sprectrum-free methods use polynomial expansions to approximate spectral filters in~\cref{eq:scnn_filters} via:
$$F_{i,j}^\ell=P^\ell_{ij}(\Lambda)$$ 
where $P^\ell_{ij}(\cdot)$ is a finite degree polynomial. 
Therefore, the total number of free parameters per filter depends on the polynomial's degree, which is independent of the graph size.   
Assuming all eigenvectors are kept in~\cref{eq:scnn_filters}, it can be rewritten as:
$$H^{\ell +1}_{:,j} = \sigma\bigg(\sum\limits_{i=1}^{d_\ell }P^\ell_{ij}(\Lambda)H^\ell _{:,i}\bigg).$$
If we write $P^\ell_{ij}(\lambda)=\sum\limits_{k=1}^K \theta_{i,j,k}^\ell \left(\lambda^k\right)$, this yields in matrix notation:
$$H^{\ell +1} = \sigma\bigg(\sum\limits_{k=1}^K \left(\widetilde{L}^k\right) H^\ell\Theta^\ell_k\bigg),$$
where $\Theta^\ell_k$ is the matrix containing the polynomials' coefficients.
These filters are $k$-localized, in the sense that the receptive field of each filter is $k$, and only nodes at a distance less than $k$ will interact in the convolution operation.
Since the normalized Laplacian is expressed in terms of the graph adjacency and degree matrices, we can write patch functions in spectrum-free method using notation from \framework:
$$f_k(W, H^\ell )=g_k(W, D).$$

\paragraph{Chebyshev Networks} (ChebNets)~\citep{defferrard2016convolutional} use the Chebyshev expansion~\citep{hammond2011wavelets} to approximate spectral filters. 
Chebyshev polynomials form an orthonormal basis in $[-1, 1]$ and can be computed efficiently with the recurrence:
\begin{equation}\label{eq:cheby_rec}
    T_0(x)=1,\ \ T_1(x)=x,\ \ \text{and}\ \ T_{k}(x)=2xT_{k-1}(x)-T_{k-2}(x)\ \ \text{for}\ k\ge 2. 
\end{equation}
In order to use Chebyshev polynomials, ChebNets rescale the normalized adjacency martrix $\widetilde{L}$ to ensure that its eigenvalues are in $[-1, 1]$. 
The convolution step in ChebNet can be written as:
$$H^{\ell  +1}=\sigma\bigg(\sum\limits_{k=1}^KT_k\bigg(\frac{2}{\lambda_{max}(\widetilde{L})}\widetilde{L}-I\bigg)H^\ell \Theta^\ell _k\bigg),$$
where $\lambda_{max}(\widetilde{L})$ is the largest eigenvalue of $\widetilde{L}$.
\paragraph{Graph Convolution Networks} (GCN)~\citep{kipf2016semi} further simplify ChebNet by letting $K=2$, adding a weight sharing constraint for the first and second convolutions $\Theta_1^\ell =-\Theta_2^\ell :=\Theta^\ell$, and assuming $\lambda_{max}(\widetilde{L})\simeq 2$. This yields:
\begin{align}
    H^{\ell  +1}&=\sigma((2I-\widetilde{L})H^\ell \Theta^\ell )\\
    &=\sigma((I+D^{-1/2}WD^{-1/2})H^\ell \Theta^\ell ),\label{eq:gcn1}
\end{align}
Furthermore, since $I+D^{-1/2}WD^{-1/2}$ has eigenvalues in $[0, 2]$, applying~\cref{eq:gcn1} multiple times might lead to numerical instabilities or exploding gradients.
To overcome this issue, GCNs use a re-normalization trick, which maps the eigenvalues of $I+D^{-1/2}WD^{-1/2}$ to $[0,1]$: 
$$I+D^{-1/2}WD^{-1/2}\rightarrow ({D+I})^{-1/2}({W+I})({D+I})^{-1/2}.$$
Using \framework{} notation, GCN patch functions can be written as:
$$g_1(W,D)=({D+I})^{-1/2}({W+I})({D+I})^{-1/2},$$ 
and the graph convolution layer (see~\ref{fig:agg} for an illustration) is:
\begin{align}
  H^{\ell  +1}=\sigma(g_1(W,D)H^{\ell  }\Theta^\ell ).\label{eq:gcn}  
\end{align}
This model has been applied to many problems including matrix completion \citep{berg2017graph}, link prediction in knowledge graphs \citep{schlichtkrull2018modeling}, and unsupervised graph embedding with variational inference \citep{kipf2016variational}.

\smallskip
\smallskip
\noindent Note that in contrast with spectrum-based methods covered in the previous section, both ChebyNet and GCN do not rely on computations of the Laplacian's eigenvectors. The convolution step is only defined over the local neighborhood of each node (as defined by the adjacency matrix $W$), and therefore we can view these methods as local message passing algorithms (see the Taxonomy in~\cref{fig:taxonomy}), even though these are motivated by spectral graph theory. 
\subsection{Spatial Graph Convolutions}\label{subsec:spatial}
\textit{Spectrum-based} methods are limited by their domain dependency and cannot be applied in inductive settings.  
Furthermore, \textit{spectrum-free} methods such as GCNs require storing the entire graph adjacency matrix, which can be computationally expensive for large graphs. 

To overcome these limitations, \textit{spatial} methods borrow ideas from standard CNNs, where convolutions are applied in the spatial domain as defined by the graph topology.
For instance, in computer vision, convolutional filters are spatially localized by using fixed rectangular patches around each pixel. 
Additionally, since pixels in images have a natural ordering (top, left, bottom, right), it is possible to reuse filters' weights at every location, significantly reducing the total number of parameters. 
While such spatial convolutions cannot directly be applied in graph domains, spatial graph convolutions use ideas such as \textit{neighborhood sampling} and \textit{attention mechanisms} to overcome challenges posed by graphs' irregularities. 
\begin{figure*}
\centering
\includegraphics[width=0.8\textwidth]{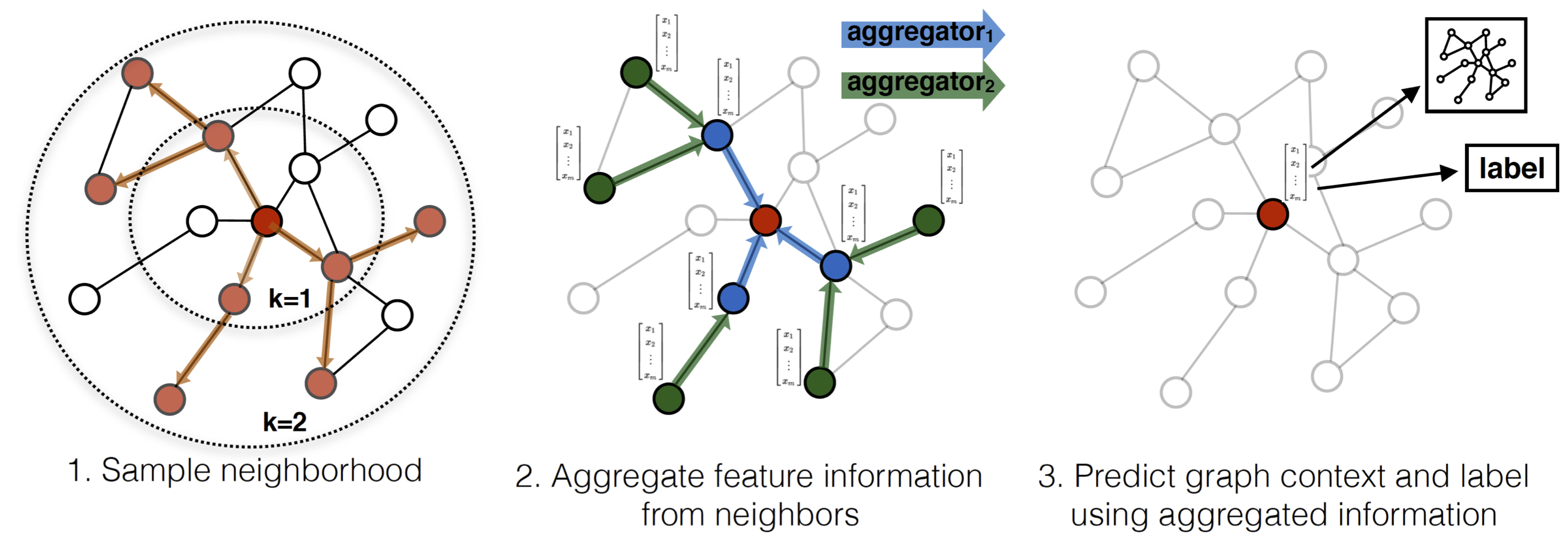}
\caption{Illustration of the GraphSAGE model. Reprinted with permission from \citep{hamilton2017inductive}.}
\label{fig:graphSage}
\end{figure*}
\subsubsection{Sampling-based spatial methods}\label{subsec:spatial_sampling}
\paragraph{Inductive representation learning on large graphs} (SAGE)~\cite{hamilton2017inductive}
While GCNs can be used in inductive settings, they were originally introduced for semi-supervised transductive settings, and the learned filters might strongly rely on the Laplacian used for training. 
Furthermore, GCNs require storing the entire graph in memory which can be computationally expensive for large graphs.

To overcome these limitations, \citet{hamilton2017inductive} propose SAGE, a general framework to learn inductive node embeddings while reducing the computational complexity of GCNs. 
Instead of averaging signals from all one-hop neighbors using multiplications with the Laplacian matrix, SAGE samples fixed neighborhoods (of size $q$) to remove the strong dependency on a fixed graph structure and generalize to new graphs. 
At every SAGE layer, nodes aggregate information from nodes sampled from their neighborhood, and the propagation rule can be written as: 
\begin{equation}
    H_{:,i}^{\ell +1}=\sigma(\Theta_1^\ell H_{:,i}^\ell + \Theta_2^\ell \mathrm{AGG}(\{H^\ell_{:,j}:j|v_j\in\mathrm{Sample}(\mathcal{N}(v_i), q)\})),
\end{equation}
where $\mathrm{AGG}(\cdot)$ is an aggregation function, which can be any permutation invariant operator such as averaging (SAGE-mean) or max-pooling (SAGE-pool).

Note that SAGE can also be described using \framework. 
For simplicity, we describe SAGE-mean using \framework{} notation, and refer to \citep{hamilton2017inductive} for details regarding other aggregation schemes. 
In \framework{} notation, SAGE-mean uses two patch learning functions with $g_1(W,D)=I$ being the identity, and $g_2(W,D)\sim\mathcal{U}_{\text{norm}}(D^{-1}W, q)$, where $\mathcal{U}_{\text{norm}}(\cdot,q)$ indicates uniformly sampling $q$ nonzero entries per row, followed by row normalization.
That is, the second patch propagates information using neighborhood sampling, and the SAGE-mean layer is:
$$H^{\ell +1}=\sigma(g_1(W,D)H^\ell \Theta_1^\ell +g_2(W,D)H^{\ell }\Theta_2^\ell ).$$

\subsubsection{Attention-based spatial methods}\label{subsec:spatial_attention}
Attention mechanisms \citep{vaswani2017attention} have been successfully used in language models, and are particularly useful when operating on long sequence inputs, they allow models to identify relevant parts of the inputs. 
Similar ideas have been applied to graph convolution networks.
Graph attention-based models learn to pay attention to important neighbors during the message passing step. 
This provides more flexibility in inductive settings, compared to methods that rely on fixed weights such as GCNs. 

Broadly speaking, attention methods learn neighbors' importance using parametric functions whose inputs are node features at the previous layer.
Using \framework, we can abstract patch functions in attention-based methods as functions of the form:
$$f_k(W, H^\ell )=\alpha(W \cdot g_k(H^\ell )),$$
where $\cdot$ indicates element-wise multiplication and $\alpha(\cdot)$ is an activation function such as softmax or ReLU.
\paragraph{Graph Attention Networks}  (GAT)~\citep{velickovic2017graph} is an attention-based version of GCNs, which incorporate self-attention mechanisms when computing patches.
At every layer, GAT attends over the neighborhood of each node and learns to selectively pick nodes which lead to the best performance for some downstream task. 
The high-level intuition is similar to SAGE \citep{hamilton2017inductive} and makes GAT suitable for inductive and transductive problems.
However, instead of limiting the convolution step to fixed size-neighborhoods as in SAGE, GAT allows each node to attend over the entirety of its neighbors and uses attention to assign different weights to different nodes in a neighborhood.  
The attention parameters are trained through backpropagation, and the GAT self-attention mechanism is: 
$$g_k(H^\ell )=\mathrm{LeakyReLU}(H^\ell B^\top b_0 \oplus b_1^\top B{H^\ell }^\top)$$
where $\oplus$ indicates summation of row and column vectors with broadcasting, and $(b_0,b_1)$ and $B$ are trainable attention weight vectors and weight matrix respectively.
The edge scores are then row normalized with softmax.
In practice, the authors propose to use multi-headed attention and combine the propagated signals with a concatenation of the average operator followed by some activation function. 
GAT can be implemented efficiently by computing the self-attention scores in parallel across edges, as well as computing the output representations in parallel across nodes. 
\paragraph{Mixture Model Networks} (MoNet)~\citet{monti2017geometric} provide a general framework that works particularly well when the node features lie in multiple domains such as 3D point clouds or meshes.
MoNet can be interpreted as an attention method as it learns patches using parametric functions in a pre-defined spatial domain (e.g. spatial coordinates), and then applies convolution filters in the graph domain.  

Note that MoNet is a generalization of previous spatial approaches such as Geodesic CNN (GCNN) \citep{masci2015geodesic} and Anisotropic CNN (ACNN) \citep{boscaini2016learning}, which both introduced constructions for convolution layers on manifolds.
However, both GCNN and ACNN use fixed patches that are defined on a specific coordinate system and therefore cannot  generalize to graph-structured data. 
The MoNet framework is more general; any pseudo-coordinates such as local graph features (e.g.  vertex degree) or manifold features ({e.g.} 3D spatial coordinates) can be used to compute the patches.
More specifically, if $U^{s}$ are pseudo-coordinates and $H^{\ell}$ are features from another domain, then using \framework{}, the MoNet layer can be expressed as: 
\begin{align}
    H^{\ell +1}=\sigma\bigg(\sum\limits_{k=1}^{K}(W\cdot g_k(U^{s}))H^{\ell }\Theta_k^\ell\bigg),
\end{align}
where $\cdot$ is element-wise multiplication and $g_k(U^{s})$ are the learned parametric patches, which are $|V|\times |V|$ matrices. 
In practice, MoNet uses Gaussian kernels to learn patches, such that:
$$g_k(U^{s})=\mathrm{exp}\bigg(-\frac{1}{2}(U^{s}-\mu_k)^\top \Sigma_k^{-1}(U^{s}-\mu_k)\bigg),$$
where $\mu_k$ and $\Sigma_k$ are learned parameters, and \citet{monti2017geometric} restrict $\Sigma_k$ to be a diagonal matrix.
\subsection{Non-Euclidean Graph Convolutions}\label{subsec:non_euclidean_sup}
Hyperbolic shallow embeddings enable embeddings of hierarchical graphs with smaller distortion than Euclidean embeddings. 
However, one major downside of shallow embeddings is that they are inherently transductive and cannot generalize to new graphs. 
On the other hand, Graph Neural Networks, which leverage node features, have achieved state-of-the-art performance on inductive graph embedding tasks. 

\begin{figure}[t!]
    \centering
    \begin{subfigure}[t]{0.45\textwidth}
        \centering
        \includegraphics[width=\linewidth]{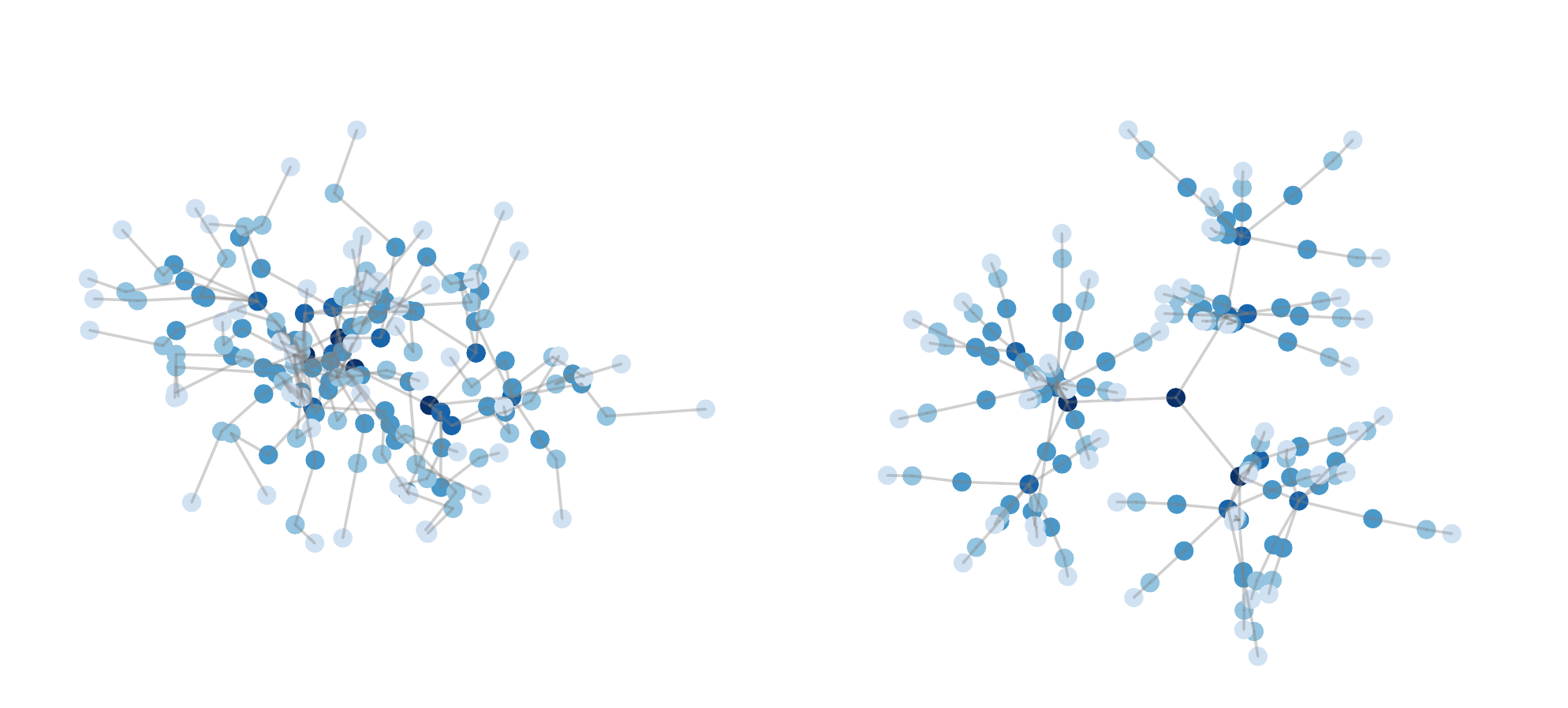}
        \caption{{GCN} layers.}\label{fig:gcn_tree}
    \end{subfigure}%
    \begin{subfigure}[t]{0.45\textwidth}
        \centering
        \includegraphics[width=\linewidth]{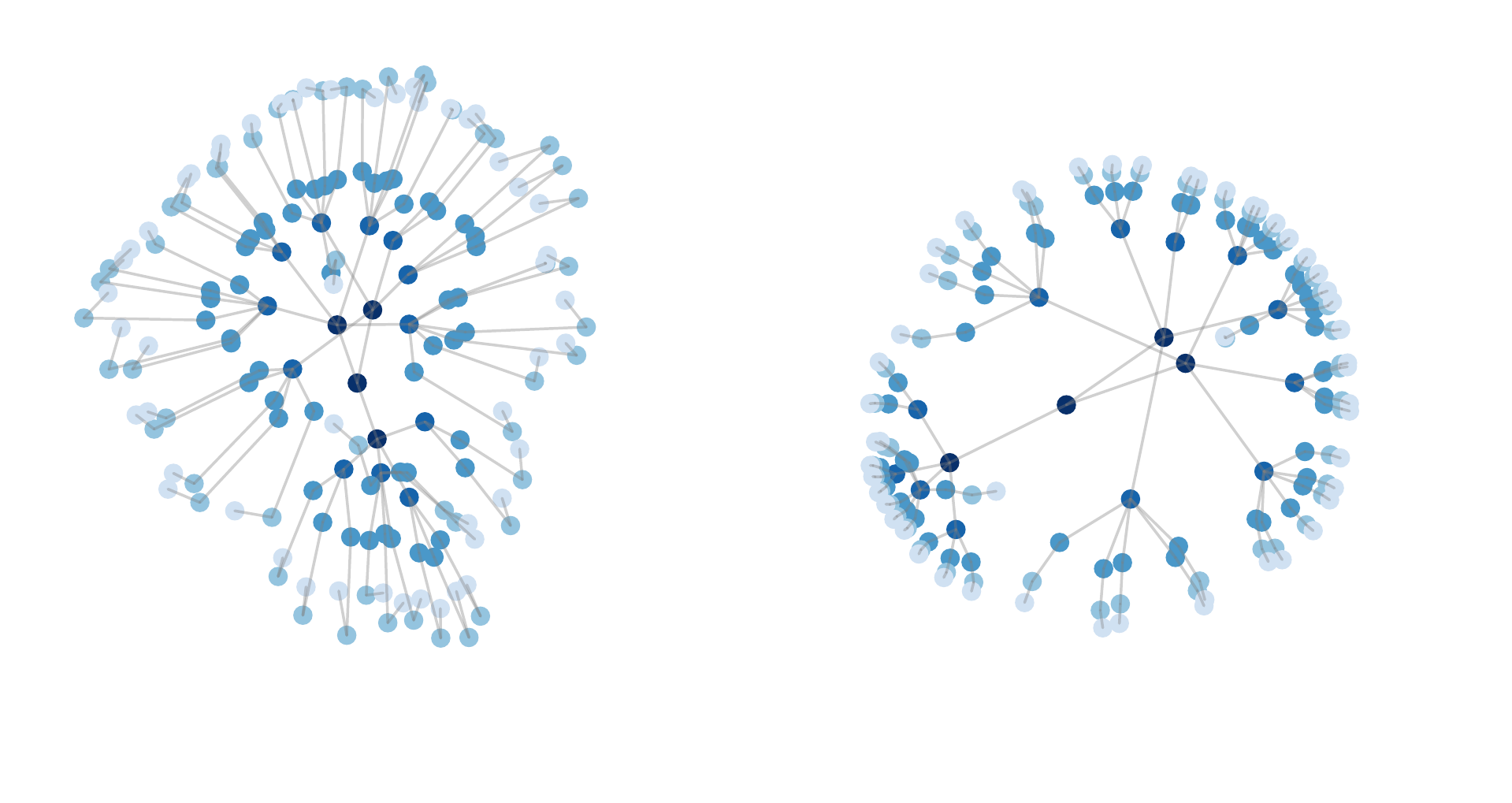}
        \caption{HGCN layers.}\label{fig:hyper_tree}
    \end{subfigure}%
    \caption{Euclidean (left) and hyperbolic (right) embeddings of a tree graph. Hyperbolic embeddings learn natural hierarchies in the embedding space (depth indicated by color). Reprinted with permission from \citep{chami2019hyperbolic}.}
    \label{fig:hgcn_viz}
\end{figure}

Recently, there has been interest in extending Graph Neural Networks to learn non-Euclidean embeddings and thus benefit from both the expressiveness of Graph Neural Networks and hyperbolic geometry. 
One major challenge in doing so is how to perform convolutions in a non-Euclidean space, where standard operations such as inner products and matrix multiplications are not defined.
\paragraph{Hyperbolic Graph Convolutional Neural Networks} (HGCN) \citep{chami2019hyperbolic} and Hyperbolic Graph Neural Networks (HGNN) \citep{liu2019hyperbolic} apply graph convolutions in hyperbolic space by leveraging the Euclidean tangent space, which provides a first-order approximation of the hyperbolic manifold at a point. 
For every graph convolution step, node embeddings are mapped to the Euclidean tangent space at the origin, where convolutions are applied, and then mapped back to the hyperbolic space. These approaches yield significant improvements on graphs that exhibit hierarchical structure~(\cref{fig:hgcn_viz}). 

\subsection{Summary of supervised graph embedding}
This section presented a number of methods that process task labels (e.g., node or graph labels) at training time. As such, model parameters are directly optimized on the upstream task.

Shallow methods use neither node features $X$ nor adjacency $W$ in the encoder (Section \ref{subsec:shallow_sup}), but utilize the adjacency to ensure consistency.
Such methods are useful in transductive settings, if only one graph is given, without node features, a fraction of nodes are labeled, and the goal is to recover labels for unlabeled nodes.

Graph regularization methods (Section \ref{subsec:reg}) utilize node features $X$ but not the adjacency $W$ in the encoder. In general, these methods are inductive (except of one version of planetoid \citep{yang2016revisiting}). In fact, they need only a graph at training time but not at inference (test) time. In general, they can be applied when node features contain rich information.

Finally, graph convolution models (Sections \ref{subsec:gc_framework}, \ref{subsec:spectral} \& \ref{subsec:spatial}) utilize node features and adjacency in the encoder. At the time of writing, these models acheive superior empirical performance on many node-classification tasks.
\section{Applications}\label{sec:applications}
Graph representation learning methods can be applied to a wide range of applications, which can be unsupervised or supervised.  
In unsupervised applications,
task-specific labels are not processed for learning embeddings. Rather, the graph is used as a form of self-supervision. Specifically, one can learn embeddings that preserve the graph (i.e. neighborhoods) or to preserve structural equivalence of nodes (see \cref{subsec:pos_struct} for distinction), for instance, by
applying
unsupervised embedding methods~(\cref{sec:unsup_models}, upper branch of the Taxonomy in~\cref{fig:taxonomy}).
On the other hand, in supervised applications, node embeddings are directly optimized for some specific task, such as classifying nodes or graphs. 
In this setting, supervised embedding methods~(\cref{sec:sup_models}, lower branch of the Taxonomy in~\cref{fig:taxonomy}) can be applied.
Table \ref{table:applications} summarizes some popular tasks in GRL, and pairs them with methods frequently used for the tasks.
We review common unsupervised and supervised graph applications next. 

\begin{table}
\vspace{-1.1cm}
\hspace*{\dimexpr -\oddsidemargin-0.5in}
\begin{minipage}{\dimexpr \paperwidth-1in}
\centering
\hspace*{-2.2cm}\begin{tabular}{r p{5cm} | c | c | c}
\hline
&
\multirow{2}{*}{\bfseries Method} &  
\multicolumn{2}{c|}{\bfseries Training complexity} &
\multirow{2}{*}{\bfseries Training input}
\\
& &  Memory & Computation & 
\\
\hline

\multirow{2}{*}{\textbf{(a)}} &
\multirow{2}{*}{DeepWalk \citepalias{perozzi2014deepwalk}} &
\multirow{2}{*}{$\mathcal{O}(|V|d)$} &
\multirow{2}{*}{$\mathcal{O}(c^2 d |V| \log_2 |V|)$} &
\multirow{12}{*}{$W$}
\\

 & & & 
\\
\cline{1-4}

\multirow{2}{*}{\textbf{(b)}} &
\multirow{2}{*}{node2vec \citepalias{grover2016node2vec}} &
\multirow{2}{*}{$\mathcal{O}(|V|d)$} &
\multirow{2}{*}{$\mathcal{O}(c^2 d |V|)$}  &
\\

 &
 &
 &
\\

\cline{1-4}

\multirow{3}{*}{\textbf{(c)}} &
LINE \citepalias{tang2015line} \newline 
HOPE \citepalias{ou2016asymmetric} \newline 
GF  \citepalias{ahmed2013distributed} &
\multirow{3}{*}{$\mathcal{O}(|V|d)$} &
\multirow{3}{*}{$\mathcal{O}(|E|d)$}  &
\\

\cline{1-4}
\multirow{2}{*}{\textbf{(d)}} &
SDNE \citepalias{wang2016structural} \newline DNGR \citepalias{cao2016deep}   &
\multirow{2}{*}{$\mathcal{O}(|V|b \mathcal{D})$} &
\multirow{2}{*}{$\mathcal{O}(|V|b \mathcal{M})$} &
 \\
 \cline{1-4}
\multirow{2}{*}{\textbf{(e)}} & 
GraRep \citepalias{cao2015grarep} \newline
WYS \citepalias{abu2017watch} &
\multirow{2}{*}{$\mathcal{O}(|V|^2)$} &
\multirow{2}{*}{$\mathcal{O}(|V|^3c + |V|^2d)$} &
\\
\cline{1-4}

\hline
\textbf{(f)} &
HARP \citepalias{chen2017harp} &
\multicolumn{2}{c|}{ \textit{inherits}} &
$W$ 
\\
\hline
\textbf{(g)} &
Splitter \citepalias{epasto2019splitter} &
\multicolumn{2}{c|}{ \textit{inherits}} &
$W$ 
\\
\hline
\multirow{1}{*}{\textbf{(h)}} &
\multirow{1}{*}{MDS \citepalias{kruskal1964multidimensional}} &
$\mathcal{O}(|V|^2)$ &
$\mathcal{O}(|V|^3)$ &
\multirow{3}{*}{\parbox{2.4cm}{$X$ induces $W$}} 
\\
\cline{1-4}
\multirow{2}{*}{\textbf{(i)}} &
LP \citepalias{zhu2002learning} \newline 
LLE \citepalias{roweis2000nonlinear} &
\multirow{2}{*}{$\mathcal{O}(|V|)$} &
\multirow{2}{*}{$\mathcal{O}(|E| \times \textrm{iters})$}  & 
\\

\hline
\textbf{(j)} &
GNN Methods &
$\mathcal{O}(|V| \mathcal{D})$ &
$\mathcal{O}(|E| \mathcal{D} + |V| \mathcal{M} )$ &
$X, W$ 
\\

\hline
\textbf{(k)} &
SAGE \citepalias{hamilton2017representation} &
$\mathcal{O}(bF^{H} \mathcal{D})$ &
$\mathcal{O}(bF^{H-1} \mathcal{D} + bF^{H} \mathcal{M})$ &
$X, W$ 
\\

\hline
\textbf{(l)} &
GTTF \citepalias{markowitz2021graph} &
$\mathcal{O}(bF^{H} \mathcal{D})$ &
$\mathcal{O}(bF^{H-1} \mathcal{D} + bF^{H} \mathcal{M})$ &
$X, W$ 
\\

\hline

\end{tabular}\hspace*{-2.2cm}
\vspace{-0.3cm}
\caption{%
\small
Summary and practical implications of GRL methods.
Columns from \textbf{right-to-left}: practical scenarios where methods have demonstrated value; the input to the methods: (adjacency matrix $W$, node features $X$, or both); the hardware cost to train the method; finally, the left-most columns indicate the method classes. We derive the \textbf{Training Complexity} as follows. The method classes \textbf{(a-h)} are node embedding methods where $c \in \mathbb{Z}$ denotes \textit{context size} (e.g. length of random walk) and $d$ denotes size of embedding dictionary.
\textbf{(a)} DeepWalk and \textbf{(b)} node2vec store the embedding dictionary (with $\mathcal{O}(|V|d)$ floating-point entries); Training simulates from every node $\in V$ a fixed number of walks each with fixed length: the dot products of all node-pairs within window of size $c$ along the simulated walks are calculated. 
For each pair either the hierarchical softmax (a) or negative sampling (b) is used.
Both left-most $|V|$ terms can be substituted by batch size $b$ to analyze per-batch complexity. However, we analyze per epoch for simplicity.
\textbf{(c)}
LINE \citepalias{tang2015line}, HOPE \citepalias{ou2016asymmetric} and GF  \citepalias{ahmed2013distributed} loop-over all edges. 
%
\textbf{(d)} SDNE and DNGR train auto-encoders over the adjacency matrix, with batch-size $b$, and $\mathcal{D} = \sum_\ell d_\ell$ denotes total dimensions of all layers. $\mathcal{M} = \sum_\ell d_\ell d_{\ell+1}$ accounts for floating-point ops of matrix multiplications.
For full-batch, $b = |V|$.
\textbf{(e)} GraRep and WYS raise transition matrix to power $c$ storing a dense square matrix with $\mathcal{O}(|V|^2)$ non-zeros.
\textbf{(f)} HARP \citepalias{chen2017harp} and
\textbf{(g)} Splitter can run any algorithm (e.g., \textbf{(a-e)}), therefore their complexity depends on the specific underlying algorithm. Here we assume that number of times HARP is invoked (i.e. \textit{scales of the graph}), and the average number of personas per node for Splitter, are both small ($\ll |V|$).
\textbf{(h)} MDS  computes all-pairs similarity and
LE  requires full eigendecomposition of the graph laplacian matrix (to recover the eigenvectors corresponding \textit{smallest} eigenvalues).
\textbf{(i)} LP and LLE  loop over edges up-to ``iters'' iterations, assuming that number of label classes is small.
\textbf{(j)} contain graph convolution methods of GCN \citepalias{kipf2016semi}, GAT \citepalias{velickovic2017graph},
MixHop \citepalias{abu2019mixhop}, GIN \citepalias{xu2018powerful}, 
GGNN \citepalias{li2015gated}, MPNN \citepalias{gilmer2017neural},  
ChebNet \citepalias{defferrard2016convolutional} and MoNet \citepalias{monti2017geometric}.
We assume the \textit{naive} full-batch implementation provided by authors of those methods. At each layer, every node averages its neighbors (total of $|E|\mathcal{D}$ floating-point operations) followed by multiplying by the layer filter (total of $|V|\mathcal{M}$ floating-point ops). Finally, to scale learning to larger graphs, sampling methods like \textbf{(k-l)}
reduce the hardware requirement of training algorithm,
seperating memory complexity from graph size.
\textbf{(k)} SAGE and
\textbf{(l)} GTTF
sample $F$ nodes for every node in batch (of size $b$), and $F$ of their neighbors, and so on, until tree height reaches $H$. For both \textbf{(k)} and \textbf{(l}, we ignore runtime complexity of data pre-processing, as it is to be computed once per graph, irrespective of the number of (hyperparameter) sweep runs.
}
\label{table:applications}
\end{minipage}
\end{table}

\subsection{Unsupervised applications}\label{section:unsup_taks}
\subsubsection{Graph reconstruction}
The most standard unsupervised graph application is graph reconstruction.
In this setting, the goal is to learn mapping functions (which can be parametric or not) that map nodes to dense distributed representations which preserve  graph properties such as node similarity.
Graph reconstruction doesn't require any supervision and models can be trained by minimizing a reconstruction error, which is the error in recovering the original graph from learned embeddings.
Several algorithms were designed specifically for this task, and we refer to~\cref{sec:unsup_models} for some examples of reconstruction objectives.
At a high level, graph reconstruction is similar to dimensionality reduction in the sense that the main goal is summarize some input data into a low-dimensional embedding.
Instead of compressing high dimensional vectors into low-dimensional ones as standard dimensionality reduction methods (e.g. PCA) do, the goal of graph reconstruction models is to compress data defined on graphs into low-dimensional vectors.  
%
\subsubsection{Link prediction}
Link prediction is the task of predicting links in a graph. 
In other words, the goal in link prediction tasks is to predict missing or unobserved links ({e.g.} links that may appear in the future for dynamic and temporal networks). 
Link prediction can also help identifying spurious link and remove them.
It is a major application of graph learning models in industry, and common example of applications include predicting friendships in social networks or predicting user-product interactions in recommendation systems. 

A common approach for training link prediction models is to mask some edges in the graph (positive and negative edges), train a model with the remaining edges and then test it on the masked set of edges. 
Note that link prediction is different from graph reconstruction. 
In link prediction, we aim at predicting links that are not observed in the original graph while in graph reconstruction, we only want to compute embeddings that preserve the graph structure through reconstruction error minimization. 

Finally, while link prediction has similarities with supervised tasks in the sense that we have labels for edges (positive, negative, unobserved), we group it under the unsupervised class of applications since edge labels are usually not used during training, but only used to measure the predictive quality of embeddings.
That is, models described in~\cref{sec:unsup_models} can be applied to the link prediction problem. 

\subsubsection{Clustering}
Clustering is particularly useful for discovering communities and has many real-world applications.
For instance, clusters exist in biological networks (e.g. as groups of proteins with similar properties), or in social networks (e.g. as groups of people with similar interests). 

Note that unsupervised methods introduced in this survey can be used to solve clustering problems: one can run a clustering algorithm (e.g. k-means) on embeddings that are output by an encoder.
Further, clustering can be joined with the learning algorithm while learning a shallow \citep{rozemberczki2019gemsec} or Graph Convolution \citep{chiang2019clustergcn,chen2019supervised} embedding model.

\subsubsection{Visualization}
There are many off-the-shelf tools for mapping graph nodes onto two-dimensional manifolds for the purpose of visualization.
Visualizations allow network scientists to qualitatively understand graph properties, understand relationships between nodes or visualize node clusters. 
Among the popular tools are methods based on \textit{Force-Directed Layouts}, with various web-app Javascript implementations.

Unsupervised graph embedding methods are also used for visualization purposes: by first training an encoder-decoder model (corresponding to a shallow embedding or graph convolution network), and then mapping every node representation onto a two-dimensional space using, t-distributed stochastic neighbor embeddings (t-SNE) \citep{maaten2008visualizing} or PCA \citep{jolliffe2011principal}.
Such a process (embedding $\rightarrow$ dimensionality reduction) is commonly used to qualitatively evaluate the performance of graph learning algorithms.
If nodes have attributes, one can use these attributes to color the nodes on 2D visualization plots.
Good embedding algorithms embed nodes that have similar attributes nearby in the embedding space, as demonstrated in visualizations of various methods \citep{perozzi2014deepwalk, kipf2016semi, abu2017watch}.
Finally, beyond mapping every node to a 2D coordinate, methods which map every graph to a representation \citep{alrfou2019ddgk} can similarly be projected into two dimensions to visualize and qualitatively analyze graph-level properties. 

\subsection{Supervised applications}\label{section:sup_taks}
\subsubsection{Node classification}\label{sec:nc}
Node classification is an important supervised graph application, where the goal is to learn node representations that can accurately predict node labels. 
For instance, node labels could be scientific topics in citation networks, or gender and other attributes in social networks. 

Since labelling large graphs can be time-consuming and expensive, semi-supervised node classification is a particularly common application.
In semi-supervised settings, only a small fraction of nodes is labelled and the goal is to leverage links between nodes to predict attributes of unlabelled nodes.
This setting is transductive since there is only one partially labelled fixed graph.
It is also possible to do inductive node classification, which corresponds to the task of classifying nodes in multiple graphs.

Note that node features can significantly boost the performance on node classification tasks if these are descriptive for the target label.
Indeed, recent methods such as GCN \citep{kipf2016semi} or GraphSAGE \citep{hamilton2017inductive} have achieved state-of-the-art performance on multiple node classification benchmarks due to their ability to combine structural information and semantics coming from features.
On the other hand, other methods such as random walks on graphs fail to leverage feature information and therefore achieve lower performance on these tasks. 

\subsubsection{Graph classification}\label{sec:gc}
Graph classification is a supervised application
the task is to predict graph-level labels given an input  graph.
Graph classification tasks are inherently inductive, as new graphs are presented at test time.
Many popular tasks are biochemical and some others are online social networks. In the biochemical domain, a common application uses graphs corresponding to molecules.
In these graphs, each node represents an atom (e.g.\ with a feature vector that's a 1-hot encoding of its atomic number) and an edge between two nodes indicates a bond (feature vector could indicate bond type).
The graph-level label is task dependant, e.g., indicating mutagenicity of a drug against bacteria, such as MUTANG \citep{debnath1991structure}.
In online social networks, nodes usually correspond to users and edges represent relationships or interactions. For instance, the Reddit graph classification tasks \citep{yanardag2015deep} contain many graphs. Each graph corresponds to a discussion thread: user commenting on a user's comment, an edge will connect the two. The goal is to predict the community (\textit{sub-reddit}) where discussion took place, given the graph of comments.

Different than tasks of node-level (e.g., node classification) and edge-level (e.g., link prediction) prediction, graph classification tasks require an additional type of pooling, in order to aggregate node-level information into graph-level information.
As discussed earlier, generalizing this notion of pooling to arbitrary graphs is non-trivial, and is an active research area.
The pooling function should be invariant to the node order.
Many methods use simple pooling, such as mean or sum of all graph node-level latent vectors e.g. \citep{xu2018powerful}.
Other methods use differentiable pooling \citep{ying2018hierarchical, cangea2018towards, goa2019graphunets, lee2019selfattention}.

In addition to these supervised methods, a number of unsupervised methods for learning graph-level representations have been proposed \citep{tsitsulin2018netlsd, alrfou2019ddgk,tsitsulin2020slaq}. In fact, a notable class of unsupervised graph-level models are known as \textbf{graph kernels} (GKs), see \citep{vishwanathan2010graphkernels, krieg2020survey} for reviews.

While GKs are outside our main focus,
here we briefly mention connections of GKs to \name.
GKs can be applied to graph-level tasks such as graph classification.
GK can implicitly implement a similarity function that maps any two graphs into a scalar.
Traditional GKs compute similarity of two graphs by counting how many walks (or paths) the two graphs share in common -- e.g., each walk can be encoded as a sequence of node labels. If nodes are not labeled, it is common to use the node degrees as labels.
GKs are often analyzed in their ability to detect (sub-)graph isomorphism. Two (sub-)graphs are isomorphic if they are identical when ignoring node ordering.
As sub-graph isomorphism is NP-hard, the $1$-dimensional Weisfeiler-Leman ($1$-WL) heuristic
deems two sub-graphs as isomorphic as follows. For each graph, node statistics are counted as histograms (e.g., count nodes with label ``A'', and how many of those have an edge to nodes with label ``B'', etc).
The $1$-WL heuristic deems two graphs as isomorphic if their histograms, extracted from 1-hop neighborhood, are identical.
Certain GNNs, such as the Graph Isomorphism Network \citep[GIN,][]{xu2018powerful}
have been proven to realize the $1$-WL heuristic i.e. it can map two graphs to the same latent vector if-and-only-if they would be deemed isomorphic by the $1$-WL heuristic.
Some recent work combines GKs and GNNs. As examples, \citet{chen2020cgkn}
extract walk patterns; \citet{du2019gntk}
model similarity of two graphs using the similarity of the ``tangent space'' of the objective w.r.t. the Gaussian-initialized GNN parameters. In both \citep{du2019gntk, chen2020cgkn}, there is no actual GNN training.
The training rather uses kernelized methods such as kernel support vector machines, on the pairwise Gram matrix.
As such, these methods cannot be readily plugged into our our frameworks of \framework\, or \name. 
On other hand, other methods explicitly map a graph to the high-dimensional latent space, rather than implicitly compute graph-to-graph similarity scalar score. As an example, the 
$k$-GNN network of
\citet{morris2019weisfeiler} can realize the $k$-WL heuristic (similar to 1-WL, but here histograms are computed up-to $k$-hop neighbors), yet it is explicitly programmed as a GNN. As such, the $k$-GNN model classes can be described in our frameworks of \framework~and \name.

%
\section{Conclusion and Open Research Directions}\label{sec:conc}
In this survey, we introduced a unified framework to compare machine learning models for graph-structured data. 
We presented a generalized \name{} framework, previously applied to unsupervised network embedding, that encapsulates shallow graph embedding methods, graph auto-encoders, graph regularization methods and graph neural networks. 
We also introduced a graph convolution framework (\framework), which is used to describe and compare convolution-based graph neural networks, including spatial and spectral graph convolutions. 
Using this framework, we introduced a comprehensive taxonomy of GRL methods, encapsulating over thirty methods for graph embedding (both supervised and unsupervised). 

We hope that this survey will help and encourage future research in GRL, to hopefully solve the challenges that these models are currently facing. 
In particular, practitioners can reference the taxonomy to better understand the available tools and applications, and easily identify the best method for a given problem. 
Additionally, researchers with new research questions can use the taxonomy to better classify their research questions, reference the existing work, identify the right baselines to compare to, and find the appropriate tools to answer their questions.

While GRL methods have achieved state-of-the-art performance on node classification or link prediction tasks, many challenges remain unsolved.
Next, we discuss ongoing research directions and challenges that graph embedding models are facing. 

\paragraph{Evaluation and benchmarks}
The methods covered in this survey are typically evaluated using standard node classification or link prediction benchmarks. 
For instance, citation networks are very often used as benchmarks to evaluate graph embedding methods. 
However, these small citation benchmarks have drawbacks since results might significantly vary based on datasets' splits, or training procedures (e.g. early stopping), as shown in recent work \citep{shchur2018pitfalls}. 

To better advance GRL methods, it is important to use robust and unified evaluation protocols, and evaluate these methods beyond small node classification and link prediction benchmarks. 
Recently, there has been progress in this direction, including new graph benchmarks with leaderboards~\citep{ogb_2020,dwivedi2020benchmarking} and graph embedding libraries~\citep{fey2019fast,wang2019deep,goyal2018gem}.
On a similar vein, \citet{sinha2020evaluating} recently proposed a set of tasks grounded in first-order logic, to evaluate the reasoning capabilities of GNNs. 

\paragraph{Fairness in Graph Learning}
The emerging field of Fairness in Machine Learning seeks to ensure that models avoid correlation between `sensitive' features and the model's predicted output \citep{mehrabi2019survey}.
These concerns can be especially relevant for graph learning problems, where we must also consider the correlation of the graph structure (the edges) in addition to the feature vectors of the nodes with the final output.

The most popular technique for adding fairness constraints to models relies on using adversarial learning to debias the model's predictions relative to the sensitive feature(s), and can be extended to GRL \citep{bose2019compositional}.
However, adversarial methods do not offer strong guarantees about the actual amount of bias removed.
In addition, many debiasing methods may not be effective at the debiasing task in practice \citep{gonen2019lipstick}.
Recent work in the area aims to provide provable guarantees for debiasing GRL  \citep{palowitch2019monet}.

\paragraph{Application to large and realistic graphs}
Most learning methods on graphs are applied only on smaller datasets, with sizes of up to hundred of thousands of nodes.
However, many real-world graphs are much larger, containing up to billions of nodes.
Methods that scale for large graphs \citep{lerer2019biggraph, ying2018pinsage} require a Distributed Systems setup with many machines, such as MapReduce \citep{dean08mapreduce}.
Given a large graph that fits on a single hard disk (e.g. with one terabyte size) but does not fit on RAM, how can a researcher apply a learning method on such a large graphs, using just a personal computer?
Contrast this with a computer vision task by considering a large image dataset \citep{deng2009imagenet, kuznetsova2020openimages}. 
It is possible to train such models on personal computers, as long as the model can fit on RAM, regardless how large the dataset is.
This problem may be particularly challenging for graph embedding models, especially those which have parameters that scale with the number of nodes in the graph.

Sometimes in industry, even choosing the best graph to use as input is difficult.  \cite{halcrow-grale} describes Grale, a system at Google used for learning the correct graph from a variety of different features.  Grale relies on techniques from similarity search (like locality sensitive hashing) to scale graph learning to extremely large datasets.  Recent work extends the Grale model with an attention network \citep{rozemberczki-pathfinder} to allow end-to-end learning.

We foresee additional engineering and mathematical challenges in learning methods for large graphs, while still being operable on a single machine. We hope that researchers can focus on this direction to expose such learning tools to non-expert practitioners, such as a Neurologist wishing to analyze the sub-graph of the human brain given its neurons and synapses, stored as nodes and edges.

\paragraph{Molecule generation}
Learning on graphs has a great potential for helping molecular scientists to reduce cost and time in the laboratory. Researchers proposed methods for predicting quantum properties of molecules \citep{gilmer2017neural, duvenaud2015convolutional} and for generating molecules with some desired properties \citep{liu2018constrained, de2018molgan,li2018learning,simonovsky2018graphvae,you2018graphrnn}. A review of recent methods can be found in \citep{elton2019molecular}. Many of these methods are concerned with manufacturing materials with certain properties (e.g. conductance and malleability), and others are concerned drug design \citep{jin2018junction,Ragoza2017proteinligand,feng2018padme}. 

\paragraph{Combinatorial optimization}
Computationally hard problems arise in a broad range of areas including routing science, cryptography, decision making and planning. 
Broadly speaking, a problem is computationally hard when the algorithms that compute the optimal solution scale poorly with the problem size.
There has been recent interest in leveraging machine learning techniques (e.g. reinforcement learning) to solve combinatorial optimization problems and we refer to~\citep{bengio2018machine} for a review of these methods. 

Many hard problems (e.g. SAT, vertex cover...) can be expressed in terms of graphs and more recently, there has been interest in leveraging graph embeddings to approximate solutions of NP-hard problems~\citep{khalil2017learning,nowak2017revised,selsam2018learning,prates2019learning}. 
These methods tackle computationally hard problems from a data-driven perspective, where given multiple instances of a problem, the task is to predict whether a particular instance (e.g. node) belongs to the optimal solution. 
Other work focuses on optimizing graph partitions \citep{bianchi2020spectral,tsitsulin2020graph}, finding assignments that aim to fulfill an objective (e.g.\ the minimum conductance cut). 

One motivation for all these approaches is the relational inductive biases found in GNNs which enable them to better represent graphs compared to standard neural networks (e.g. permutation invariance). 
While these data-driven methods are still outperformed by existing solvers, promising results show that GNNs can generalize to larger problem instances \citep{nowak2017revised,prates2019learning}.
We refer to the recent survey on neural symbolic learning by~\citet{lamb2020graph} for an extensive review of GNN-based methods for combinatorial optimization. 

\paragraph{Non-Euclidean embeddings}
As we saw in~\cref{subsec:non_euclidean_unsup} and~\cref{subsec:non_euclidean_sup}, an important aspect of graph embeddings is the underlying space geometry. 
Graphs are discrete, high-dimensional, non-Euclidean structures, and there is no straightforward way to encode this information into low-dimensional Euclidean embeddings that preserve the graph topology \citep{bronstein2017geometric}. 
Recently, there has been interest and progress into learning non-Euclidean embeddings such as hyperbolic \citep{nickel2017poincare} or mixed-product space \citep{gu2018learning} embeddings. 
These non-Euclidean embeddings provide a promise for more expressive embeddings, compared to Euclidean embeddings.
For instance, hyperbolic embeddings can represent hierarchical data with much smaller distortion than Euclidean embeddings~\citep{sarkar2011low} and have led to state-of-the-art results in many modern applications such as link prediction in knowledge graphs~\citep{balazevic2019multi,chami2020low} and linguistics tasks~\citep{tifrea2018poincare,le2019inferring}. 

Two common challenges arise with non-Euclidean embeddings: precision issues (e.g. near the boundary of the Poincar\'e ball) in hyperbolic space~\citep{de2018representation,yu2019numerically} and challenging Riemannian optimization~\citep{bonnabel2013stochastic,becigneul2018riemannian}.
Additionally, it is also unclear how to pick the right geometry for a given input graph. 
While there exists some discrete measures for the tree-likeliness of graphs (e.g. Gromov's four-point condition \citep{jonckheere2008scaled} and others~\citep{abu2016metric,chen2013hyperbolicity,adcock2013tree}), an interesting open research direction is how to pick or learn the right geometry for a given discrete graph. 

\paragraph{Theoretical guarantees}
There have been significant advances in the design of graph embedding models, which improved over the state-of-the-art in many applications. 
However, there is still limited understanding about theoretical guarantees and limitations of graph embedding models. 
Understanding the representational power of GNNs is a nascent area of research, and recent works adapt existing results from learning theory to the problem of GRL \citep{chen2019equivalence,garg2020generalization,loukas2019graph,maron2018invariant,morris2019weisfeiler,verma2019stability,xu2018powerful}.
The development of theoretical frameworks is critical to pursue in order to understand the theoretical guarantees and limitations of graph embedding methods.

\small
\section*{Acknowledgements}
\normalsize
We thank Meissane Chami, Aram Galstyan, Megan Leszczynski, John Palowitch, Laurel Orr, and Nimit Sohoni for their helpful feedback and discussions.
We also thank Carlo Vittorio Cannistraci, Thomas Kipf, Luis Lamb, Bruno Ribeiro and Petar Veli{\v{c}}kovi{\'{c}} for their helpful feedback and comments on the first version of this work.
We gratefully acknowledge the support of DARPA under Nos. FA86501827865 (SDH) and FA86501827882 (ASED); NIH under No. U54EB020405 (Mobilize), NSF under Nos. CCF1763315 (Beyond Sparsity), CCF1563078 (Volume to Velocity), and 1937301 (RTML); ONR under No. N000141712266 (Unifying Weak Supervision); the Moore Foundation, NXP, Xilinx, LETI-CEA, Intel, IBM, Microsoft, NEC, Toshiba, TSMC, ARM, Hitachi, BASF, Accenture, Ericsson, Qualcomm, Analog Devices, the Okawa Foundation, American Family Insurance, Google Cloud, Swiss Re, the HAI-AWS Cloud Credits for Research program, TOTAL, and members of the Stanford DAWN project: Teradata, Facebook, Google, Ant Financial, NEC, VMWare, and Infosys. The U.S. Government is authorized to reproduce and distribute reprints for Governmental purposes notwithstanding any copyright notation thereon. Any opinions, findings, and conclusions or recommendations expressed in this material are those of the authors and do not necessarily reflect the views, policies, or endorsements, either expressed or implied, of DARPA, NIH, ONR, or the U.S. Government.
\small

\newpage

\bibliographystyle{plainnat}
\bibliography{ref}

\begin{thebibliography}{169}
\providecommand{\natexlab}[1]{#1}
\providecommand{\url}[1]{\texttt{#1}}
\expandafter\ifx\csname urlstyle\endcsname\relax
  \providecommand{\doi}[1]{doi: #1}\else
  \providecommand{\doi}{doi: \begingroup \urlstyle{rm}\Url}\fi

\bibitem[Abu-Ata and Dragan(2016)]{abu2016metric}
Muad Abu-Ata and Feodor~F Dragan.
\newblock Metric tree-like structures in real-world networks: an empirical
  study.
\newblock \emph{Networks}, 67\penalty0 (1):\penalty0 49--68, 2016.

\bibitem[Abu-El-Haija et~al.(2017)Abu-El-Haija, Perozzi, and
  Al-Rfou]{haija2017lowrank}
Sami Abu-El-Haija, Bryan Perozzi, and Rami Al-Rfou.
\newblock Learning edge representations via low-rank asymmetric projections.
\newblock In \emph{Proceedings of the 2017 ACM on Conference on Information and
  Knowledge Management}, CIKM ’17, page 1787–1796, 2017.

\bibitem[Abu-El-Haija et~al.(2018)Abu-El-Haija, Perozzi, Al-Rfou, and
  Alemi]{abu2017watch}
Sami Abu-El-Haija, Bryan Perozzi, Rami Al-Rfou, and Alexander~A Alemi.
\newblock Watch your step: Learning node embeddings via graph attention.
\newblock In \emph{Advances in Neural Information Processing Systems}, pages
  9180--9190, 2018.

\bibitem[Abu-El-Haija et~al.(2019)Abu-El-Haija, Perozzi, Kapoor, Alipourfard,
  Lerman, Harutyunyan, Ver~Steeg, and Galstyan]{abu2019mixhop}
Sami Abu-El-Haija, Bryan Perozzi, Amol Kapoor, Nazanin Alipourfard, Kristina
  Lerman, Hrayr Harutyunyan, Greg Ver~Steeg, and Aram Galstyan.
\newblock Mixhop: Higher-order graph convolutional architectures via sparsified
  neighborhood mixing.
\newblock In \emph{International Conference on Machine Learning}, pages 21--29,
  2019.

\bibitem[Adcock et~al.(2013)Adcock, Sullivan, and Mahoney]{adcock2013tree}
Aaron~B Adcock, Blair~D Sullivan, and Michael~W Mahoney.
\newblock Tree-like structure in large social and information networks.
\newblock In \emph{2013 IEEE 13th International Conference on Data Mining},
  pages 1--10. IEEE, 2013.

\bibitem[Ahmed et~al.(2013)Ahmed, Shervashidze, Narayanamurthy, Josifovski, and
  Smola]{ahmed2013distributed}
Amr Ahmed, Nino Shervashidze, Shravan Narayanamurthy, Vanja Josifovski, and
  Alexander~J Smola.
\newblock Distributed large-scale natural graph factorization.
\newblock In \emph{Proceedings of the 22nd international conference on World
  Wide Web}, pages 37--48. ACM, 2013.

\bibitem[Al-Rfou et~al.(2019)Al-Rfou, Zelle, and Perozzi]{alrfou2019ddgk}
Rami Al-Rfou, Dustin Zelle, and Bryan Perozzi.
\newblock Ddgk: Learning graph representations for deep divergence graph
  kernels.
\newblock \emph{Proceedings of the 2019 World Wide Web Conference on World Wide
  Web}, 2019.

\bibitem[Alanis-Lobato et~al.(2016)Alanis-Lobato, Mier, and
  Andrade-Navarro]{alanis2016efficient}
Gregorio Alanis-Lobato, Pablo Mier, and Miguel~A Andrade-Navarro.
\newblock Efficient embedding of complex networks to hyperbolic space via their
  laplacian.
\newblock \emph{Scientific reports}, 6:\penalty0 30108, 2016.

\bibitem[Almeida(1987)]{almeida1987learning}
Luis~B Almeida.
\newblock A learning rule for asynchronous perceptrons with feedback in a
  combinatorial environment.
\newblock In \emph{Proceedings, 1st First International Conference on Neural
  Networks}, volume~2, pages 609--618. IEEE, 1987.

\bibitem[Balazevic et~al.(2019)Balazevic, Allen, and
  Hospedales]{balazevic2019multi}
Ivana Balazevic, Carl Allen, and Timothy Hospedales.
\newblock Multi-relational poincar{\'e} graph embeddings.
\newblock In \emph{Advances in Neural Information Processing Systems}, pages
  4463--4473, 2019.

\bibitem[Battaglia et~al.(2016)Battaglia, Pascanu, Lai, Rezende,
  et~al.]{battaglia2016interaction}
Peter Battaglia, Razvan Pascanu, Matthew Lai, Danilo~Jimenez Rezende, et~al.
\newblock Interaction networks for learning about objects, relations and
  physics.
\newblock In \emph{Advances in Neural Information Processing Systems}, pages
  4502--4510, 2016.

\bibitem[Battaglia et~al.(2018)Battaglia, Hamrick, Bapst, Sanchez-Gonzalez,
  Zambaldi, Malinowski, Tacchetti, Raposo, Santoro, Faulkner,
  et~al.]{battaglia2018relational}
Peter~W Battaglia, Jessica~B Hamrick, Victor Bapst, Alvaro Sanchez-Gonzalez,
  Vinicius Zambaldi, Mateusz Malinowski, Andrea Tacchetti, David Raposo, Adam
  Santoro, Ryan Faulkner, et~al.
\newblock Relational inductive biases, deep learning, and graph networks.
\newblock \emph{arXiv preprint arXiv:1806.01261}, 2018.

\bibitem[Becigneul and Ganea(2018)]{becigneul2018riemannian}
Gary Becigneul and Octavian-Eugen Ganea.
\newblock Riemannian adaptive optimization methods.
\newblock In \emph{International Conference on Learning Representations}, 2018.

\bibitem[Belkin and Niyogi(2002)]{belkin2002laplacian}
Mikhail Belkin and Partha Niyogi.
\newblock Laplacian eigenmaps and spectral techniques for embedding and
  clustering.
\newblock In \emph{Advances in neural information processing systems}, pages
  585--591, 2002.

\bibitem[Belkin and Niyogi(2004)]{belkin2004semi}
Mikhail Belkin and Partha Niyogi.
\newblock Semi-supervised learning on riemannian manifolds.
\newblock \emph{Machine learning}, 56\penalty0 (1-3):\penalty0 209--239, 2004.

\bibitem[Belkin et~al.(2006)Belkin, Niyogi, and Sindhwani]{belkin2006manifold}
Mikhail Belkin, Partha Niyogi, and Vikas Sindhwani.
\newblock Manifold regularization: A geometric framework for learning from
  labeled and unlabeled examples.
\newblock \emph{Journal of machine learning research}, 7\penalty0
  (Nov):\penalty0 2399--2434, 2006.

\bibitem[Bengio et~al.(2013)Bengio, Courville, and
  Vincent]{bengio2013representation}
Yoshua Bengio, Aaron Courville, and Pascal Vincent.
\newblock Representation learning: A review and new perspectives.
\newblock \emph{IEEE transactions on pattern analysis and machine
  intelligence}, 35\penalty0 (8):\penalty0 1798--1828, 2013.

\bibitem[Bengio et~al.(2018)Bengio, Lodi, and Prouvost]{bengio2018machine}
Yoshua Bengio, Andrea Lodi, and Antoine Prouvost.
\newblock Machine learning for combinatorial optimization: a methodological
  tour d'horizon.
\newblock \emph{arXiv preprint arXiv:1811.06128}, 2018.

\bibitem[Berg et~al.(2017)Berg, Kipf, and Welling]{berg2017graph}
Rianne van~den Berg, Thomas~N Kipf, and Max Welling.
\newblock Graph convolutional matrix completion.
\newblock \emph{arXiv preprint arXiv:1706.02263}, 2017.

\bibitem[Bianchi et~al.(2020)Bianchi, Grattarola, and
  Alippi]{bianchi2020spectral}
Filippo~Maria Bianchi, Daniele Grattarola, and Cesare Alippi.
\newblock Spectral clustering with graph neural networks for graph pooling.
\newblock In \emph{International Conference on Machine Learning}, pages
  874--883. PMLR, 2020.

\bibitem[Bonnabel(2013)]{bonnabel2013stochastic}
Silvere Bonnabel.
\newblock Stochastic gradient descent on riemannian manifolds.
\newblock \emph{IEEE Transactions on Automatic Control}, 58\penalty0
  (9):\penalty0 2217--2229, 2013.

\bibitem[Boscaini et~al.(2016)Boscaini, Masci, Rodol{\`a}, and
  Bronstein]{boscaini2016learning}
Davide Boscaini, Jonathan Masci, Emanuele Rodol{\`a}, and Michael Bronstein.
\newblock Learning shape correspondence with anisotropic convolutional neural
  networks.
\newblock In \emph{Advances in Neural Information Processing Systems}, pages
  3189--3197, 2016.

\bibitem[Bose and Hamilton(2019)]{bose2019compositional}
Avishek~Joey Bose and William Hamilton.
\newblock Compositional fairness constraints for graph embeddings.
\newblock \emph{arXiv preprint arXiv:1905.10674}, 2019.

\bibitem[Bronstein et~al.(2017)Bronstein, Bruna, LeCun, Szlam, and
  Vandergheynst]{bronstein2017geometric}
Michael~M Bronstein, Joan Bruna, Yann LeCun, Arthur Szlam, and Pierre
  Vandergheynst.
\newblock Geometric deep learning: going beyond euclidean data.
\newblock \emph{IEEE Signal Processing Magazine}, 34\penalty0 (4):\penalty0
  18--42, 2017.

\bibitem[Bruna et~al.(2014)Bruna, Zaremba, Szlam, and Lecun]{bruna2013spectral}
Joan Bruna, Wojciech Zaremba, Arthur Szlam, and Yann Lecun.
\newblock Spectral networks and locally connected networks on graphs
  international conference on learning representations (iclr2014).
\newblock \emph{CBLS, April}, 2014.

\bibitem[Bui et~al.(2018)Bui, Ravi, and Ramavajjala]{bui2017neural}
Thang~D Bui, Sujith Ravi, and Vivek Ramavajjala.
\newblock Neural graph learning: Training neural networks using graphs.
\newblock In \emph{Proceedings of the Eleventh ACM International Conference on
  Web Search and Data Mining}, pages 64--71, 2018.

\bibitem[Cai et~al.(2018)Cai, Zheng, and Chang]{cai2018comprehensive}
Hongyun Cai, Vincent~W Zheng, and Kevin Chang.
\newblock A comprehensive survey of graph embedding: problems, techniques and
  applications.
\newblock \emph{IEEE Transactions on Knowledge and Data Engineering}, 2018.

\bibitem[Cangea et~al.(2018)Cangea, Velickovic, Jovanovic, Kipf, , and
  Lio]{cangea2018towards}
Catalina Cangea, Petar Velickovic, Nikola Jovanovic, Thomas Kipf, , and Pietro
  Lio.
\newblock Towards sparse hierarchical graph classifiers.
\newblock In \emph{arXiv:1811.01287}, 2018.

\bibitem[Cao et~al.(2015)Cao, Lu, and Xu]{cao2015grarep}
Shaosheng Cao, Wei Lu, and Qiongkai Xu.
\newblock Grarep: Learning graph representations with global structural
  information.
\newblock In \emph{Proceedings of the 24th ACM International on Conference on
  Information and Knowledge Management}, pages 891--900. ACM, 2015.

\bibitem[Cao et~al.(2016)Cao, Lu, and Xu]{cao2016deep}
Shaosheng Cao, Wei Lu, and Qiongkai Xu.
\newblock Deep neural networks for learning graph representations.
\newblock In \emph{AAAI}, pages 1145--1152, 2016.

\bibitem[Chamberlain et~al.(2017)Chamberlain, Clough, and
  Deisenroth]{chamberlain2017neural}
Benjamin~Paul Chamberlain, James Clough, and Marc~Peter Deisenroth.
\newblock Neural embeddings of graphs in hyperbolic space.
\newblock \emph{arXiv preprint arXiv:1705.10359}, 2017.

\bibitem[Chami et~al.(2019)Chami, Ying, R{\'e}, and
  Leskovec]{chami2019hyperbolic}
Ines Chami, Zhitao Ying, Christopher R{\'e}, and Jure Leskovec.
\newblock Hyperbolic graph convolutional neural networks.
\newblock In \emph{Advances in Neural Information Processing Systems}, pages
  4869--4880, 2019.

\bibitem[Chami et~al.(2020)Chami, Wolf, Juan, Sala, Ravi, and
  R{\'e}]{chami2020low}
Ines Chami, Adva Wolf, Da-Cheng Juan, Frederic Sala, Sujith Ravi, and
  Christopher R{\'e}.
\newblock Low-dimensional hyperbolic knowledge graph embeddings.
\newblock In \emph{Proceedings of the 58th Annual Meeting of the Association
  for Computational Linguistics}, 2020.

\bibitem[Chapelle et~al.(2009)Chapelle, Scholkopf, and Zien]{chapellesemi}
Olivier Chapelle, Bernhard Scholkopf, and Alexander Zien.
\newblock Semi-supervised learning (chapelle, o. et al., eds.; 2006)[book
  reviews].
\newblock \emph{IEEE Transactions on Neural Networks}, 20\penalty0
  (3):\penalty0 542--542, 2009.

\bibitem[Chen et~al.(2020)Chen, Jacob, and Mairal]{chen2020cgkn}
Dexiong Chen, Laurent Jacob, and Julien Mairal.
\newblock Convolutional kernel networks for graph-structured data.
\newblock In \emph{International Conference on Machine Learning}, 2020.

\bibitem[Chen et~al.(2018{\natexlab{a}})Chen, Perozzi, Al-Rfou, and
  Skiena]{chen2018tutorial}
Haochen Chen, Bryan Perozzi, Rami Al-Rfou, and Steven Skiena.
\newblock A tutorial on network embeddings.
\newblock \emph{arXiv preprint arXiv:1808.02590}, 2018{\natexlab{a}}.

\bibitem[Chen et~al.(2018{\natexlab{b}})Chen, Perozzi, Hu, and
  Skiena]{chen2017harp}
Haochen Chen, Bryan Perozzi, Yifan Hu, and Steven Skiena.
\newblock Harp: Hierarchical representation learning for networks.
\newblock In \emph{Thirty-Second AAAI Conference on Artificial Intelligence},
  2018{\natexlab{b}}.

\bibitem[Chen et~al.(2018{\natexlab{c}})Chen, Sun, Tian, Perozzi, Chen, and
  Skiena]{chen2018edgelabels}
Haochen Chen, Xiaofei Sun, Yingtao Tian, Bryan Perozzi, Muhao Chen, and Steven
  Skiena.
\newblock Enhanced network embeddings via exploiting edge labels.
\newblock In \emph{Proceedings of the 27th ACM International Conference on
  Information and Knowledge Management}, CIKM ’18, page 1579–1582,
  2018{\natexlab{c}}.

\bibitem[Chen et~al.(2013)Chen, Fang, Hu, and Mahoney]{chen2013hyperbolicity}
Wei Chen, Wenjie Fang, Guangda Hu, and Michael~W Mahoney.
\newblock On the hyperbolicity of small-world and treelike random graphs.
\newblock \emph{Internet Mathematics}, 9\penalty0 (4):\penalty0 434--491, 2013.

\bibitem[Chen et~al.(2019{\natexlab{a}})Chen, Estrach, and
  Li]{chen2019supervised}
Zhengdao Chen, Joan~Bruna Estrach, and Lisha Li.
\newblock Supervised community detection with line graph neural networks.
\newblock In \emph{7th International Conference on Learning Representations,
  ICLR 2019}, 2019{\natexlab{a}}.

\bibitem[Chen et~al.(2019{\natexlab{b}})Chen, Villar, Chen, and
  Bruna]{chen2019equivalence}
Zhengdao Chen, Soledad Villar, Lei Chen, and Joan Bruna.
\newblock On the equivalence between graph isomorphism testing and function
  approximation with gnns.
\newblock In \emph{Advances in Neural Information Processing Systems}, pages
  15894--15902, 2019{\natexlab{b}}.

\bibitem[Chiang et~al.(2019)Chiang, Liu, Si, Li, Bengio, and
  Hsieh]{chiang2019clustergcn}
Wei-Lin Chiang, Xuanqing Liu, Si~Si, Yang Li, Samy Bengio, and Cho-Jui Hsieh.
\newblock Cluster-gcn: An efficient algorithm for training deep and large graph
  convolutional networks.
\newblock In \emph{ACM SIGKDD Conference on Knowledge Discovery and Data Mining
  (KDD)}, 2019.
\newblock URL \url{https://arxiv.org/pdf/1905.07953.pdf}.

\bibitem[Cho et~al.(2014)Cho, Van~Merri{\"e}nboer, Bahdanau, and
  Bengio]{cho2014properties}
Kyunghyun Cho, Bart Van~Merri{\"e}nboer, Dzmitry Bahdanau, and Yoshua Bengio.
\newblock On the properties of neural machine translation: Encoder-decoder
  approaches.
\newblock \emph{arXiv preprint arXiv:1409.1259}, 2014.

\bibitem[Cox and Cox(2008)]{cox2008multidimensional}
Michael~AA Cox and Trevor~F Cox.
\newblock Multidimensional scaling.
\newblock In \emph{Handbook of data visualization}, pages 315--347. Springer,
  2008.

\bibitem[Cruz et~al.(2019)Cruz, Kipf, and Welling]{cruz2019modular}
Daniel Fernando~Daza Cruz, Thomas Kipf, and Max Welling.
\newblock A modular framework for unsupervised graph representation learning.
\newblock 2019.

\bibitem[De~Cao and Kipf(2018)]{de2018molgan}
Nicola De~Cao and Thomas Kipf.
\newblock Molgan: An implicit generative model for small molecular graphs.
\newblock \emph{arXiv preprint arXiv:1805.11973}, 2018.

\bibitem[Dean and Ghemawat(2008)]{dean08mapreduce}
Jeffrey Dean and Sanjay Ghemawat.
\newblock Mapreduce: Simplified data processing on large clusters.
\newblock \emph{Commun. ACM}, page 107–113, 2008.
\newblock \doi{10.1145/1327452.1327492}.
\newblock URL \url{https://doi.org/10.1145/1327452.1327492}.

\bibitem[Debnath et~al.(1991)Debnath, de~Compadre, Debnath, Shusterman, , and
  Hansch]{debnath1991structure}
Asim~Kumar Debnath, Rosa L.~Lopez de~Compadre, Gargi Debnath, Alan~J.
  Shusterman, , and Corwin Hansch.
\newblock Structure-activity relationship of mutagenic aromatic and
  heteroaromatic nitro compounds. correlation with molecular orbital energies
  and hydrophobicity.
\newblock In \emph{J. Med. Chem.}, pages 786--797, 1991.

\bibitem[Defferrard et~al.(2016)Defferrard, Bresson, and
  Vandergheynst]{defferrard2016convolutional}
Micha{\"e}l Defferrard, Xavier Bresson, and Pierre Vandergheynst.
\newblock Convolutional neural networks on graphs with fast localized spectral
  filtering.
\newblock In \emph{Advances in Neural Information Processing Systems}, pages
  3844--3852, 2016.

\bibitem[Deng et~al.(2009)Deng, Dong, Socher, Li, Li, and
  Fei-Fei]{deng2009imagenet}
J.~Deng, W.~Dong, R.~Socher, L.-J. Li, K.~Li, and L.~Fei-Fei.
\newblock {ImageNet: A Large-Scale Hierarchical Image Database}.
\newblock In \emph{CVPR09}, 2009.

\bibitem[Du et~al.(2019)Du, Hou, Salakhutdinov, Poczos, Wang, and
  Xu]{du2019gntk}
Simon~S Du, Kangcheng Hou, Russ~R Salakhutdinov, Barnabas Poczos, Ruosong Wang,
  and Keyulu Xu.
\newblock Graph neural tangent kernel: Fusing graph neural networks with graph
  kernels.
\newblock In \emph{Advances in Neural Information Processing Systems}, 2019.

\bibitem[Duvenaud et~al.(2015)Duvenaud, Maclaurin, Iparraguirre, Bombarell,
  Hirzel, Aspuru-Guzik, and Adams]{duvenaud2015convolutional}
David~K Duvenaud, Dougal Maclaurin, Jorge Iparraguirre, Rafael Bombarell,
  Timothy Hirzel, Al{\'a}n Aspuru-Guzik, and Ryan~P Adams.
\newblock Convolutional networks on graphs for learning molecular fingerprints.
\newblock In \emph{Advances in neural information processing systems}, pages
  2224--2232, 2015.

\bibitem[Dwivedi et~al.(2020)Dwivedi, Joshi, Laurent, Bengio, and
  Bresson]{dwivedi2020benchmarking}
Vijay~Prakash Dwivedi, Chaitanya~K Joshi, Thomas Laurent, Yoshua Bengio, and
  Xavier Bresson.
\newblock Benchmarking graph neural networks.
\newblock \emph{arXiv preprint arXiv:2003.00982}, 2020.

\bibitem[Elton et~al.(2019)Elton, Boukouvalas, Fuge, and
  Chung]{elton2019molecular}
Daniel~C Elton, Zois Boukouvalas, Mark~D Fuge, and Peter~W Chung.
\newblock Deep learning for molecular design—a review of the state of the
  art.
\newblock \emph{Molecular Systems Design \& Engineering}, 4\penalty0
  (4):\penalty0 828--849, 2019.

\bibitem[Epasto and Perozzi(2019)]{epasto2019splitter}
Alessandro Epasto and Bryan Perozzi.
\newblock Is a single embedding enough? learning node representations that
  capture multiple social contexts.
\newblock In \emph{The World Wide Web Conference}, WWW ’19, page 394–404,
  New York, NY, USA, 2019. Association for Computing Machinery.
\newblock ISBN 9781450366748.
\newblock \doi{10.1145/3308558.3313660}.
\newblock URL \url{https://doi.org/10.1145/3308558.3313660}.

\bibitem[Epasto et~al.(2017)Epasto, Lattanzi, and Paes~Leme]{epasto2017persona}
Alessandro Epasto, Silvio Lattanzi, and Renato Paes~Leme.
\newblock Ego-splitting framework: From non-overlapping to overlapping
  clusters.
\newblock In \emph{Proceedings of the 23rd ACM SIGKDD International Conference
  on Knowledge Discovery and Data Mining}, KDD ’17, page 145–154, New York,
  NY, USA, 2017. Association for Computing Machinery.
\newblock ISBN 9781450348874.
\newblock \doi{10.1145/3097983.3098054}.
\newblock URL \url{https://doi.org/10.1145/3097983.3098054}.

\bibitem[Feng et~al.(2018)Feng, Dueva, Cherkasov, and Ester]{feng2018padme}
Qingyuan Feng, Evgenia Dueva, Artem Cherkasov, and Martin Ester.
\newblock Padme: A deep learning-based framework for drug-target interaction
  prediction.
\newblock \emph{arXiv preprint arXiv:1807.09741}, 2018.

\bibitem[Fey and Lenssen(2019)]{fey2019fast}
Matthias Fey and Jan~Eric Lenssen.
\newblock Fast graph representation learning with pytorch geometric.
\newblock \emph{arXiv preprint arXiv:1903.02428}, 2019.

\bibitem[Gao and Ji(2019)]{goa2019graphunets}
Hongyang Gao and Shuiwang Ji.
\newblock Graph u-nets.
\newblock In \emph{Proceedings of the 36th International Conference on Machine
  Learning}, 2019.

\bibitem[Garcia and Bruna(2018)]{bruna18fewshot}
Victor Garcia and Joan Bruna.
\newblock Few-shot learning with graph neural networks.
\newblock In \emph{International Conference on Learning Representations
  (ICLR)}, 2018.

\bibitem[Garg et~al.(2020)Garg, Jegelka, and Jaakkola]{garg2020generalization}
Vikas~K Garg, Stefanie Jegelka, and Tommi Jaakkola.
\newblock Generalization and representational limits of graph neural networks.
\newblock \emph{arXiv preprint arXiv:2002.06157}, 2020.

\bibitem[Gilmer et~al.(2017)Gilmer, Schoenholz, Riley, Vinyals, and
  Dahl]{gilmer2017neural}
Justin Gilmer, Samuel~S Schoenholz, Patrick~F Riley, Oriol Vinyals, and
  George~E Dahl.
\newblock Neural message passing for quantum chemistry.
\newblock In \emph{Proceedings of the 34th International Conference on Machine
  Learning-Volume 70}, pages 1263--1272. JMLR. org, 2017.

\bibitem[Godec(2018)]{godec_2018}
Primož Godec.
\newblock
  \url{https://towardsdatascience.com/graph-embeddings-the-summary-cc6075aba007},
  2018.

\bibitem[Gonen and Goldberg(2019)]{gonen2019lipstick}
Hila Gonen and Yoav Goldberg.
\newblock Lipstick on a pig: Debiasing methods cover up systematic gender
  biases in word embeddings but do not remove them.
\newblock \emph{arXiv preprint arXiv:1903.03862}, 2019.

\bibitem[Gori et~al.(2005)Gori, Monfardini, and Scarselli]{gori2005new}
Marco Gori, Gabriele Monfardini, and Franco Scarselli.
\newblock A new model for learning in graph domains.
\newblock In \emph{Proceedings. 2005 IEEE International Joint Conference on
  Neural Networks, 2005.}, volume~2, pages 729--734. IEEE, 2005.

\bibitem[Goyal and Ferrara(2018{\natexlab{a}})]{goyal2018gem}
Palash Goyal and Emilio Ferrara.
\newblock Gem: a python package for graph embedding methods.
\newblock \emph{Journal of Open Source Software}, 3\penalty0 (29):\penalty0
  876, 2018{\natexlab{a}}.

\bibitem[Goyal and Ferrara(2018{\natexlab{b}})]{goyal2018graph}
Palash Goyal and Emilio Ferrara.
\newblock Graph embedding techniques, applications, and performance: A survey.
\newblock \emph{Knowledge-Based Systems}, 151:\penalty0 78--94,
  2018{\natexlab{b}}.

\bibitem[Grover and Leskovec(2016)]{grover2016node2vec}
Aditya Grover and Jure Leskovec.
\newblock node2vec: Scalable feature learning for networks.
\newblock In \emph{Proceedings of the 22nd ACM SIGKDD international conference
  on Knowledge discovery and data mining}, pages 855--864. ACM, 2016.

\bibitem[Grover et~al.(2019)Grover, Zweig, and Ermon]{grover2018graphite}
Aditya Grover, Aaron Zweig, and Stefano Ermon.
\newblock Graphite: Iterative generative modeling of graphs.
\newblock In \emph{International Conference on Machine Learning}, pages
  2434--2444, 2019.

\bibitem[Gu et~al.(2018)Gu, Sala, Gunel, and R{\'e}]{gu2018learning}
Albert Gu, Frederic Sala, Beliz Gunel, and Christopher R{\'e}.
\newblock Learning mixed-curvature representations in product spaces.
\newblock \emph{International Conference on Learning Representations}, 2018.

\bibitem[Halcrow et~al.(2020)Halcrow, Mosoi, Ruth, and Perozzi]{halcrow-grale}
Jonathan Halcrow, Alexandru Mosoi, Sam Ruth, and Bryan Perozzi.
\newblock Grale: Designing networks for graph learning.
\newblock In \emph{Proceedings of the 26th ACM SIGKDD International Conference
  on Knowledge Discovery \& Data Mining}, KDD '20, page 2523–2532, New York,
  NY, USA, 2020. Association for Computing Machinery.
\newblock ISBN 9781450379984.
\newblock \doi{10.1145/3394486.3403302}.
\newblock URL \url{https://doi.org/10.1145/3394486.3403302}.

\bibitem[Hamilton et~al.(2017{\natexlab{a}})Hamilton, Ying, and
  Leskovec]{hamilton2017inductive}
Will Hamilton, Zhitao Ying, and Jure Leskovec.
\newblock Inductive representation learning on large graphs.
\newblock In \emph{Advances in Neural Information Processing Systems}, pages
  1024--1034, 2017{\natexlab{a}}.

\bibitem[Hamilton et~al.(2017{\natexlab{b}})Hamilton, Ying, and
  Leskovec]{hamilton2017representation}
William~L Hamilton, Rex Ying, and Jure Leskovec.
\newblock Representation learning on graphs: Methods and applications.
\newblock \emph{arXiv preprint arXiv:1709.05584}, 2017{\natexlab{b}}.

\bibitem[Hammond et~al.(2011)Hammond, Vandergheynst, and
  Gribonval]{hammond2011wavelets}
David~K Hammond, Pierre Vandergheynst, and R{\'e}mi Gribonval.
\newblock Wavelets on graphs via spectral graph theory.
\newblock \emph{Applied and Computational Harmonic Analysis}, 30\penalty0
  (2):\penalty0 129--150, 2011.

\bibitem[Henaff et~al.(2015)Henaff, Bruna, and LeCun]{henaff2015deep}
Mikael Henaff, Joan Bruna, and Yann LeCun.
\newblock Deep convolutional networks on graph-structured data.
\newblock \emph{arXiv preprint arXiv:1506.05163}, 2015.

\bibitem[Hochreiter and Schmidhuber(1997)]{hochreiter1997long}
Sepp Hochreiter and J{\"u}rgen Schmidhuber.
\newblock Long short-term memory.
\newblock \emph{Neural computation}, 9\penalty0 (8):\penalty0 1735--1780, 1997.

\bibitem[Hu et~al.(2020)Hu, Fey, Zitnik, Dong, Ren, Liu, Catasta, and
  Leskovec]{ogb_2020}
Weihua Hu, Matthias Fey, Marinka Zitnik, Yuxiao Dong, Hongyu Ren, Bowen Liu,
  Michele Catasta, and Jure Leskovec.
\newblock Open graph benchmark: Datasets for machine learning on graphs.
\newblock \emph{arXiv preprint arXiv:2005.00687}, 2020.

\bibitem[Huang et~al.(2020)Huang, He, Huang, Sun, Abu-El-Haija, Perozzi,
  Lerman, Morstatter, and Galstyan]{huang2020personalized}
Di~Huang, Zihao He, Yuzhong Huang, Kexuan Sun, Sami Abu-El-Haija, Bryan
  Perozzi, Kristina Lerman, Fred Morstatter, and Aram Galstyan.
\newblock Graph embedding with personalized context distribution.
\newblock In \emph{Companion Proceedings of the Web Conference 2020}, WWW
  ’20, page 655–661, 2020.

\bibitem[Jin et~al.(2018)Jin, Barzilay, and Jaakkola]{jin2018junction}
Wengong Jin, Regina Barzilay, and Tommi Jaakkola.
\newblock Junction tree variational autoencoder for molecular graph generation.
\newblock In \emph{International Conference on Machine Learning}, 2018.

\bibitem[Jolliffe(2011)]{jolliffe2011principal}
Ian Jolliffe.
\newblock Principal component analysis.
\newblock In \emph{International encyclopedia of statistical science}, pages
  1094--1096. Springer, 2011.

\bibitem[Jonckheere et~al.(2008)Jonckheere, Lohsoonthorn, and
  Bonahon]{jonckheere2008scaled}
Edmond Jonckheere, Poonsuk Lohsoonthorn, and Francis Bonahon.
\newblock Scaled gromov hyperbolic graphs.
\newblock \emph{Journal of Graph Theory}, 2008.

\bibitem[Khalil et~al.(2017)Khalil, Dai, Zhang, Dilkina, and
  Song]{khalil2017learning}
Elias Khalil, Hanjun Dai, Yuyu Zhang, Bistra Dilkina, and Le~Song.
\newblock Learning combinatorial optimization algorithms over graphs.
\newblock In \emph{Advances in Neural Information Processing Systems}, pages
  6348--6358, 2017.

\bibitem[Kipf and Welling(2016{\natexlab{a}})]{kipf2016semi}
Thomas~N Kipf and Max Welling.
\newblock Semi-supervised classification with graph convolutional networks.
\newblock \emph{arXiv preprint arXiv:1609.02907}, 2016{\natexlab{a}}.

\bibitem[Kipf and Welling(2016{\natexlab{b}})]{kipf2016variational}
Thomas~N Kipf and Max Welling.
\newblock Variational graph auto-encoders.
\newblock \emph{arXiv preprint arXiv:1611.07308}, 2016{\natexlab{b}}.

\bibitem[Kleinberg(2007)]{kleinberg2007geographic}
Robert Kleinberg.
\newblock Geographic routing using hyperbolic space.
\newblock In \emph{IEEE INFOCOM 2007-26th IEEE International Conference on
  Computer Communications}, pages 1902--1909. IEEE, 2007.

\bibitem[Konstas et~al.(2009)Konstas, Stathopoulos, and Jose]{socrecgraph}
Ioannis Konstas, Vassilios Stathopoulos, and Joemon~M. Jose.
\newblock On social networks and collaborative recommendation.
\newblock In \emph{Proceedings of the 32nd international ACM SIGIR conference
  on Research and development in information retrieval}, pages 195--202, 2009.

\bibitem[Kriege et~al.(2020)Kriege, Johansson, and Morris]{krieg2020survey}
N.~M. Kriege, F.~D. Johansson, and C.~Morris.
\newblock A survey on graph kernels.
\newblock In \emph{Applied Network Science}, pages 1--42, 2020.

\bibitem[Krioukov et~al.(2010)Krioukov, Papadopoulos, Kitsak, Vahdat, and
  Bogun{\'a}]{krioukov2010hyperbolic}
Dmitri Krioukov, Fragkiskos Papadopoulos, Maksim Kitsak, Amin Vahdat, and
  Mari{\'a}n Bogun{\'a}.
\newblock Hyperbolic geometry of complex networks.
\newblock \emph{Physical Review E}, 82\penalty0 (3):\penalty0 036106, 2010.

\bibitem[Kruskal(1964)]{kruskal1964multidimensional}
Joseph~B Kruskal.
\newblock Multidimensional scaling by optimizing goodness of fit to a nonmetric
  hypothesis.
\newblock \emph{Psychometrika}, 29\penalty0 (1):\penalty0 1--27, 1964.

\bibitem[Kuznetsova et~al.(2020)Kuznetsova, Rom, Alldrin, Uijlings, Krasin,
  Pont-Tuset, Kamali, Popov, Malloci, Kolesnikov, Duerig, and
  Ferrari]{kuznetsova2020openimages}
Alina Kuznetsova, Hassan Rom, Neil Alldrin, Jasper Uijlings, Ivan Krasin, Jordi
  Pont-Tuset, Shahab Kamali, Stefan Popov, Matteo Malloci, Alexander
  Kolesnikov, Tom Duerig, and Vittorio Ferrari.
\newblock The open images dataset v4: Unified image classification, object
  detection, and visual relationship detection at scale.
\newblock \emph{IJCV}, 2020.

\bibitem[Lamb et~al.(2020)Lamb, Garcez, Gori, Prates, Avelar, and
  Vardi]{lamb2020graph}
Luis Lamb, Artur Garcez, Marco Gori, Marcelo Prates, Pedro Avelar, and Moshe
  Vardi.
\newblock Graph neural networks meet neural-symbolic computing: A survey and
  perspective.
\newblock \emph{arXiv preprint arXiv:2003.00330}, 2020.

\bibitem[Le et~al.(2019)Le, Roller, Papaxanthos, Kiela, and
  Nickel]{le2019inferring}
Matthew Le, Stephen Roller, Laetitia Papaxanthos, Douwe Kiela, and Maximilian
  Nickel.
\newblock Inferring concept hierarchies from text corpora via hyperbolic
  embeddings.
\newblock In \emph{Proceedings of the 57th Annual Meeting of the Association
  for Computational Linguistics}, pages 3231--3241, 2019.

\bibitem[LeCun et~al.(1989)LeCun, Boser, Denker, Henderson, Howard, Hubbard,
  and Jackel]{lecun1989backpropagation}
Yann LeCun, Bernhard Boser, John~S Denker, Donnie Henderson, Richard~E Howard,
  Wayne Hubbard, and Lawrence~D Jackel.
\newblock Backpropagation applied to handwritten zip code recognition.
\newblock \emph{Neural computation}, 1\penalty0 (4):\penalty0 541--551, 1989.

\bibitem[Lee et~al.(2019)Lee, Lee, , and Kang]{lee2019selfattention}
Junhyun Lee, Inyeop Lee, , and Jaewoo Kang.
\newblock Self-attention graph pooling.
\newblock In \emph{In International Conference on Machine Learning}, 2019.

\bibitem[Lerer et~al.(2019)Lerer, Wu, Shen, Lacroix, Wehrstedt, Bose, and
  Peysakhovich]{lerer2019biggraph}
Adam Lerer, Ledell Wu, Jiajun Shen, Timothee Lacroix, Luca Wehrstedt, Abhijit
  Bose, and Alex Peysakhovich.
\newblock {PyTorch-BigGraph: A Large-scale Graph Embedding System}.
\newblock In \emph{Proceedings of the 2nd SysML Conference}, Palo Alto, CA,
  USA, 2019.

\bibitem[Levy and Goldberg(2014)]{levy2014neural}
Omer Levy and Yoav Goldberg.
\newblock Neural word embedding as implicit matrix factorization.
\newblock In \emph{Advances in neural information processing systems}, pages
  2177--2185, 2014.

\bibitem[Li et~al.(2015)Li, Tarlow, Brockschmidt, and Zemel]{li2015gated}
Yujia Li, Daniel Tarlow, Marc Brockschmidt, and Richard Zemel.
\newblock Gated graph sequence neural networks.
\newblock \emph{arXiv preprint arXiv:1511.05493}, 2015.

\bibitem[Li et~al.(2018)Li, Vinyals, Dyer, Pascanu, and
  Battaglia]{li2018learning}
Yujia Li, Oriol Vinyals, Chris Dyer, Razvan Pascanu, and Peter Battaglia.
\newblock Learning deep generative models of graphs.
\newblock \emph{arXiv preprint arXiv:1803.03324}, 2018.

\bibitem[Liben-Nowell and Kleinberg(2007)]{liben2007link}
David Liben-Nowell and Jon Kleinberg.
\newblock The link-prediction problem for social networks.
\newblock \emph{Journal of the American society for information science and
  technology}, 58\penalty0 (7):\penalty0 1019--1031, 2007.

\bibitem[Liu et~al.(2018)Liu, Allamanis, Brockschmidt, and
  Gaunt]{liu2018constrained}
Qi~Liu, Miltiadis Allamanis, Marc Brockschmidt, and Alexander Gaunt.
\newblock Constrained graph variational autoencoders for molecule design.
\newblock In \emph{Advances in Neural Information Processing Systems}, pages
  7795--7804, 2018.

\bibitem[Liu et~al.(2019)Liu, Nickel, and Kiela]{liu2019hyperbolic}
Qi~Liu, Maximilian Nickel, and Douwe Kiela.
\newblock Hyperbolic graph neural networks.
\newblock In \emph{Advances in Neural Information Processing Systems}, pages
  8228--8239, 2019.

\bibitem[Loukas(2019)]{loukas2019graph}
Andreas Loukas.
\newblock What graph neural networks cannot learn: depth vs width.
\newblock \emph{arXiv preprint arXiv:1907.03199}, 2019.

\bibitem[Maaten and Hinton(2008)]{maaten2008visualizing}
Laurens van~der Maaten and Geoffrey Hinton.
\newblock Visualizing data using t-sne.
\newblock \emph{Journal of machine learning research}, 9\penalty0
  (Nov):\penalty0 2579--2605, 2008.

\bibitem[Markowitz et~al.(2021)Markowitz, Balasubramanian, Mirtaheri,
  Abu-El-Haija, Perozzi, Steeg, and Galstyan]{markowitz2021graph}
Elan~Sopher Markowitz, Keshav Balasubramanian, Mehrnoosh Mirtaheri, Sami
  Abu-El-Haija, Bryan Perozzi, Greg~Ver Steeg, and Aram Galstyan.
\newblock Graph traversal with tensor functionals: A meta-algorithm for
  scalable learning.
\newblock In \emph{International Conference on Learning Representations}, 2021.

\bibitem[Maron et~al.(2018)Maron, Ben-Hamu, Shamir, and
  Lipman]{maron2018invariant}
Haggai Maron, Heli Ben-Hamu, Nadav Shamir, and Yaron Lipman.
\newblock Invariant and equivariant graph networks.
\newblock In \emph{International Conference on Learning Representations}, 2018.

\bibitem[Masci et~al.(2015)Masci, Boscaini, Bronstein, and
  Vandergheynst]{masci2015geodesic}
Jonathan Masci, Davide Boscaini, Michael Bronstein, and Pierre Vandergheynst.
\newblock Geodesic convolutional neural networks on riemannian manifolds.
\newblock In \emph{Proceedings of the IEEE international conference on computer
  vision workshops}, pages 37--45, 2015.

\bibitem[Mehrabi et~al.(2019)Mehrabi, Morstatter, Saxena, Lerman, and
  Galstyan]{mehrabi2019survey}
Ninareh Mehrabi, Fred Morstatter, Nripsuta Saxena, Kristina Lerman, and Aram
  Galstyan.
\newblock A survey on bias and fairness in machine learning.
\newblock \emph{arXiv preprint arXiv:1908.09635}, 2019.

\bibitem[Mikolov et~al.(2013)Mikolov, Sutskever, Chen, Corrado, and
  Dean]{mikolov2013distributed}
Tomas Mikolov, Ilya Sutskever, Kai Chen, Greg~S Corrado, and Jeff Dean.
\newblock Distributed representations of words and phrases and their
  compositionality.
\newblock In \emph{Advances in neural information processing systems}, pages
  3111--3119, 2013.

\bibitem[Monti et~al.(2017)Monti, Boscaini, Masci, Rodola, Svoboda, and
  Bronstein]{monti2017geometric}
Federico Monti, Davide Boscaini, Jonathan Masci, Emanuele Rodola, Jan Svoboda,
  and Michael~M Bronstein.
\newblock Geometric deep learning on graphs and manifolds using mixture model
  cnns.
\newblock In \emph{Proceedings of the IEEE Conference on Computer Vision and
  Pattern Recognition}, pages 5115--5124, 2017.

\bibitem[Morris et~al.(2019)Morris, Ritzert, Fey, Hamilton, Lenssen, Rattan,
  and Grohe]{morris2019weisfeiler}
Christopher Morris, Martin Ritzert, Matthias Fey, William~L Hamilton, Jan~Eric
  Lenssen, Gaurav Rattan, and Martin Grohe.
\newblock Weisfeiler and leman go neural: Higher-order graph neural networks.
\newblock In \emph{Proceedings of the AAAI Conference on Artificial
  Intelligence}, volume~33, pages 4602--4609, 2019.

\bibitem[Muscoloni et~al.(2017)Muscoloni, Thomas, Ciucci, Bianconi, and
  Cannistraci]{muscoloni2017machine}
Alessandro Muscoloni, Josephine~Maria Thomas, Sara Ciucci, Ginestra Bianconi,
  and Carlo~Vittorio Cannistraci.
\newblock Machine learning meets complex networks via coalescent embedding in
  the hyperbolic space.
\newblock \emph{Nature communications}, 8\penalty0 (1):\penalty0 1--19, 2017.

\bibitem[Nickel and Kiela(2017)]{nickel2017poincare}
Maximillian Nickel and Douwe Kiela.
\newblock Poincar{\'e} embeddings for learning hierarchical representations.
\newblock In \emph{Advances in neural information processing systems}, pages
  6338--6347, 2017.

\bibitem[Nickel and Kiela(2018)]{nickel2018learning}
Maximillian Nickel and Douwe Kiela.
\newblock Learning continuous hierarchies in the lorentz model of hyperbolic
  geometry.
\newblock In \emph{International Conference on Machine Learning}, pages
  3779--3788, 2018.

\bibitem[Nowak et~al.(2017)Nowak, Villar, Bandeira, and
  Bruna]{nowak2017revised}
Alex Nowak, Soledad Villar, Afonso~S Bandeira, and Joan Bruna.
\newblock Revised note on learning algorithms for quadratic assignment with
  graph neural networks.
\newblock \emph{arXiv preprint arXiv:1706.07450}, 2017.

\bibitem[Ou et~al.(2016)Ou, Cui, Pei, Zhang, and Zhu]{ou2016asymmetric}
Mingdong Ou, Peng Cui, Jian Pei, Ziwei Zhang, and Wenwu Zhu.
\newblock Asymmetric transitivity preserving graph embedding.
\newblock In \emph{Proceedings of the 22nd ACM SIGKDD international conference
  on Knowledge discovery and data mining}, pages 1105--1114. ACM, 2016.

\bibitem[Palowitch and Perozzi(2019)]{palowitch2019monet}
John Palowitch and Bryan Perozzi.
\newblock Monet: Debiasing graph embeddings via the metadata-orthogonal
  training unit.
\newblock \emph{arXiv preprint arXiv:1909.11793}, 2019.

\bibitem[Papadopoulos et~al.(2012)Papadopoulos, Kitsak, Serrano, Bogun{\'a},
  and Krioukov]{papadopoulos2012popularity}
Fragkiskos Papadopoulos, Maksim Kitsak, M~{\'A}ngeles Serrano, Mari{\'a}n
  Bogun{\'a}, and Dmitri Krioukov.
\newblock Popularity versus similarity in growing networks.
\newblock \emph{Nature}, 489\penalty0 (7417):\penalty0 537--540, 2012.

\bibitem[Papadopoulos et~al.(2014)Papadopoulos, Psomas, and
  Krioukov]{papadopoulos2014network}
Fragkiskos Papadopoulos, Constantinos Psomas, and Dmitri Krioukov.
\newblock Network mapping by replaying hyperbolic growth.
\newblock \emph{IEEE/ACM Transactions on Networking}, 23\penalty0 (1):\penalty0
  198--211, 2014.

\bibitem[Peng et~al.(2020)Peng, Huang, Luo, Zheng, Rong, Xu, and
  Huang]{peng2020graph}
Zhen Peng, Wenbing Huang, Minnan Luo, Qinghua Zheng, Yu~Rong, Tingyang Xu, and
  Junzhou Huang.
\newblock {Graph Representation Learning via Graphical Mutual Information
  Maximization}.
\newblock In \emph{Proceedings of The Web Conference}, 2020.
\newblock \doi{https://doi.org/10.1145/3366423.3380112}.

\bibitem[Pennington et~al.(2014)Pennington, Socher, and
  Manning]{pennington2014glove}
Jeffrey Pennington, Richard Socher, and Christopher Manning.
\newblock Glove: Global vectors for word representation.
\newblock In \emph{Proceedings of the 2014 conference on empirical methods in
  natural language processing (EMNLP)}, pages 1532--1543, 2014.

\bibitem[Perozzi et~al.(2014)Perozzi, Al-Rfou, and Skiena]{perozzi2014deepwalk}
Bryan Perozzi, Rami Al-Rfou, and Steven Skiena.
\newblock Deepwalk: Online learning of social representations.
\newblock In \emph{Proceedings of the 20th ACM SIGKDD international conference
  on Knowledge discovery and data mining}, pages 701--710. ACM, 2014.

\bibitem[Pineda(1988)]{pineda1988generalization}
Fernando~J Pineda.
\newblock Generalization of back propagation to recurrent and higher order
  neural networks.
\newblock In \emph{Neural information processing systems}, pages 602--611,
  1988.

\bibitem[Prates et~al.(2019)Prates, Avelar, Lemos, Lamb, and
  Vardi]{prates2019learning}
Marcelo Prates, Pedro~HC Avelar, Henrique Lemos, Luis~C Lamb, and Moshe~Y
  Vardi.
\newblock Learning to solve np-complete problems: A graph neural network for
  decision tsp.
\newblock In \emph{Proceedings of the AAAI Conference on Artificial
  Intelligence}, volume~33, pages 4731--4738, 2019.

\bibitem[Qiu et~al.(2018)Qiu, Dong, Ma, Li, Wang, and Tang]{qiu2018network}
Jiezhong Qiu, Yuxiao Dong, Hao Ma, Jian Li, Kuansan Wang, and Jie Tang.
\newblock Network embedding as matrix factorization: Unifying deepwalk, line,
  pte, and node2vec.
\newblock In \emph{Proceedings of the Eleventh ACM International Conference on
  Web Search and Data Mining}, pages 459--467, 2018.

\bibitem[Qiu et~al.(2019)Qiu, Dong, Ma, Li, Wang, Wang, and
  Tang]{qiu2019NetSMF}
Jiezhong Qiu, Yuxiao Dong, Hao Ma, Jian Li, Chi Wang, Kuansan Wang, and Jie
  Tang.
\newblock Netsmf: Large-scale network embedding as sparse matrix factorization.
\newblock In \emph{The World Wide Web Conference}, WWW ’19, page 1509–1520,
  New York, NY, USA, 2019. Association for Computing Machinery.
\newblock ISBN 9781450366748.
\newblock \doi{10.1145/3308558.3313446}.
\newblock URL \url{https://doi.org/10.1145/3308558.3313446}.

\bibitem[Ragoza et~al.(2017)Ragoza, Hochuli, Idrobo, Sunseri, and
  Koes]{Ragoza2017proteinligand}
Matthew Ragoza, Joshua Hochuli, Elisa Idrobo, Jocelyn Sunseri, and David~Ryan
  Koes.
\newblock Protein–ligand scoring with convolutional neural networks.
\newblock \emph{Journal of Chemical Information and Modeling}, 57\penalty0
  (4):\penalty0 942--957, 2017.
\newblock \doi{10.1021/acs.jcim.6b00740}.
\newblock URL \url{https://doi.org/10.1021/acs.jcim.6b00740}.
\newblock PMID: 28368587.

\bibitem[Roweis and Saul(2000)]{roweis2000nonlinear}
Sam~T Roweis and Lawrence~K Saul.
\newblock Nonlinear dimensionality reduction by locally linear embedding.
\newblock \emph{science}, 290\penalty0 (5500):\penalty0 2323--2326, 2000.

\bibitem[Rozemberczki et~al.(2019)Rozemberczki, Davies, Sarkar, and
  Sutton]{rozemberczki2019gemsec}
Benedek Rozemberczki, Ryan Davies, Rik Sarkar, and Charles Sutton.
\newblock Gemsec: Graph embedding with self clustering.
\newblock In \emph{Proceedings of the 2019 IEEE/ACM International Conference on
  Advances in Social Networks Analysis and Mining}, ASONAM ’19, page 65–72,
  New York, NY, USA, 2019. Association for Computing Machinery.
\newblock ISBN 9781450368681.
\newblock \doi{10.1145/3341161.3342890}.
\newblock URL \url{https://doi.org/10.1145/3341161.3342890}.

\bibitem[Rozemberczki et~al.(2021)Rozemberczki, Englert, Kapoor, Blais, and
  Perozzi]{rozemberczki-pathfinder}
Benedek Rozemberczki, Peter Englert, Amol Kapoor, Martin Blais, and Bryan
  Perozzi.
\newblock Pathfinder discovery networks for neural message passing.
\newblock In \emph{Proceedings of the Web Conference 2021}, WWW '21, page
  2547–2558, New York, NY, USA, 2021. Association for Computing Machinery.
\newblock ISBN 9781450383127.
\newblock \doi{10.1145/3442381.3449882}.
\newblock URL \url{https://doi.org/10.1145/3442381.3449882}.

\bibitem[Sala et~al.(2018)Sala, De~Sa, Gu, and Re]{de2018representation}
Frederic Sala, Chris De~Sa, Albert Gu, and Christopher Re.
\newblock Representation tradeoffs for hyperbolic embeddings.
\newblock In \emph{International Conference on Machine Learning}, pages
  4460--4469, 2018.

\bibitem[Sarkar(2011)]{sarkar2011low}
Rik Sarkar.
\newblock Low distortion delaunay embedding of trees in hyperbolic plane.
\newblock In \emph{International Symposium on Graph Drawing}, pages 355--366.
  Springer, 2011.

\bibitem[Scarselli et~al.(2009)Scarselli, Gori, Tsoi, Hagenbuchner, and
  Monfardini]{scarselli2009graph}
Franco Scarselli, Marco Gori, Ah~Chung Tsoi, Markus Hagenbuchner, and Gabriele
  Monfardini.
\newblock The graph neural network model.
\newblock \emph{IEEE Transactions on Neural Networks}, 20\penalty0
  (1):\penalty0 61--80, 2009.

\bibitem[Schlichtkrull et~al.(2018)Schlichtkrull, Kipf, Bloem, van~den Berg,
  Titov, and Welling]{schlichtkrull2018modeling}
Michael Schlichtkrull, Thomas~N Kipf, Peter Bloem, Rianne van~den Berg, Ivan
  Titov, and Max Welling.
\newblock Modeling relational data with graph convolutional networks.
\newblock In \emph{European Semantic Web Conference}, pages 593--607. Springer,
  2018.

\bibitem[Selsam et~al.(2018)Selsam, Lamm, B{\"u}nz, Liang, de~Moura, and
  Dill]{selsam2018learning}
Daniel Selsam, Matthew Lamm, Benedikt B{\"u}nz, Percy Liang, Leonardo de~Moura,
  and David~L Dill.
\newblock Learning a sat solver from single-bit supervision.
\newblock \emph{arXiv preprint arXiv:1802.03685}, 2018.

\bibitem[Shchur et~al.(2018)Shchur, Mumme, Bojchevski, and
  G{\"u}nnemann]{shchur2018pitfalls}
Oleksandr Shchur, Maximilian Mumme, Aleksandar Bojchevski, and Stephan
  G{\"u}nnemann.
\newblock Pitfalls of graph neural network evaluation.
\newblock \emph{arXiv preprint arXiv:1811.05868}, 2018.

\bibitem[Simonovsky and Komodakis(2018)]{simonovsky2018graphvae}
Martin Simonovsky and Nikos Komodakis.
\newblock Graphvae: Towards generation of small graphs using variational
  autoencoders.
\newblock \emph{arXiv preprint arXiv:1802.03480}, 2018.

\bibitem[Sinha et~al.(2020)Sinha, Sodhani, Pineau, and
  Hamilton]{sinha2020evaluating}
Koustuv Sinha, Shagun Sodhani, Joelle Pineau, and William~L Hamilton.
\newblock Evaluating logical generalization in graph neural networks.
\newblock \emph{arXiv preprint arXiv:2003.06560}, 2020.

\bibitem[Srinivasan and Ribeiro(2020)]{Srinivasan2020On}
Balasubramaniam Srinivasan and Bruno Ribeiro.
\newblock On the equivalence between positional node embeddings and structural
  graph representations.
\newblock In \emph{International Conference on Learning Representations}, 2020.
\newblock URL \url{https://openreview.net/forum?id=SJxzFySKwH}.

\bibitem[Stark et~al.(2006)Stark, Breitkreutz, Reguly, Boucher, Breitkreutz,
  and Tyers]{10.1093/nar/gkj109}
Chris Stark, Bobby-Joe Breitkreutz, Teresa Reguly, Lorrie Boucher, Ashton
  Breitkreutz, and Mike Tyers.
\newblock Biogrid: a general repository for interaction datasets.
\newblock \emph{Nucleic acids research}, 34\penalty0 (suppl\_1):\penalty0
  D535--D539, 2006.

\bibitem[Tang et~al.(2015)Tang, Qu, Wang, Zhang, Yan, and Mei]{tang2015line}
Jian Tang, Meng Qu, Mingzhe Wang, Ming Zhang, Jun Yan, and Qiaozhu Mei.
\newblock Line: Large-scale information network embedding.
\newblock In \emph{Proceedings of the 24th International Conference on World
  Wide Web}, pages 1067--1077. International World Wide Web Conferences
  Steering Committee, 2015.

\bibitem[Tenenbaum et~al.(2000)Tenenbaum, De~Silva, and
  Langford]{tenenbaum2000global}
Joshua~B Tenenbaum, Vin De~Silva, and John~C Langford.
\newblock A global geometric framework for nonlinear dimensionality reduction.
\newblock \emph{science}, 290\penalty0 (5500):\penalty0 2319--2323, 2000.

\bibitem[Tifrea et~al.(2018)Tifrea, Becigneul, and Ganea]{tifrea2018poincare}
Alexandru Tifrea, Gary Becigneul, and Octavian-Eugen Ganea.
\newblock Poincare glove: Hyperbolic word embeddings.
\newblock In \emph{International Conference on Learning Representations}, 2018.

\bibitem[Tsitsulin et~al.(2018)Tsitsulin, Mottin, Karras, Bronstein, and
  M\"{u}ller]{tsitsulin2018netlsd}
Anton Tsitsulin, Davide Mottin, Panagiotis Karras, Alexander Bronstein, and
  Emmanuel M\"{u}ller.
\newblock Netlsd: Hearing the shape of a graph.
\newblock In \emph{Proceedings of the 24th ACM SIGKDD International Conference
  on Knowledge Discovery \& Data Mining}, KDD ’18, page 2347–2356, 2018.

\bibitem[Tsitsulin et~al.(2020{\natexlab{a}})Tsitsulin, Munkhoeva, and
  Perozzi]{tsitsulin2020slaq}
Anton Tsitsulin, Marina Munkhoeva, and Bryan Perozzi.
\newblock Just slaq when you approximate: Accurate spectral distances for
  web-scale graphs.
\newblock In \emph{Proceedings of The Web Conference 2020}, WWW ’20, page
  2697–2703, 2020{\natexlab{a}}.

\bibitem[Tsitsulin et~al.(2020{\natexlab{b}})Tsitsulin, Palowitch, Perozzi, and
  M{\"u}ller]{tsitsulin2020graph}
Anton Tsitsulin, John Palowitch, Bryan Perozzi, and Emmanuel M{\"u}ller.
\newblock Graph clustering with graph neural networks.
\newblock \emph{arXiv preprint arXiv:2006.16904}, 2020{\natexlab{b}}.

\bibitem[Vaswani et~al.(2017)Vaswani, Shazeer, Parmar, Uszkoreit, Jones, Gomez,
  Kaiser, and Polosukhin]{vaswani2017attention}
Ashish Vaswani, Noam Shazeer, Niki Parmar, Jakob Uszkoreit, Llion Jones,
  Aidan~N Gomez, {\L}ukasz Kaiser, and Illia Polosukhin.
\newblock Attention is all you need.
\newblock In \emph{Advances in neural information processing systems}, pages
  5998--6008, 2017.

\bibitem[Veli{\v{c}}kovi{\'{c}} et~al.(2018)Veli{\v{c}}kovi{\'{c}}, Cucurull,
  Casanova, Romero, Lio, and Bengio]{velickovic2017graph}
Petar Veli{\v{c}}kovi{\'{c}}, Guillem Cucurull, Arantxa Casanova, Adriana
  Romero, Pietro Lio, and Yoshua Bengio.
\newblock Graph attention networks.
\newblock In \emph{International Conference on Learning Representations}, 2018.

\bibitem[Veli{\v{c}}kovi{\'{c}} et~al.(2019)Veli{\v{c}}kovi{\'{c}}, Fedus,
  Hamilton, Liò, Bengio, and Hjelm]{velickovic2018dgi}
Petar Veli{\v{c}}kovi{\'{c}}, William Fedus, William~L. Hamilton, Pietro Liò,
  Yoshua Bengio, and R~Devon Hjelm.
\newblock Deep graph infomax.
\newblock In \emph{International Conference on Learning Representations}, 2019.

\bibitem[Verma and Zhang(2019)]{verma2019stability}
Saurabh Verma and Zhi-Li Zhang.
\newblock Stability and generalization of graph convolutional neural networks.
\newblock In \emph{Proceedings of the 25th ACM SIGKDD International Conference
  on Knowledge Discovery \& Data Mining}, pages 1539--1548, 2019.

\bibitem[Vincent et~al.(2010)Vincent, Larochelle, Lajoie, Bengio, and
  Manzagol]{vincent2010stacked}
Pascal Vincent, Hugo Larochelle, Isabelle Lajoie, Yoshua Bengio, and
  Pierre-Antoine Manzagol.
\newblock Stacked denoising autoencoders: Learning useful representations in a
  deep network with a local denoising criterion.
\newblock \emph{Journal of machine learning research}, 11\penalty0
  (Dec):\penalty0 3371--3408, 2010.

\bibitem[Vishwanathan et~al.(2010)Vishwanathan, Schraudolph, Kondor, and
  Borgwardt]{vishwanathan2010graphkernels}
S.~V.~N. Vishwanathan, N.~N. Schraudolph, R.~Kondor, and K.~M Borgwardt.
\newblock Graph kernels.
\newblock In \emph{Journal of Machine Learning Research}, pages 1201--1242,
  2010.

\bibitem[Wang et~al.(2016)Wang, Cui, and Zhu]{wang2016structural}
Daixin Wang, Peng Cui, and Wenwu Zhu.
\newblock Structural deep network embedding.
\newblock In \emph{Proceedings of the 22nd ACM SIGKDD international conference
  on Knowledge discovery and data mining}, pages 1225--1234. ACM, 2016.

\bibitem[Wang et~al.(2019)Wang, Yu, Zheng, Gan, Gai, Ye, Li, Zhou, Huang, Ma,
  et~al.]{wang2019deep}
Minjie Wang, Lingfan Yu, Da~Zheng, Quan Gan, Yu~Gai, Zihao Ye, Mufei Li,
  Jinjing Zhou, Qi~Huang, Chao Ma, et~al.
\newblock Deep graph library: Towards efficient and scalable deep learning on
  graphs.
\newblock \emph{arXiv preprint arXiv:1909.01315}, 2019.

\bibitem[Weston et~al.(2008)Weston, Ratle, and Collobert]{weston2012deep}
Jason Weston, Fr{\'e}d{\'e}ric Ratle, and Ronan Collobert.
\newblock Deep learning via semi-supervised embedding.
\newblock In \emph{Proceedings of the 25th international conference on Machine
  learning}, pages 1168--1175. ACM, 2008.

\bibitem[Wu et~al.(2019)Wu, Pan, Chen, Long, Zhang, and
  Yu]{wu2019comprehensive}
Zonghan Wu, Shirui Pan, Fengwen Chen, Guodong Long, Chengqi Zhang, and Philip~S
  Yu.
\newblock A comprehensive survey on graph neural networks.
\newblock \emph{arXiv preprint arXiv:1901.00596}, 2019.

\bibitem[Xu et~al.(2018)Xu, Hu, Leskovec, and Jegelka]{xu2018powerful}
Keyulu Xu, Weihua Hu, Jure Leskovec, and Stefanie Jegelka.
\newblock How powerful are graph neural networks?
\newblock \emph{arXiv preprint arXiv:1810.00826}, 2018.

\bibitem[Yanardag and Vishwanathan(2015)]{yanardag2015deep}
Pinar Yanardag and S.V.N. Vishwanathan.
\newblock Deep graph kernels.
\newblock In \emph{Proceedings of the 21th ACM SIGKDD International Conference
  on Knowledge Discovery and Data Mining}, pages 1365--1374. Association for
  Computing Machinery, 2015.

\bibitem[Yang et~al.(2016)Yang, Cohen, and Salakhutdinov]{yang2016revisiting}
Zhilin Yang, William~W Cohen, and Ruslan Salakhutdinov.
\newblock Revisiting semi-supervised learning with graph embeddings.
\newblock In \emph{Proceedings of the 33rd International Conference on
  International Conference on Machine Learning-Volume 48}, pages 40--48. JMLR.
  org, 2016.

\bibitem[Ying et~al.(2018{\natexlab{a}})Ying, He, Chen, Eksombatchai, Hamilton,
  and Leskovec]{ying2018pinsage}
Rex Ying, Ruining He, Kaifeng Chen, Pong Eksombatchai, William Hamilton, and
  Jure Leskovec.
\newblock Graph convolutional neural networks for web-scale recommender
  systems.
\newblock In \emph{Proceedings of the 24th ACM SIGKDD International Conference
  on Knowledge Discovery \& Data Mining}, 2018{\natexlab{a}}.

\bibitem[Ying et~al.(2018{\natexlab{b}})Ying, You, Morris, Ren, Hamilton, and
  Leskovec]{ying2018hierarchical}
Zhitao Ying, Jiaxuan You, Christopher Morris, Xiang Ren, Will Hamilton, and
  Jure Leskovec.
\newblock Hierarchical graph representation learning with differentiable
  pooling.
\newblock In S.~Bengio, H.~Wallach, H.~Larochelle, K.~Grauman, N.~Cesa-Bianchi,
  and R.~Garnett, editors, \emph{Advances in Neural Information Processing
  Systems 31}, pages 4800--4810. Curran Associates, Inc., 2018{\natexlab{b}}.
\newblock URL
  \url{http://papers.nips.cc/paper/7729-hierarchical-graph-representation-learning-with-differentiable-pooling.pdf}.

\bibitem[You et~al.(2018)You, Ying, Ren, Hamilton, and
  Leskovec]{you2018graphrnn}
Jiaxuan You, Rex Ying, Xiang Ren, William~L Hamilton, and Jure Leskovec.
\newblock Graphrnn: A deep generative model for graphs.
\newblock \emph{arXiv preprint arXiv:1802.08773}, 2018.

\bibitem[You et~al.(2019)You, Ying, and Leskovec]{you2019position}
Jiaxuan You, Rex Ying, and Jure Leskovec.
\newblock Position-aware graph neural networks.
\newblock \emph{arXiv preprint arXiv:1906.04817}, 2019.

\bibitem[Yu and De~Sa(2019)]{yu2019numerically}
Tao Yu and Christopher~M De~Sa.
\newblock Numerically accurate hyperbolic embeddings using tiling-based models.
\newblock In \emph{Advances in Neural Information Processing Systems}, pages
  2023--2033, 2019.

\bibitem[Zhang et~al.(2018{\natexlab{a}})Zhang, Yin, Zhu, and
  Zhang]{zhang2018network}
Daokun Zhang, Jie Yin, Xingquan Zhu, and Chengqi Zhang.
\newblock Network representation learning: A survey.
\newblock \emph{IEEE Transactions on Big Data}, 2018{\natexlab{a}}.

\bibitem[Zhang et~al.(2018{\natexlab{b}})Zhang, Cui, Neumann, and
  Chen]{zhang2018end}
Muhan Zhang, Zhicheng Cui, Marion Neumann, and Yixin Chen.
\newblock An end-to-end deep learning architecture for graph classification.
\newblock In \emph{Thirty-Second AAAI Conference on Artificial Intelligence},
  2018{\natexlab{b}}.

\bibitem[Zhang et~al.(2018{\natexlab{c}})Zhang, Cui, and Zhu]{zhang2018deep}
Ziwei Zhang, Peng Cui, and Wenwu Zhu.
\newblock Deep learning on graphs: A survey.
\newblock \emph{arXiv preprint arXiv:1812.04202}, 2018{\natexlab{c}}.

\bibitem[Zhou et~al.(2004)Zhou, Bousquet, Lal, Weston, and
  Sch{\"o}lkopf]{zhou2004learning}
Dengyong Zhou, Olivier Bousquet, Thomas~N Lal, Jason Weston, and Bernhard
  Sch{\"o}lkopf.
\newblock Learning with local and global consistency.
\newblock In \emph{Advances in neural information processing systems}, pages
  321--328, 2004.

\bibitem[Zhou et~al.(2018)Zhou, Cui, Zhang, Yang, Liu, Wang, Li, and
  Sun]{zhou2018graph}
Jie Zhou, Ganqu Cui, Zhengyan Zhang, Cheng Yang, Zhiyuan Liu, Lifeng Wang,
  Changcheng Li, and Maosong Sun.
\newblock Graph neural networks: A review of methods and applications.
\newblock \emph{arXiv preprint arXiv:1812.08434}, 2018.

\bibitem[Zhu and Ghahramani(2002)]{zhu2002learning}
Xiaojin Zhu and Zoubin Ghahramani.
\newblock Learning from labeled and unlabeled data with label propagation.
\newblock 2002.

\end{thebibliography}
\end{document}